\documentclass{article}

\usepackage[main, preprint]{neurips_2026}
\usepackage[utf8]{inputenc}
\usepackage[T1]{fontenc}

\usepackage{microtype}
\usepackage{graphicx}
\usepackage{subcaption}
\usepackage{booktabs} 
\usepackage{multirow}
\usepackage{hyperref}
\usepackage{url}

\usepackage{amsmath}
\usepackage{amssymb}
\usepackage{mathtools}
\usepackage[most]{tcolorbox}
\usepackage{amsthm}
\usepackage{adjustbox}
\usepackage{xspace}
\usepackage{xcolor}
\usepackage{textcomp}
\usepackage{stfloats}

\usepackage{tikz}
\usetikzlibrary{calc}
\usepackage{pgfplots}
\usepackage{adjustbox}
\usetikzlibrary{shapes.geometric, arrows.meta, positioning, calc}

\pgfplotsset{compat=1.18}

\usepackage[capitalize,noabbrev]{cleveref}

\theoremstyle{plain}

\theoremstyle{definition}

\theoremstyle{remark}

\newcommand{\stdf}[1]{\ \mbox{\scriptsize $\pm$ #1}}
\newcommand{\ours}{NNS\xspace}

\title{Simplifying Neural Networks During Training}

\author{%
  Lorenzo Sciandra\thanks{Equal contribution.} \\
  Department of Computer Science, University of Turin, Turin, Italy \\
  \texttt{lorenzo.sciandra@unito.it}
  \And
  Samuele Fonio\footnotemark[1] \\
  Department of Computer Science, University of Turin, Turin, Italy \\
  \texttt{samuele.fonio@unito.it}
  \And
  Roberto Esposito \\
  Department of Computer Science, University of Turin, Turin, Italy \\
  \texttt{roberto.esposito@unito.it}
}

\definecolor{RoyalBlue}{RGB}{65, 105, 225}

\begin{document}

\maketitle

\begin{abstract}
  Understanding and exploiting the training dynamics of overparameterized deep neural networks remains a central challenge in modern machine learning. Recent evidence on Neural Collapse (NC) shows that class representations and classifiers exhibit highly structured geometry, while the Tunnel Effect suggests that only a subset of layers is essential for feature extraction. We combine these two perspectives and propose an NC-inspired training framework for simplifying deep networks during training. Our method monitors representation dynamics through the Inverse Fisher Criterion, a stable and efficient proxy for the variability collapse behavior, to identify both the split point between feature extraction and classification and the training stage at which simplification becomes viable. We then replace the trailing layers with a lightweight classification head and continue training the reduced model. Experiments on image-classification benchmarks across MLP, VGG, and ResNet architectures show that the proposed method achieves substantial parameter reductions while maintaining accuracy comparable to that of the full model. Code to reproduce the experiments can be found at: \url{https://github.com/LorenzoSciandra/NNS}.
\end{abstract}

\section{Introduction}
\label{sec:intro}

Over the past decade, Deep Neural Networks (DNNs) have become the dominant paradigm in modern machine learning, consistently advancing the state of the art on benchmark tasks and achieving superhuman performance in a wide range of domains \cite{wang_scientific_2023}. Despite this empirical success, our theoretical understanding has lagged behind, and many fundamental questions remain open. A notable example is the double-descent phenomenon \cite{belkin_reconciling_2019,nakkiran_deep_2020}, which suggests that overparameterization can improve both optimization and generalization, seemingly contradicting the predictions of classical learning theory \cite{geman_neural_1992,hastie_elements_2009}. In this context, a recent line of work has identified an intriguing phenomenon, termed Neural Collapse (NC), which emerges during the terminal phase of training in a broad class of classification problems. Specifically, in both balanced \cite{papyan2020prevalence, han_neural_2022} and imbalanced \cite{fang_exploring_2021, behnia_implicit_2023} dataset settings, the last-layer representations and classifier of a trained DNN have been observed to converge to a simple and highly structured simplex Equiangular Tight Frame (ETF); see Definition 1 in~\citealp{papyan2020prevalence}.

\citealp{han_flatness_2025} further show, under classical assumptions, that even if NC is not necessary for generalization, it can induce relative flatness of the loss landscape, a property that may contribute to generalization \cite{hochreiter_flat_1997,petzka_relative_2021}. Precisely, they exploit the grokking phenomenon \cite{power_grokking_2022,nanda_progress_2023}, i.e., the ability of a model to generalize after achieving complete memorization, to disentangle necessity from sufficiency phenomena for generalization.

Pursuing a complementary direction, the Information Bottleneck Principle (IBP)~\cite{tishby2000information} offers an information-theoretic interpretation of deep models. Under this view, a DNN can be decomposed into two functional components: a feature extractor and a classification head. The former progressively transforms and compresses the input across layers, while the latter uses the resulting representation to solve the task. Although this structural separation is often assumed implicitly, it has only recently been quantified through the Tunnel Effect (TE)~\cite{masarczyk2023tunnel}. Inspired by IBP, TE emphasizes the role of depth in representation formation and aims to identify the point at which a minimal sufficient statistic has already been attained. Beyond that point, the subsequent layers, referred to as \textit{the tunnel}, mainly propagate and further compress the learned representation.

Since NC is primarily understood as an optimization-driven phenomenon~\cite{hui_limitations_2022, sukenik_neural_2024} that may also emerge in earlier layers~\cite{rangamani2023feature}, we integrate this perspective with the Tunnel Effect to develop an NC-inspired training pipeline for reducing the complexity of modern DNNs that we termed \textbf{Neural Network Simplification} (\ours). \ours starts from a standard overparameterized model and uses a representation metric to identify the \emph{split point} at which the network transitions from feature extraction to classification. Once this transition is detected reliably, we simplify the model during training by removing the trailing layers. Crucially, our goal is not to assert that full Neural Collapse, or the complete ETF geometry, has already been attained before truncation. Rather, we use early variability collapse behavior as a practical signal that the later layers are primarily acting as task-specific compressors, which makes it possible to retain comparable performance with a smaller model on the image-classification architectures considered here.
%
Beyond improving our understanding of overparameterized models, this approach has practical implications for the design of training pipelines that favor smaller models and may therefore help mitigate the environmental cost of increasingly deep networks.

In this context, our contributions are threefold:
\begin{itemize}
    \item We introduce \ours, a training-time method that combines insights from Neural Collapse and the Tunnel Effect to dynamically simplify network architectures during optimization.
    \item We adopt the Inverse Fisher Criterion, a stable and comparatively efficient NC proxy, for detecting the extractor–classifier split across layers and epochs.
    \item We show, through experiments on multiple image-classification architectures and datasets, that \ours achieves large parameter reductions while preserving accuracy comparable to that of the full model.
\end{itemize}


\section{Related Works}
\label{sec:rel_works}

 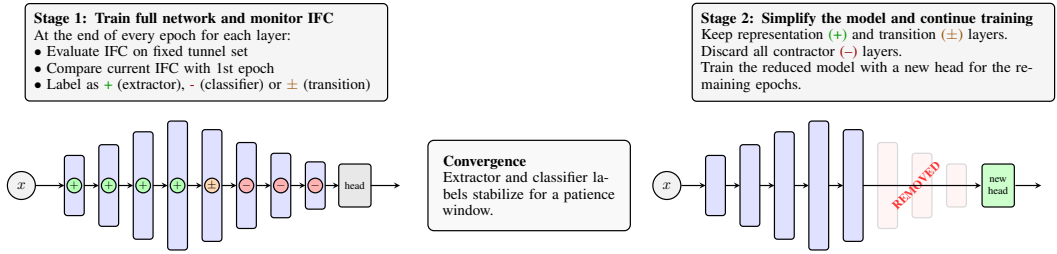
\begin{figure*}[t]

\centering

\resizebox{\textwidth}{!}{

\begin{tikzpicture}[x=1cm,y=1cm,>=stealth]

    \tikzstyle{layer}=[draw=black, rounded corners=2pt, thick, fill=blue!12]

    \tikzstyle{head}=[draw=black, rounded corners=2pt, thick, fill=green!18]

    \tikzstyle{metric}=[draw=black, rounded corners=2pt, thick, fill=orange!15, align=center]

    \tikzstyle{note}=[draw=black, rounded corners=3pt, thick, fill=gray!8, align=center]

    \tikzstyle{line}=[->, thick]

    \tikzstyle{dashedline}=[->, thick, dashed]

    \tikzset{
        layer_keep/.style={draw=black, rounded corners=2pt, thick, fill=green!25},
        layer_discard/.style={draw=black, rounded corners=2pt, thick, fill=red!25},
        head/.style={draw=black, rounded corners=2pt, thick, fill=green!20, minimum height=1cm, minimum width=1.1cm},
        labelbox/.style={draw=black, rounded corners=3pt, thick, fill=gray!8, align=left, inner sep=6pt, text width=5.5cm},
        line/.style={->, thick},
        dashedline/.style={->, thick, dashed},
        marker/.style={draw, circle, thick, minimum size=0.45cm, inner sep=0pt, font=\bfseries}
    }


    \node[draw, circle, thick, minimum size=0.65cm, fill=gray!10] (xin) at (0.2,1.9) {$x$};

    \foreach \x/\h/\name in {1.2/1.4/l1, 2.0/1.9/l2, 2.8/2.5/l3, 3.6/3.0/l4, 4.4/2.6/l5, 5.2/2.0/l6, 6.0/1.5/l7, 6.8/1.1/l8} {
        \draw[layer] (\x, {1.9-\h/2}) rectangle ++(0.46, \h);
    }

    \node[head, fill=gray!18, minimum width=0.75cm, minimum height=1.0cm] (fullhead) at (7.95,1.9) {};

    \node[font=\scriptsize] at (7.95,1.9) {head};

    \draw[line] (xin) -- (1.2,1.9);

    \foreach \x/\xn in {1.66/2.0,2.46/2.8,3.26/3.6,4.06/4.4,4.86/5.2,5.66/6.0,6.46/6.8,7.26/7.58} \draw[line] (\x,1.9) -- (\xn,1.9);

    \draw[line] (8.32,1.9) -- (9.0,1.9);


    \node[labelbox, text width=8cm] (ifcbox) at (4.5,5.0)
        {\textbf{Stage 1: Train full network and monitor IFC}\\
        At the end of every epoch for each layer: \\
         $\bullet$ Evaluate IFC on fixed tunnel set\\
         $\bullet$ Compare current IFC with 1st epoch\\
         $\bullet$ Label as \textcolor{green!60!black}{+} (extractor), \textcolor{red!60!black}{-} (classifier) or \textcolor{orange!60!black}{$\pm$} (transition)};


    \node[draw, circle, thick, fill=green!30, minimum size=0.34cm, inner sep=0pt, font=\scriptsize] at (1.43,1.9) {$+$};
    \node[draw, circle, thick, fill=green!30, minimum size=0.34cm, inner sep=0pt, font=\scriptsize] at (2.23,1.9) {$+$};
    \node[draw, circle, thick, fill=green!30, minimum size=0.34cm, inner sep=0pt, font=\scriptsize] at (3.03,1.9) {$+$};
    \node[draw, circle, thick, fill=green!30, minimum size=0.34cm, inner sep=0pt, font=\scriptsize] at (3.83,1.9) {$+$};
    \node[draw, circle, thick, fill=orange!30, minimum size=0.34cm, inner sep=0pt, font=\scriptsize] at (4.63,1.9) {$\pm$};
    \node[draw, circle, thick, fill=red!30, minimum size=0.34cm, inner sep=0pt, font=\scriptsize] at (5.43,1.9) {$-$};
    \node[draw, circle, thick, fill=red!30, minimum size=0.34cm, inner sep=0pt, font=\scriptsize] at (6.23,1.9) {$-$};
    \node[draw, circle, thick, fill=red!30, minimum size=0.34cm, inner sep=0pt, font=\scriptsize] at (7.03,1.9) {$-$};




    \node[note, draw=black, thick, fill=gray!5, rounded corners=4pt, inner sep=10pt, align=left, text width=4.0cm] (conv) at (12.0, 1.9) {
        \textbf{Convergence}\\
        Extractor and classifier labels stabilize for a patience window.
    };


    \node[draw, circle, thick, minimum size=0.65cm, fill=gray!10] (xin2) at (15.2,1.9) {$x$};

    \foreach \x/\h in {16.1/1.4, 16.9/1.9, 17.7/2.5, 18.5/3.0, 19.3/2.6} {
        \draw[layer] (\x, {1.9-\h/2}) rectangle ++(0.46, \h);
    }


    \begin{scope}[opacity=0.15]
        \draw[layer_discard] (20.1, 1.9-1.0) rectangle ++(0.46, 2.0);
        \draw[layer_discard] (20.9, 1.9-0.75) rectangle ++(0.46, 1.5);
        \draw[layer_discard] (21.7, 1.9-0.5) rectangle ++(0.46, 1.0);
    \end{scope}
    \node[red!80, font=\bfseries\footnotesize, rotate=45] at (21.0, 1.9) {REMOVED};

    \node[head, minimum width=0.75cm, minimum height=1.0cm] (newhead) at (22.9,1.9) {};

    \node[font=\scriptsize, align=center] at (22.9,1.9) {new\\head};

    \draw[line] (xin2) -- (16.1,1.9);

    \foreach \x/\xn in {16.56/16.9,17.36/17.7,18.16/18.5,18.96/19.3,19.76/22.5} \draw[line] (\x,1.9) -- (\xn,1.9);

    \draw[line] (23.3,1.9) -- (23.9,1.9);

    \node[labelbox, text width=8cm] (split_info) at (20.0, 5.0)
        {\textbf{Stage 2: Simplify the model and continue training}\\
         Keep representation \textcolor{green!60!black}{(+)} and transition \textcolor{orange!60!black}{($\pm$)} layers.\\
         Discard all contractor \textcolor{red!60!black}{(--)} layers.\\
         Train the reduced model with a new head for the remaining epochs.};



\end{tikzpicture}

}

\caption{Visual overview of the \ours pipeline. A full overparameterized network is trained while the Inverse Fisher Criterion is monitored after each epoch on the fixed tunnel set for all the layers. Once the layer-wise trends stabilize, the transition region is identified, the classifier-like tail is removed, a lightweight head is attached, and training continues on the reduced model for the remaining epochs.}

\label{fig:nns_pipeline}

\end{figure*} 

\paragraph{Neural Collapse and Simplex ETFs}
Neural Collapse is an empirically observed phenomenon that characterizes the geometry of penultimate-layer features in overparameterized networks trained to zero training loss.
From a theoretical perspective, the ETF structure of the class means and final-layer weights is the unique set of global optima for several loss functions when features are treated as free parameters \citep{zhu2021geometric, zhou2022all, zhou_optimization_2022}. \citealp{rangamani2023feature} then investigate whether NC extends to intermediate layers, and they empirically show that ETF-like geometry persists across several of the final layers. Their interpretation is that the network first maps the input into a high-dimensional space in which linear decision boundaries become viable, and then refines these representations by reducing within-class variability and increasing between-class separation. Notably, they also show that fixing these layers post hoc as simplex ETFs can significantly reduce the number of parameters with only negligible degradation in accuracy. Building on this intuition, we instead aim to discard redundant late layers during training and replace them with a lightweight classification head.
\citealp{markou2024guiding} exploit NC from a different angle, guiding optimization toward the nearest simplex ETF at each iteration. Concretely, the classifier weights are set implicitly through a Riemannian optimization problem.
We evaluate this technique in our framework as one possible classification head for the simplified model.

\paragraph{Intrinsic Dimension and Tunnel Effect}
The work of \citealp{ansuini2019intrinsic} studies the expressivity of the feature spaces induced across the layers of a DNN. They define the Intrinsic Dimension (ID) of a representation as the minimum number of coordinates needed to describe the data without significant information loss. Their analysis shows that, in trained networks, data representations often lie on curved manifolds whose dimensionality is orders of magnitude smaller than that of the ambient embedding space.
Moreover, ID typically increases in the early layers and then decreases in the final ones. This pattern is consistent with a standard view of network dynamics: early layers improve discrimination by embedding data in increasingly expressive spaces, while later layers perform task-specific compression. \citealp{masarczyk2023tunnel} sharpen this picture through the Tunnel Effect. They show empirically that the initial layers, referred to as the \textit{extractor}, establish linearly separable representations, whereas the subsequent layers, the \textit{tunnel}, compress those representations with little effect on final performance. They further show that this extractor-tunnel split, which depends strongly on the interplay between model capacity and task complexity, emerges early in training and persists thereafter.
Their experiments also demonstrate that a shallower network can match the deep model as long as it retains at least the capacity of the extractor. This motivates our effort to track the dynamics of this split point during training. In this sense, our contribution is not a new NC metric per se, but the use of an NC-inspired online signal to operationalize the TE split during optimization.

\paragraph{Pruning and Lottery Ticket Hypothesis}
Although our training pipeline is related to pruning, it differs from standard pruning practice in several fundamental ways. The usual \emph{progressive pruning and retraining} paradigm trains a large model to convergence, prunes it, and then fine-tunes the reduced architecture to recover performance. By contrast, our method acts during training by identifying and removing redundant final layers before full convergence. In this respect, it is reminiscent of \citealp{you_drawing_2020}, which seeks small subnetworks early in training and evaluates them either from scratch or by continuing optimization. That line of work builds on the \emph{Lottery Ticket Hypothesis} \cite{frankle_lottery_2019}, which states that dense, randomly initialized networks contain sparse subnetworks
capable of matching the full model’s performance.
In \citealp{you_drawing_2020}, binary masks are constructed at each epoch by pruning weights with the smallest $l_1$ norms, and stabilization is detected via the Hamming distance between consecutive masks.
Despite these conceptual connections, our method differs substantially from existing pruning strategies. First, it does not require a predefined pruning budget, but determines the degree of reduction directly from the observed training dynamics. Second, guided by representation-learning considerations, it restricts simplification to the final layers, where redundancy is most likely to arise. The two approaches are also compatible rather than exclusive: in our experiments, we additionally tested a hybrid setting in which EB-LTH is applied after \ours on the simplified model.


\section{Neural Network Simplification}

The goal of this paper is to introduce Neural Network Simplification, a method that combines the Tunnel Effect perspective with the Neural Collapse phenomenon to simplify overparameterized deep networks \emph{during} training. Concretely, we study whether the final layers of a network can be removed at an intermediate training stage while preserving performance comparable to that of the original model. The key idea is that, if the dynamics of an overparameterized DNN can be monitored across epochs and the transition from representation to classification can be identified reliably, then the remaining tail of the network can be replaced by a lightweight head. TE indicates that such a split exists, but identifies it only \emph{a posteriori}. By contrast, NC provides training signals that stabilize over time. We therefore do not attempt to detect the full onset of all NC properties before truncation; instead, we use an \textbf{NC-related representation metric} as a signal to track when late layers behave primarily as task-specific compressors.

To reach our objective, we must answer two key questions:
\begin{itemize}
    \item \textbf{Where} in a DNN lies the splitting point that separates the feature extractor from the task-specific component?
    \item \textbf{When}, during training, can this division be confidently identified, enabling a simplification of the model? 
\end{itemize}

To do so, we seek a representation metric that can determine whether a layer belongs to the \textit{feature extractor} or to the \textit{classification head}. Such a metric should satisfy four requirements:
 
$i)$ it must be computationally efficient, since it is evaluated at every epoch and for all monitored layers; $ii)$ it must remain tractable on large datasets;
$iii)$ it should converge reliably, so that an early split decision remains valid throughout training; and $iv)$ it should converge quickly, thereby reducing the overhead of training the full model.

Requirement $ii)$ can be addressed by evaluating the metrics on a small class-balanced subset of the training data, which we call the \textbf{tunnel set}. In the experiments reported here, the tunnel set contains 5\% of the training data and is sampled to preserve the original class distribution. All representation metrics are computed on this subset. Appendix~\ref{app:tunnel_set} reports experiments on the tunnel set size.

\subsection{The Inverse Fisher Criterion}

\label{sec:rep_metrics}

To identify a principled criterion for simplifying deep neural networks during training, we build upon the NC phenomenon, which describes a set of geometric regularities emerging in the representations and in the classification head of overparameterized models as training progresses. More precisely, NC is characterized by four properties. In our context, the most relevant one is \emph{variability collapse} (NC1), which describes the progressive concentration of features around their class means. For completeness, we recall that NC also includes \emph{convergence to simplex equiangular tight frame} (NC2), \emph{convergence to self-duality} (NC3), and \emph{simplification to nearest class center} (NC4). A formal definition of all the properties is in Appendix \ref{app:nc_definition}.

Before introducing the representation metric we used in \ours, it is worth noting that in~\citealp{masarczyk2023tunnel}, the authors identify the split point between the \emph{extractor} and the \emph{tunnel} only \emph{after} full training, by attaching a linear probe to each layer and selecting the first layer whose probe reaches at least 95\% of the final accuracy of the original model. This provides a useful \emph{a posteriori} characterization, but it is not directly applicable in our setting because it requires training the full network to convergence. Our goal is therefore to replace this post-hoc criterion with an online signal that can be monitored during training.

Several representation metrics have been proposed as computationally efficient proxies for the NC properties. After evaluating a range of such metrics across layers and epochs, we found the \textbf{Inverse Fisher Criterion} (IFC) to be the most reliable and consistent indicator, and we therefore use it as our split signal. IFC, defined as $\sigma_W/\sigma_B$, is a practical proxy for the variability-collapse aspect of NC1 since it measures the ratio between total within-class variance $\sigma_W$ and total between-class variance $\sigma_B$ without requiring the full covariance-based NC1 computation, which is too expensive to evaluate repeatedly across layers and epochs.
This metric can be interpreted as the inverse of the Fisher linear discriminant criterion \cite{fisher_use_1936}, and was already used in \citealp{hui_limitations_2022, sukenik_neural_2024} to measure the degree of feature collapse since it is considered more stable than other metrics. Our contribution is therefore not the introduction of IFC itself, but its use as an online criterion for split-point detection. Intuitively, the IFC captures the isotropic collapse of within-class variability characteristic of NC, while avoiding the numerical instability and normalization issues of alternative metrics. In this sense, IFC can be interpreted as a minimal and comparatively efficient relaxation of the NC1 condition, suitable for online monitoring during training.


Formal definitions of all candidate representation metrics we tested for \ours are in Appendix~\ref{app:metric_def}.

\paragraph{IFC behaviour}
\begin{figure*}[htbp]
    \centering
\begin{subfigure}{0.33\columnwidth}
  \centering
  \includegraphics[width=\columnwidth]{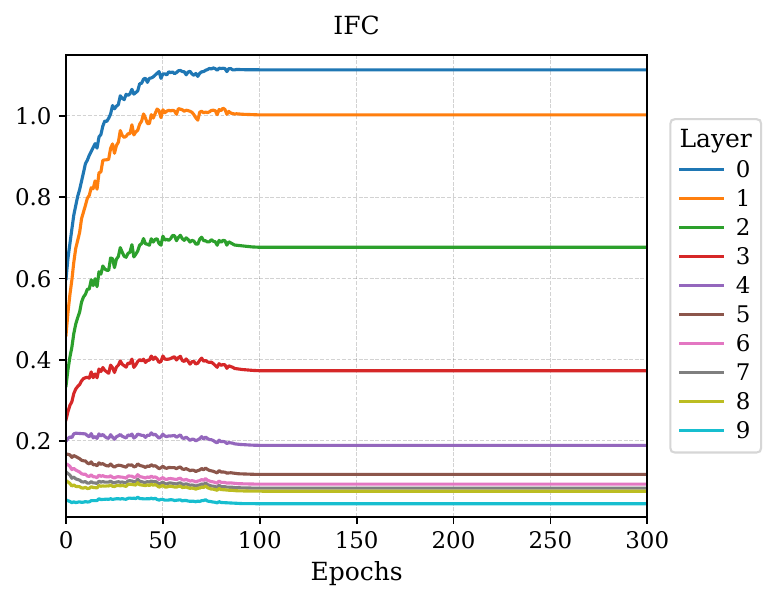}
  \caption{MLP10 Fashion-MNIST}
  \label{subfig:proxy_nc_mlp10_fashionmnist}
\end{subfigure}%
\begin{subfigure}{0.33\columnwidth}
  \centering
  \includegraphics[width=\columnwidth]{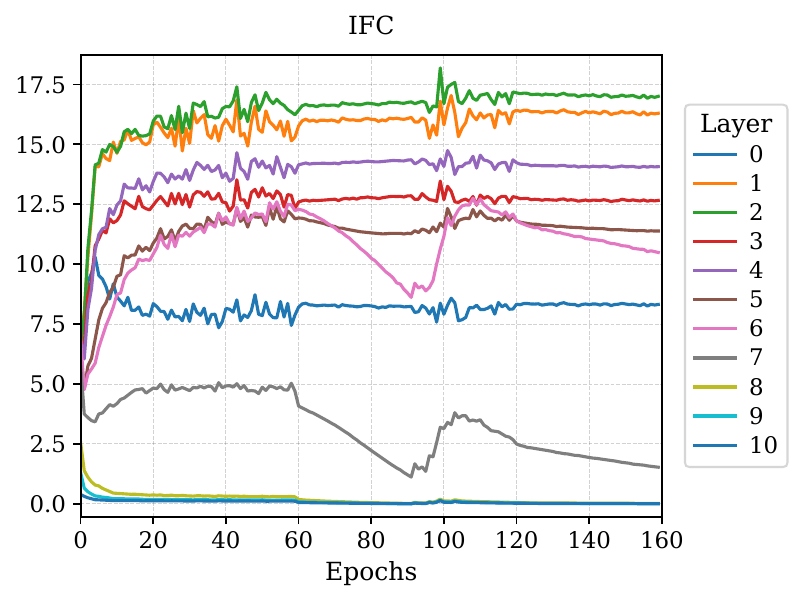}
  \caption{ResNet18 CIFAR-10}
  \label{subfig:proxy_nc_resnet18_cifar10}
\end{subfigure}
\begin{subfigure}{0.33\columnwidth}
  \centering
  \includegraphics[width=\columnwidth]{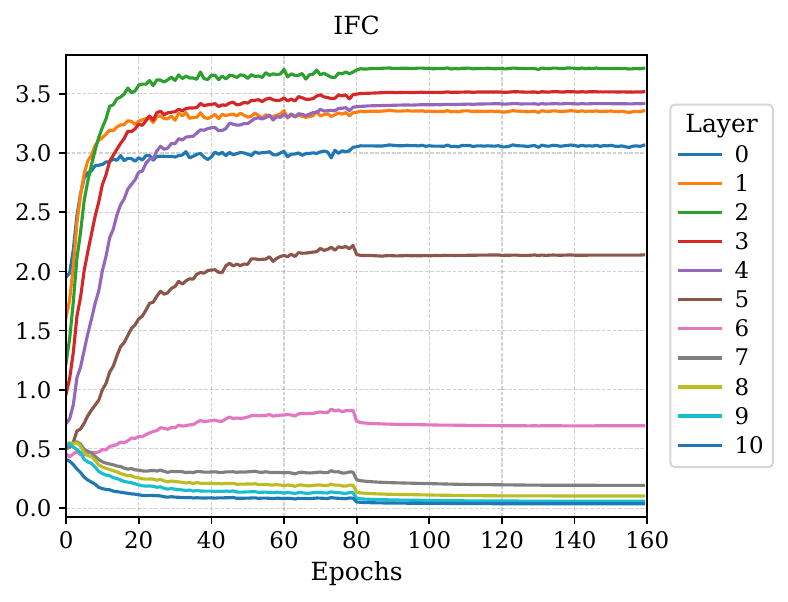}
  \caption{VGG11 CIFAR-100}
  \label{subfig:proxy_nc_vgg11_cifar100}
\end{subfigure}

    \caption{Evolution of the Inverse Fisher Criterion across training epochs for all the layers.}
    \label{fig:metrics_compare}
\end{figure*}

Since all the tested representation metrics rely heavily on the dataset and on the architecture, as they are all representation-based, we evaluate them on multiple architectures and tasks. Specifically, we consider datasets of varied complexity: Fashion-MNIST~\citep{xiao_fashion-mnist_2017}, CIFAR-10~\citep{krizhevsky_learning_2009}, CIFAR-100~\citep{krizhevsky_learning_2009}, and CUB~\citep{wah_cub-200-2011_2011}. For the architecture, we use three different families with varying depth: MLP~\citep{bebis_feed-forward_1994}, VGGs~\citep{simonyan_very_2015}, and ResNets~\citep{he_deep_2016}.  We refer to Appendix~\ref{app:exp_set} for a complete description of the experimental setting.

Throughout the main paper, we use three representative configurations as guiding examples: an MLP10 trained on Fashion-MNIST, a ResNet18 trained on CIFAR-10, and a VGG11 trained on CIFAR-100. Note that, while for MLPs and VGGs, each layer can be a potential splitting point between the feature extractor and the classifier of the network, for the ResNets, we consider each residual block as a single unit for splitting.

Empirically, as shown in Figure \ref{fig:metrics_compare}, IFC exhibits a consistent and interpretable behavior across all the architectures and datasets studied here. Early layers tend to increase the value of the criterion, reflecting a progressive enhancement of class separability, whereas later layers tend to decrease it, indicating a transition toward task-specific classification. Importantly, this pattern emerges early in training and remains stable across epochs, enabling both the identification of the split point and the determination of the split epoch. Based on its computational efficiency, numerical stability, and empirical consistency on these benchmarks, we adopt IFC as the representation metric for \ours procedure. 
Evaluating composite criteria that combine IFC with NC2-aware or manifold-geometry-aware quantities is a natural direction for future work, but designing such combinations in a way that preserves interpretability and stability is non-trivial. 

A detailed comparison with all the alternative metrics is provided in Appendix~\ref{app:metric_comparison}.

\subsection{Network split identification}
\label{ssec:find_split}

Since IFC reflects the expressive role of the underlying feature space, we label each layer $l$ according to the difference between its value at epoch $t$ and its value after the first epoch. We interpret this difference as a proxy for the layer’s degree of specialization. If the difference is positive, layer $l$ appears to map the inputs into a more expressive feature space; thus, it is likely performing representation. Indeed, if the primary role of a layer is to construct a higher-dimensional representation space in order to better disentangle samples from different classes, an increase in the IFC naturally arises as a side effect, since individual points become more separated, leading to increased variance.
Conversely, if the difference is negative, the layer tends to reduce the variability, projecting the examples into a lower-dimensional manifold. In this case, its main role is to collapse variability. So, this criterion operationalizes the extractor–tunnel dichotomy observed in the Tunnel Effect, and provides a principled, training-dynamics-based mechanism for identifying the split point. In principle, all layers are candidates for the split; in architectures where single layers are not the most meaningful unit, such as ResNets, we instead monitor residual blocks. Obviously, guided by the assumption that the layers of the neural network can be partitioned into feature extractors and variability contractors, at least one layer must be in each category. So, we will always classify the first layer as an extractor and the last layer as a contractor. 

\paragraph{Splitting the model}

Once the layer labels have stabilized sufficiently, the model can be split. Because training is stochastic, the layer categorization is often noisy in the first few epochs, mainly for the middle layers.
Therefore, we propose to trim the network once the label associated with each layer has remained unchanged for a fixed number of epochs. To this end, we introduce a patience parameter, empirically set to 15 epochs, which we found to be effective in practice. We expect the most problematic layers to lie in the middle of the network, as it is where the network transitions between roles. In light of this, we advocate trimming only the layers that can be confidently associated with the classification, while keeping those on which there is uncertainty. Operationally, when adjacent layers remain plausible splitting candidates, we interpret them as a narrow transition region and resolve the ambiguity conservatively by favoring the slightly deeper split when needed. The rationale is that retaining one additional layer is preferable to over-truncating the network and risking a drop in accuracy.

The \ours pipeline, summarized in Figure~\ref{fig:nns_pipeline}, proceeds as follows:

\begin{enumerate}
    \item Train the original network normally, and after each epoch evaluate IFC on the fixed tunnel set for every layer.
    \item Compare the current IFC of each layer with its value after the first epoch, and label the layer as extractor-like or classifier-like according to the sign of the variation, with the first and last layers fixed by construction.
    \item Track the resulting layer labels across epochs; once all labels remain unchanged for the chosen patience window, identify the stable transition region and, when ambiguity remains, choose the conservative split within that region.
    \item Remove the layers after the split point, attach the selected lightweight classification head, and resume training on the reduced model for the remaining epochs.
\end{enumerate}



Regarding the new classification head to be added to the simplified model after the split point, we propose a straightforward fully connected layer, preceded by a flattening operation when the embedding is not already in vector form. It is important to note that this choice may result in classification heads with a large number of parameters, due to the high dimensionality of feature maps for certain model–dataset combinations, especially if the number of classes is large. Alternative classification heads can be employed in this setting to ensure an effective reduction in the number of parameters. For example, for the CUB dataset, we first apply global pooling, yielding a representation whose dimensionality matches the number of channels. Similarly, one may consider averaging across channels at each spatial location and returning the flattened feature map. An analysis of these alternatives is provided in the Appendix \ref{app:exps_clas_heads}, leaving room for further improvements in future investigations.


\subsection{Split analysis}



For conciseness, here we summarize the whole experimentation, while complete results are in Appendix~\ref{app:split_analysis}.

\begin{figure*}[t]
\centering
\begin{subfigure}{0.33\columnwidth}
  \centering
  \includegraphics[width=\columnwidth]{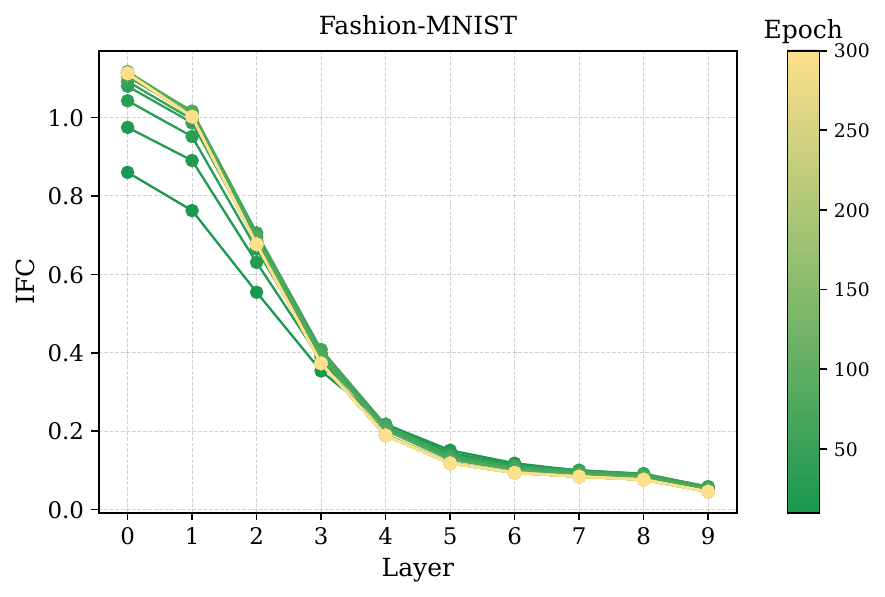}
  \caption{MLP10 Fashion-MNIST}
  \label{subfig:ifc_mlp10_fashionmnist}
\end{subfigure}%
\begin{subfigure}{0.33\columnwidth}
  \centering
  \includegraphics[width=\columnwidth]{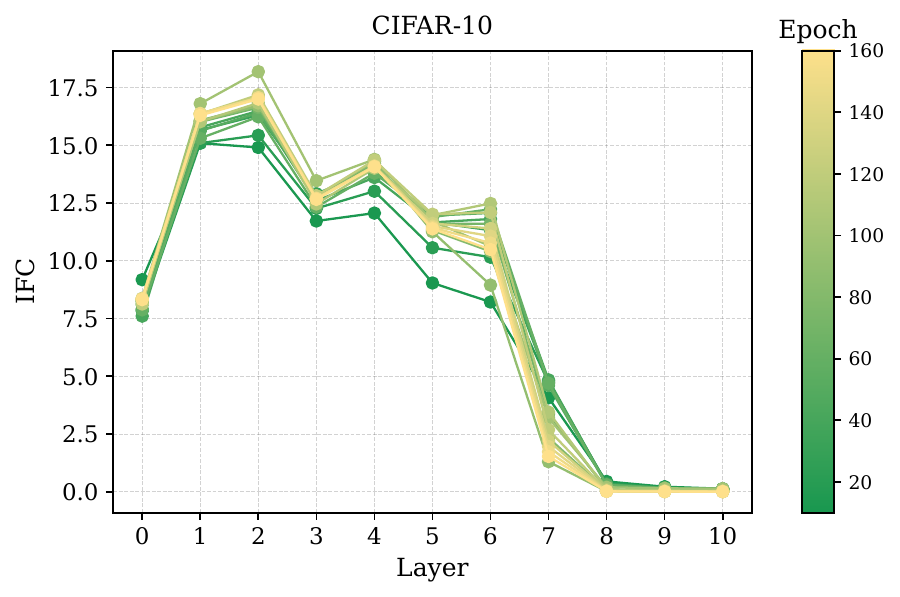}
  \caption{ResNet18 CIFAR-10}
  \label{subfig:ifc_resnet18_cifar10}
\end{subfigure}
\begin{subfigure}{0.33\columnwidth}
  \centering
  \includegraphics[width=\columnwidth]{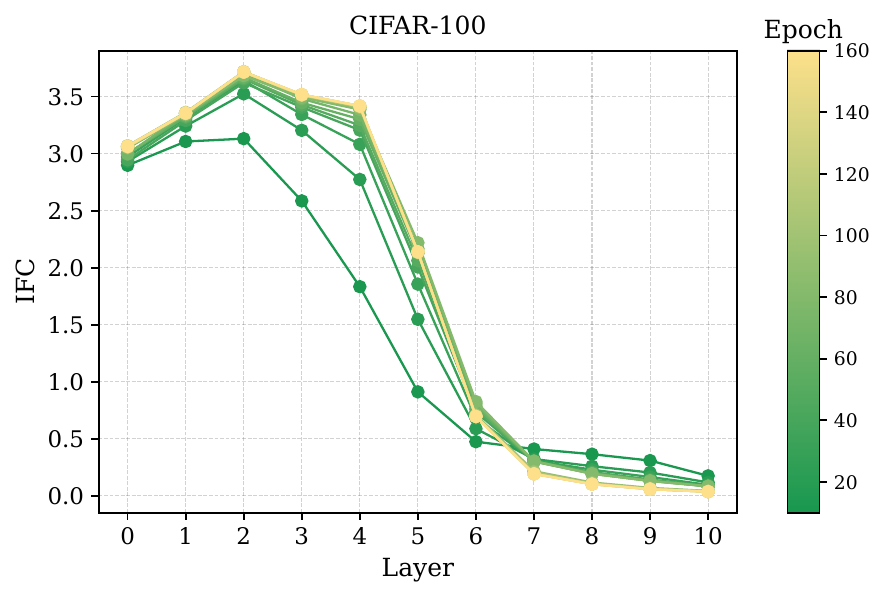}
  \caption{VGG11 CIFAR-100}
  \label{subfig:ifc_vgg11_cifar100}
\end{subfigure}
\caption{The x-axis corresponds to the network layers, ranging from the first to the final hidden layer, while the y-axis corresponds to the IFC metric. For a fixed layer, the points along the vertical line show the evolution of the IFC value across training epochs.}
\label{fig:ifc_main}
\end{figure*}

A first observation we note is that the fraction of total training epochs required before trimming the network depends on both the model \emph{and} the dataset. Nevertheless, the convergence of the IFC metric to its final value is rapid and, as shown in Figure~\ref{fig:ifc_main}, the layer-wise trend emerges early in training. The resulting fraction of epochs required ranges from $7.9\%$ to $52\%$. In the same way, the percentage of model reduction after trimming varies substantially, ranging from $35.2\%$ to $94.2\%$. Interestingly, the relative change of IFC over training empirically correlates with the extractor–classifier transition, and correctly captures the splitting point on the benchmarks studied here. Indeed, Table~\ref{tab:models_split_analysis} in Appendix \ref{app:split_analysis} shows that the splitting point is always the same across datasets for MLP10 and MLP12, which share the same base structure. 
This pattern does not apply to ResNets or VGGs, where the layers vary in size and structure as a function of the network's depth. Nonetheless, Figure~\ref{fig:ifc_params_count} in Appendix~\ref{app:split_analysis} shows that models belonging to the same architectural family display similar patterns when the end-of-training IFC value is plotted against the percentage of the total parameters contained up to a layer. So, similar models tend to use the same fraction of total parameters to first map samples into a more expressive embedding space and subsequently compress them.

It is also worth noting that the initialization of network parameters and the optimization can influence the selected split layer. Indeed, even if the size and sampling of the tunnel set could, in principle, affect the identified layer, for sufficiently large datasets, even small (percentwise) subsets are representative. As shown in Appendix~\ref{app:tunnel_set}, varying the tunnel set size has little impact on the stopping epoch and essentially no effect on the identified split layer. Moreover, fixing the tunnel set does not eliminate the variability induced by different optimization seeds, further suggesting that the split layer is mainly determined by optimization dynamics. For a fixed tunnel set, varying the optimization can change the identified split layer by retaining or not an additional layer among the ones that are at the transition phase between extraction and classification (depicted in orange in Figure~\ref{fig:nns_pipeline}).

To investigate whether the split identified by \ours corresponds to a change in the representations of the tunnel set, we analyze, in Appendix~\ref{app:CKA}, the evolution of the Centered Kernel Alignment (CKA) through epochs.

\section{Performance analysis}
\label{sec:performance_analysis}

In this section, we analyze the performance of the proposed method. 
Additional complementary analyses and experiments are provided in Appendix~\ref{app:res}.

\begin{figure*}[t]
\centering

\includegraphics[width=1.0\columnwidth]{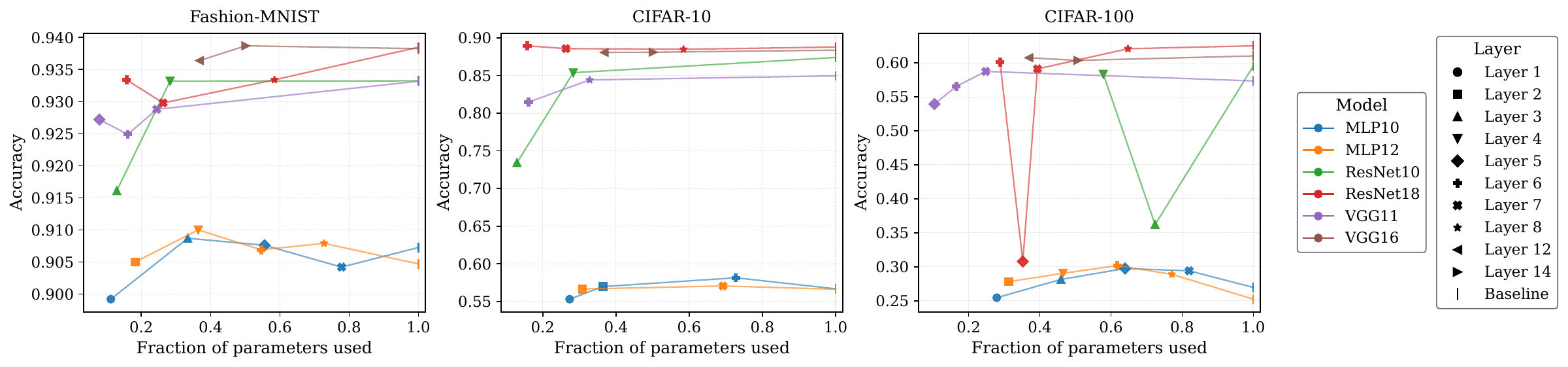}

\caption{Accuracy achieved by training from scratch different subnetworks obtained by splitting the full model at different layers. The x-axis reports the fraction of parameters in the resulting subnetwork, relative to the full model, when using linear probing.}
\label{fig:acc_params_count}
\end{figure*}

\subsection{Layer impact on generalization}

To analyze the impact of the split layer and of the resulting parameter count on generalization, we trained truncated models from scratch, starting from epoch~0, for several candidate split points. These experiments should be interpreted as an oracle analysis of split sensitivity rather than as the operational procedure used by \ours during training. In the oracle study, we evaluate the layer identified by \ours, its immediate neighborhood, and additional representative layers chosen to span a broad range of parameter budgets. Figure~\ref{fig:acc_params_count} shows the resulting performance as a function of the fraction of parameters retained relative to the full model. Importantly, the number of parameters does not necessarily increase monotonically with the depth of the split: in some cases, attaching a classification head to a larger intermediate representation can result in a higher total parameter count. This phenomenon appears, for example, in ResNet10 and ResNet18 on CIFAR-100, where certain splits perform significantly worse despite containing more parameters. In both cases, these splits correspond to earlier layers in the network, indicating that generalization depends more critically on the depth of the split than on the final parameter count. This observation is reinforced by the classification-head study in Table~\ref{tab:ablation_head} of Appendix~\ref{app:split_analysis}. There, heads that incorporate either global pooling or pixel-wise channel averaging before the final fully connected layer substantially reduce the number of parameters relative to simple linear probing while preserving performance. At the same time, once the split becomes too shallow, performance degrades even if the resulting head is larger.

\begin{figure*}[ht]
\centering
\begin{subfigure}{0.33\columnwidth}
  \centering
  \includegraphics[width=\columnwidth]{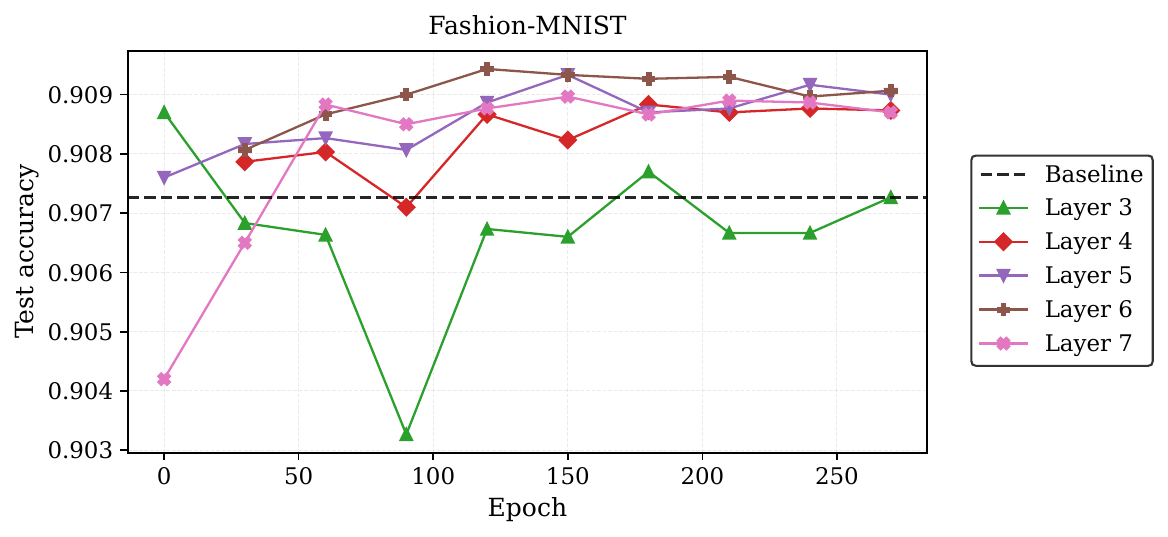}
  \caption{MLP10 Fashion-MNIST}
  \label{subfig:oracle_acc_epochs_mlp10_fashionmnist}
\end{subfigure}%
\begin{subfigure}{0.33\columnwidth}
  \centering
  \includegraphics[width=\columnwidth]{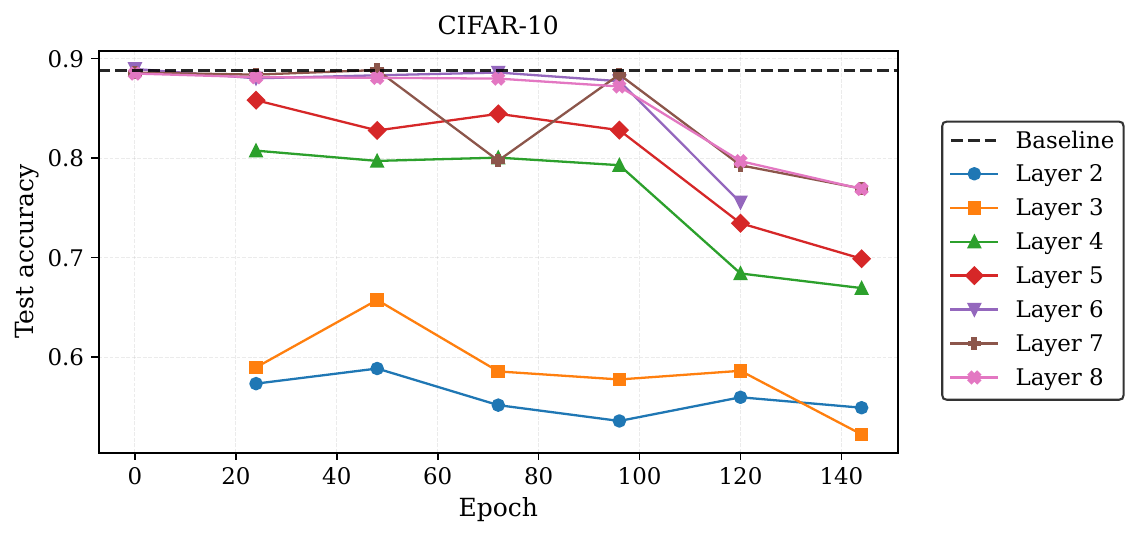}
  \caption{ResNet18 CIFAR-10}
  \label{subfig:oracle_acc_epochs_resnet18_cifar10}
\end{subfigure}
\begin{subfigure}{0.33\columnwidth}
  \centering
  \includegraphics[width=\columnwidth]{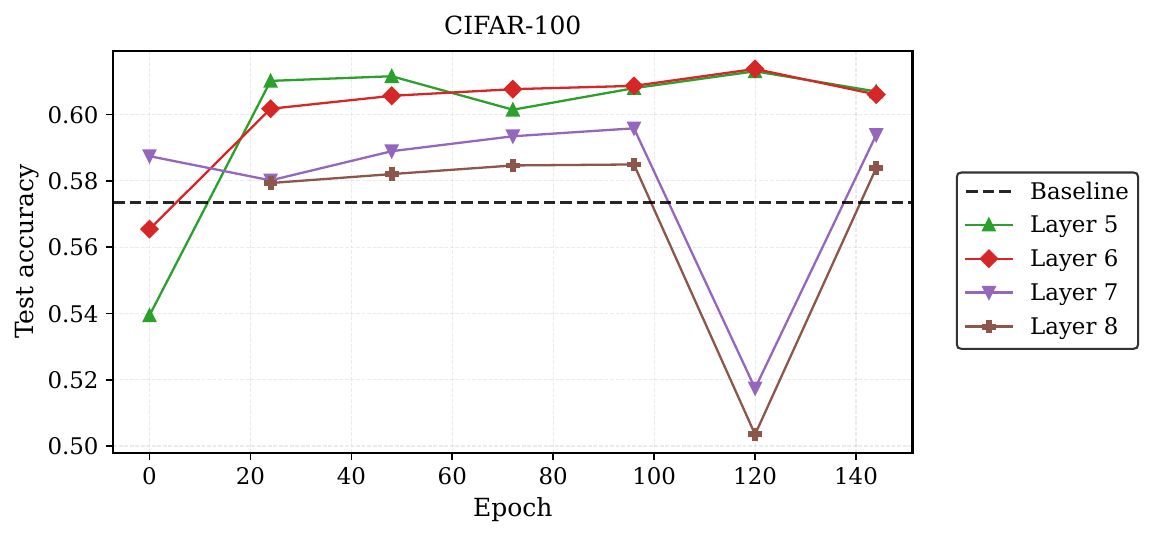}
  \caption{VGG11 CIFAR-100}
  \label{subfig:oracle_acc_epochs_vgg11_cifar100}
\end{subfigure}

\caption{Accuracy achieved by subnetworks obtained by splitting the model at different layers and epochs. The x-axis denotes the epoch at which the split is performed, after which the resulting smaller model is trained until the end.}
\label{fig:oracle_acc_epochs}
\end{figure*}

\subsection{Epoch impact on generalization}

The results in Figure~\ref{fig:acc_params_count} highlight the importance of the split point, while also showing that many layers, including the one identified by \ours, already achieve strong performance when the simplified model is trained from scratch. This suggests that the precise epoch at which the split is performed has a limited impact on the final outcome. At the same time, these oracle experiments do not eliminate the need for a criterion such as IFC: in practice, one still needs a principled way to decide which smaller architecture to instantiate, rather than selecting it a priori by trial and error. More generally, the oracle analyses in Figure~\ref{fig:acc_params_count}, Figure~\ref{fig:oracle_acc_epochs}, and Appendix~\ref{app:res} can also be viewed as controlled counterparts of fixed-epoch truncation and post-hoc split selection, since they explicitly evaluate different layers and different split times independently of the online stopping rule. In an ideal scenario, the split should therefore be applied as early as possible, in order to avoid training an overparameterized model for multiple epochs and to reduce the overall computational cost. So, the fast convergence of the IFC metric is particularly advantageous, as it allows the identification of a reliable split point after only a short initial training phase. This behavior is also evident in Figure~\ref{fig:oracle_acc_epochs}, which shows that, for a fixed splitting layer, the accuracy achieved by the truncated model remains stable across different starting epochs. More precisely, two patterns can be identified. First, for some layers, performing the split after a few initial epochs is beneficial, as it allows the layers that will be retained in the final model to be trained within the initial overparameterized network. This encourages a stronger focus on feature extraction, and it is clearly observed, for instance, in Figure~\ref{subfig:oracle_acc_epochs_vgg11_cifar100}. A similar, though milder, effect is visible in the MLP10 Fashion-MNIST setting, whereas it is less pronounced for ResNet18 on CIFAR-10. A second pattern is that the accuracy of the truncated model may deteriorate if the split is performed too late, as in Figure~\ref{subfig:oracle_acc_epochs_resnet18_cifar10}. This is expected, since the new classification head may not have sufficient remaining epochs to be adequately trained before the end of optimization.

\subsection{Comparison with full training}

To evaluate the performance of \ours, we compare linear probing ($\text{\ours}_{\text{LP}}$) with classification heads that enforce ETF structure in the final layer. Following \citealp{markou2024guiding}, we consider two variants: fixed ETF ($\text{\ours}_{\text{FIX}}$), in which the final classification layer is constrained to be a projector onto a canonical ETF, as in \cite{rangamani2023feature} work; and declarative ETF ($\text{\ours}_{\text{DCL}}$), in which the classifier is explicitly optimized toward the nearest ETF. In the \textsc{FIX} setting, the number of trainable parameters is reduced substantially because the classification head is not updated by gradient descent. By contrast, \textsc{DCL} relies on Riemannian optimization, which becomes expensive for high-dimensional embeddings and large numbers of classes. To mitigate this cost, we use a two-layer head with an initial projection to an intermediate-dimensional space and apply Riemannian optimization only to the final layer. 


For completeness, we compare \ours with the EB-LTH pruning method. Despite an apparent similarity, EB-LTH focuses on sparsifying a network through pruning in order to improve computational efficiency while maintaining the original depth, whereas \ours explicitly targets the construction of a shallower architecture. Although these approaches are not mutually exclusive and can be combined (as we did in Appendix \ref{app:acc_res}), a direct comparison is useful for highlighting their distinct perspectives. Following the original EB-LTH formulation, we evaluate pruning levels corresponding to 30\%, 50\%, and 70\% parameter reduction without re-initialization when the model is pruned.

Given the large number of experimental configurations explored, in this section we summarize the results using critical-difference diagrams, a widely used visualization technique for comparing multiple methods across several datasets. In the diagram, a lower average rank corresponds to stronger overall performance, and methods connected by a horizontal line are not significantly different according to the Nemenyi post-hoc test applied after Friedman’s test. When a method is not available for a specific configuration, as in the case of EB-LTH on MLPs, we assign it the same rank as full training for that configuration.

\begin{figure}[h]
\centering
\begin{tikzpicture}[
  treatment line/.style={rounded corners=1.5pt, line cap=round, shorten >=1pt, thick},
  treatment label/.style={font=\small, inner sep=2pt}, 
  group line/.style={ultra thick, gray!60},
]

\definecolor{CustomBlue}{RGB}{65, 105, 225}

\begin{axis}[
  clip={false},
  axis x line={center},
  axis y line={none},
  axis line style={-},
  xmin={1},
  xmax={7},
  ymax={0},
  ymin={-3.2}, 
  scale only axis={true},
  width={0.85\columnwidth}, 
  ticklabel style={anchor=south, yshift=1.1*\pgfkeysvalueof{/pgfplots/major tick length}, font=\small},
  every tick/.style={draw=black},
  major tick style={yshift=.5*\pgfkeysvalueof{/pgfplots/major tick length}},
  height={5\baselineskip}, 
  xtick={1,2,3,4,5,6,7},
  x dir={reverse},
]

\draw[color=CustomBlue, treatment line] ([yshift=-2pt] axis cs:1.86, 0) |- (axis cs:1.7, -0.9)
  node[treatment label, anchor=west, font=\small\bfseries] {$\text{NNS}_{\text{LP}}$};

\draw[treatment line] ([yshift=-2pt] axis cs:2.86, 0) |- (axis cs:1.7, -1.75)
  node[treatment label, anchor=west] {Full Training};

\draw[treatment line] ([yshift=-2pt] axis cs:3.52, 0) |- (axis cs:1.7, -2.6)
  node[treatment label, anchor=west] {$\text{EB-LTH}_{30}$};

\draw[treatment line] ([yshift=-2pt] axis cs:3.78, 0) |- (axis cs:7.0, -2.6)
  node[treatment label, anchor=east] {$\text{NNS}_{\text{DCL}}$};

\draw[treatment line] ([yshift=-2pt] axis cs:4.05, 0) |- (axis cs:7.0, -1.93)
  node[treatment label, anchor=east] {$\text{EB-LTH}_{50}$};

\draw[treatment line] ([yshift=-2pt] axis cs:5.0, 0) |- (axis cs:7.0, -1.266)
  node[treatment label, anchor=east] {$\text{EB-LTH}_{70}$};

\draw[treatment line] ([yshift=-2pt] axis cs:6.48, 0) |- (axis cs:7.0, -0.6)
  node[treatment label, anchor=east] {$\text{NNS}_{\text{FIX}}$};


\draw[group line] (axis cs:1.86,-0.5) -- (axis cs:3.78,-0.5);

\draw[group line] (axis cs:2.86,-1.50) -- (axis cs:4.05,-1.50);

\draw[group line] (axis cs:3.52,-1.0) -- (axis cs:5.00,-1.0);

\draw[group line] (axis cs:5.00,-0.3) -- (axis cs:6.48,-0.3);

\end{axis}
\end{tikzpicture}
\caption{Critical-difference diagram based on average ranks. Methods connected by a horizontal bar are not significantly different according to the Nemenyi post-hoc test. Lower rank is better.}
\label{fig:crit_diff_optimized}
\end{figure}
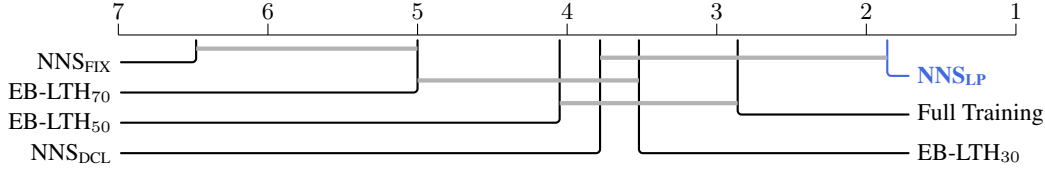

Figure~\ref{fig:crit_diff_optimized} summarizes the comparison across the 21 considered model--dataset configurations. Among all methods, $\text{\ours}_{\text{LP}}$ achieves the best average rank (1.86), followed by full training (2.86), while $\text{\ours}_{\text{FIX}}$ is consistently the weakest variant (6.48). The non-significant groupings are also informative: $\text{\ours}_{\text{LP}}$ belongs to a clique with full training, $\text{EB-LTH}_{30}$, and $\text{\ours}_{\text{DCL}}$, indicating that regardless of the rank differences these methods remain statistically competitive with the full training. At the opposite end, $\text{\ours}_{\text{FIX}}$ and $\text{EB-LTH}_{70}$ form the weakest statistically indistinguishable group, suggesting that, under such aggressive parameter reductions, neither approach remains fully competitive with full training. At the same time, the diagram shows a clear separation between the LP variant and the more unstable fixed-ETF configuration, which confirms that most of the benefit of \ours comes from the split strategy itself rather than from imposing a rigid classifier geometry after truncation. These results should also be read together with the parameter reductions reported in Table~\ref{tab:models_split_analysis}. Depending on the architecture and dataset, the split identified by \ours removes between 35.2\% and 94.2\% of the original parameters, with especially aggressive reductions on VGG models and several ResNet10 configurations. Therefore, the competitive average rank of $\text{\ours}_{\text{LP}}$ is achieved not by matching the full model size, but by preserving accuracy after a substantial reduction in network capacity. Overall, these results empirically confirm our hypothesis that \ours can achieve performance comparable to full training while dynamically reducing the number of parameters needed.

The complete set of tables with all accuracies, the FLOPs comparison, and a first preliminary study of the combination between \ours and EB-LTH is provided in Appendix~\ref{app:acc_res}.

\section{Conclusion}
\label{sec:conclusion}

In this work, we introduced \ours, a training-time framework that bridges the Tunnel Effect perspective with the Neural Collapse phenomenon to simplify overparameterized deep neural networks. By monitoring representation dynamics across layers and epochs, \ours identifies both when and where a network can be safely truncated, replacing redundant layers with a lightweight head while preserving generalization performance. Central to our approach is the Inverse Fisher Criterion, a stable and comparatively efficient proxy for Neural Collapse that captures the transition from feature extraction to classification on the image-classification architectures studied in this work. Extensive empirical evaluations show that \ours consistently produces substantially shallower models, achieving parameter reductions of up to 94\%, with accuracy comparable to that of fully trained networks. These results highlight the value of training-dynamics-aware model design and suggest that Neural Collapse can serve as a practical signal for adaptive architecture simplification. At the same time, our experiments indicate that the method is most reliable when the network exhibits a clear and stable transition between extractor-like and classifier-like layers; characterizing this condition more formally, and understanding cases in which the IFC profile is noisy or only weakly separated, remains an important direction for future work. A second open question is to investigate the behavior of \ours outside the convolutional setting studied here, even if the interaction between NC-inspired signals and transformer-like training dynamics may require different metrics or different split criteria altogether. We believe that \ours opens new directions for more parameter-efficient learning systems, and provides a foundation for future work on adaptive architectures and training-time model modification.

\bibliography{mybib}
\bibliographystyle{plainnat}

\newpage
\appendix
\onecolumn

\section{Experimental Setup}
\label{app:exp_set}

\subsection{Datasets}

We report results on the following benchmark datasets. Together, they span increasing levels of difficulty and different class cardinalities, from 10 classes up to 200, and include the fine-grained and less balanced CUB benchmark.

\paragraph{Fashion-MNIST}
Fashion-MNIST \cite{xiao_fashion-mnist_2017} is a dataset of Zalando apparel images, consisting of $60{,}000$ training examples and $10{,}000$ test examples. Each sample is a $28 \times 28$ grayscale image associated with one of 10 clothing categories, including T-shirts, trousers, and sneakers. The dataset was introduced as a more challenging drop-in replacement for the original MNIST \cite{lecun2010mnist} handwritten digits dataset.

\paragraph{CIFAR-10}
CIFAR-10 \cite{krizhevsky_learning_2009} is a standard benchmark for image classification in computer vision. It comprises $60{,}000$ color images of size $32 \times 32$, evenly distributed across 10 classes. The dataset is split into $50{,}000$ training images and $10{,}000$ test images and includes diverse object categories such as animals and vehicles.

\paragraph{CIFAR-100}
CIFAR-100 \cite{krizhevsky_learning_2009} extends CIFAR-10 by increasing the number of classes to 100, each containing 600 images, for a total of $60{,}000$ color images with resolution $32 \times 32$. The dataset is divided into $50{,}000$ training images and $10{,}000$ test images. Compared to CIFAR-10, CIFAR-100 provides a finer-grained classification task, with classes organized into more detailed semantic categories, such as specific animal species, household objects, and plant types.

\paragraph{CUB-200-2011}
The CUB-200-2011 dataset \cite{wah_cub-200-2011_2011} is a benchmark for fine-grained visual classification, focusing on bird-species recognition. It contains $11{,}788$ images of size $224 \times 224$ spanning 200 bird categories, split into $5{,}994$ training and $5{,}794$ test images. Each image is annotated with class labels, bounding boxes, part locations, and attribute annotations. The dataset is particularly challenging because of its high intra-class variability and subtle inter-class differences; the annotations were curated and validated through multiple rounds of Amazon Mechanical Turk labeling.

\subsection{Architectures and Hyperparameters}

In this section, we describe the model architectures considered in our experiments and summarize the corresponding architectural choices and hyperparameter settings.

\paragraph{MLP}
We consider Multi-Layer Perceptrons (MLPs) \cite{bebis_feed-forward_1994} composed of fully connected layers with ReLU activations. Specifically, we evaluate MLPs with 10 and 12 layers. The architecture consists of an input layer that flattens the input and projects it to a hidden representation of dimension 1024, followed by a sequence of hidden layers that preserve this dimensionality. All hidden layers use ReLU activations, while the output layer is a linear classifier without bias. No dropout or normalization techniques are employed in any layer.

\paragraph{ResNet}
We evaluate three variants of Residual Networks \cite{he_deep_2016}: ResNet-10, ResNet-18, and ResNet-34. All models follow a four-stage design based on residual blocks with identity shortcuts. The network begins with an initial convolutional layer followed by batch normalization and ReLU activation. For CUB-200-2011 this layer uses a $7 \times 7$ convolution with a stride of 2, whereas for the other dataset, a $3 \times 3$ kernel is employed. The subsequent stages consist of residual blocks with channel dimensions $64$, $128$, $256$, and $512$, respectively, with spatial downsampling performed via strided convolutions at the beginning of each stage except the first. Each residual block contains two $3 \times 3$ convolutional layers, each followed by batch normalization and ReLU activation. The network concludes with adaptive average pooling and a fully connected classification layer.

\paragraph{VGG}
We consider two Visual Geometry Group (VGG) architectures \cite{simonyan_very_2015}, namely VGG-11 and VGG-16. Both models are composed of five convolutional stages, where each stage consists of one or more $3 \times 3$ convolutional layers with ReLU activation, followed by max-pooling. Batch normalization is applied after each convolutional layer. The number of channels across successive stages is $64$, $128$, $256$, $512$, and $512$. The convolutional backbone is followed by a classifier composed of three fully connected layers: the first two have 4096 units with ReLU activation and dropout, while the final layer is a linear classifier mapping to the target number of classes.

\paragraph{Hyperparameters}

All models were trained with standard Stochastic Gradient Descent (SGD): 160 epochs for ResNet-10/18/34 and VGG-11/16, and 300 epochs for MLP-10/12. We use a 5-epoch warm-up followed by a step-decay schedule. For ResNet variants, the initial learning rate is 0.1, the weight decay is 0.0005, the momentum is 0.9, and the learning rate is decayed at epochs 60 and 120 with $\gamma=0.2$. VGG architectures use weight decay 0.0001 and decay the learning rate at epochs 80 and 120 with $\gamma=0.1$. MLPs are trained with base learning rate 0.05, no weight decay, no momentum, and learning-rate decays at epochs 100 and 200. Checkpoints are saved every 24 epochs for ResNets, every 6 epochs for VGGs, and every 30 epochs for MLPs. These settings follow standard image-classification practice and are kept fixed across experiments in order to evaluate the method across diverse model-dataset pairs, rather than to tune the split criterion itself.

\subsection{Machines}

Experiments were conducted on multiple Linux machines. The first system was equipped with an Intel Xeon Cascade Lake processor (4 CPU cores), 40 GB of RAM, and two NVIDIA Tesla T4 GPUs with 16 GB of memory each. The second system featured an AMD EPYC Rome processor (8 CPU cores), 62 GB of RAM, and a single NVIDIA A40 GPU with 46 GB of memory. The third was a cluster of four ARM machines, each equipped with an Ampere Altra Q80-30 CPU (80-core Arm Neoverse N1), 512 GB of memory, and two NVIDIA A100 GPUs with 40 GB of memory each.

Upon acceptance, we intend to release the complete experimental logs and aggregated results in order to facilitate reproducibility and support further work by the community.


\section{Neural Collapse and Representation Metrics}
\label{app:metrics}

In this appendix, $\mathbf{h}_i$ denotes the activation of the monitored layer after the layer transformation for all representation metrics. Whenever the layer output is not already a vector, the activation tensor is flattened per example before computing the class-wise and global statistics entering IFC. 

\subsection{Neural Collapse Definition}
\label{app:nc_definition}
Let $C$ be the number of classes and let $\mathbf{h}_i\in\mathbb{R}^d$ denote the last hidden layer activation of input $\mathbf{x}_i$.
Specifically, let $\mathbf{w}_k\in\mathbb{R}^d$ be the classifier weight vector for class $c\in\{1,\dots, C\}$, and $\mathcal{D}_c=\{\mathbf{x}_i\}_{i=1}^{N_c}$ the corresponding set of examples with mean $  \overline{\mathbf{h}}_c=\frac1{N_c}\sum_{x_i\in\mathcal{D}_c} \mathbf{h}_i$. Let $\overline{\mathbf{h}}=\frac1C\sum_{c=1}^C \overline{\mathbf{h}}_c$ be the global mean.
Formally, Neural Collapse is defined by the following four properties.

\noindent
NC1 - \textbf{Variability collapse:}  the individual class features collapse to their means
\[
\frac{\text{Tr}(\Sigma_W  \Sigma_B^\dagger )}{C}  \;\longrightarrow\; 0,
\]
Here, $\dagger$ is the pseudoinverse symbol, $\text{Tr}(\cdot)$ is the trace operator, and $\Sigma_W, \Sigma_B \in \mathbb{R}^{d \times d}$ are respectively the within-class and between-class covariance matrices:
\begin{equation}
\label{eq:variance_matrices}
   \begin{split}
       \Sigma_W &= \frac{1}{N} \sum_{c=1}^C \sum_{i=1}^{N_c} (\mathbf{h}_{i} -\overline{\mathbf{h}}_c)  (\mathbf{h}_{i}- \overline{\mathbf{h}}_c)^\top \\
       \Sigma_B &= \frac{1}{C}\sum_{c=1}^C (\overline{\mathbf{h}}_c - \overline{\mathbf{h}})  (\overline{\mathbf{h}}_c - \overline{\mathbf{h}})^\top
   \end{split}
\end{equation}
\medskip
\noindent
NC2 - \textbf{Convergence to the ETF:} the class means form a simplex ETF characterized by a maximum separation between each pair, and the same distance from the global one
\begin{equation*}
\begin{split}
| \| \overline{\mathbf{h}}_c - \overline{\mathbf{h}} \|_2  - \| \overline{\mathbf{h}}_{c'} - \overline{\mathbf{h}} \|_2 | &\quad \longrightarrow\quad 0 \qquad \qquad\: \quad\forall c,c'\\
\langle \widetilde{\mathbf{h}}_c, \widetilde{\mathbf{h}}_{c'} \rangle 
&\quad \longrightarrow\quad
\begin{cases}
1, & c=c',\\[4pt]
-\dfrac{1}{C-1}, &\forall c\neq c'.
\end{cases}
\end{split}
\end{equation*}
where $\widetilde{\mathbf{h}}_c = (\overline{\mathbf{h}}_c - \overline{\mathbf{h}}) / \| \overline{\mathbf{h}}_c - \overline{\mathbf{h}} \|_2$. 

\medskip
\noindent
NC3 - \textbf{Convergence to self-duality:} the weights of the penultimate layer converge to the same ETF structure of the embeddings
\[
\left \| \frac{W}{\|W||_F} - \frac{H}{\|H\|_F} \right\|_F \;\longrightarrow\; 0
\]
where $W\in \mathbb{R}^{d\times C}$ is the last layer weight matrix and $H= [\overline{\mathbf{h}}_c - \overline{\mathbf{h}}, \: c =1, \dots, C]\in \mathbb{R}^{d \times C}$ is obtained by stacking the centered class means as column vectors.

\medskip
\noindent
NC4 - \textbf{Simplification to nearest class center:} in this symmetric ETF structure, the nearest neighbour classification becomes the same as the maximisation of the logits
\[
\arg \max_c \:\: \mathbf{w}_c^\top \mathbf{h} \;\longrightarrow\; \arg \min_c \:\: \|\mathbf{h} - \overline{\mathbf{h}}_c\|_2. 
\]

We now present the candidate metrics considered in this work and analyze their behavior across architectures and datasets.

\subsection{Candidate representation metrics}
\label{app:metric_def}
Since our intuition is grounded in the Tunnel Effect (TE) phenomenon, it is worth examining it more closely. Although the split layer is defined in this work as the layer at which linear probing reaches at least 95\% of the final model accuracy, this layer often coincides with the point at which the Numerical Rank (NR) begins to decline. For calculating the NR for a single layer $\ell$, one computes the eigen-decomposition of the sample covariance matrix $\Sigma_l$ to estimate its rank by counting the number of eigenvalues exceeding a threshold $\tau$, set to $10^{-3}$ times the largest eigenvalue.  This quantity measures the effective dimensionality, or equivalently the degree of degeneracy, of the representation space at layer $\ell$. While this metric might represent a good candidate, it is computationally expensive. However, we will show the behavior of NR across various datasets and models in Appendix \ref{app:numerical_rank}.

An alternative way to identify the splitting point is to leverage metrics associated with the four NC properties, even if not all of these are tractable. In particular, evaluating NC4 would require training an auxiliary classifier after each network layer, which is computationally prohibitive. Similarly, NC3 cannot be monitored, as it measures the convergence of the final-layer weights to the class means in the Frobenius norm.

By contrast, NC2 admits an efficient computation by checking whether the centered class means attain maximal pairwise separation, making it a suitable candidate metric. Accordingly, we define and track the \textsc{COS} metric, which quantifies the average deviation between the angles formed by the empirical class means and those prescribed by an ETF:

\begin{equation}
     \text{COS} = \langle \widetilde{\mathbf{h}}_c, \widetilde{\mathbf{h}}_{c'} \rangle - \left(\frac{C}{C-1} \delta_{c,c'} - \frac{1}{C-1} \right)
 \;\longrightarrow\; 0 \qquad \qquad  \forall c,c'
 \end{equation}
Here $\delta_{c,c'}$ denotes the Kronecker delta, and this deviation measures the distance to the convergence regime implied by NC2.


Regarding NC1, even if it involves the pseudo-inverse of the between-class covariance matrix, which is computationally prohibitive at our scales, some approximations can be defined. 
A well-known proxy is the Neural Collapse Clustering (NCC) metric \cite{galanti_role_2022}, which captures tight intra-class clustering and strong inter-class separation in a pairwise manner:
\begin{equation}
   \text{NCC} = \sum_{c\neq c'} \frac{V_c + V_{c'}}{2 \|\overline{\mathbf{h}}_c - \overline{\mathbf{h}}_c'\|_2^2} 
\end{equation}
where, for each class $c$, 
\[
V_c = \sum_{i=1}^{N_c}\| \mathbf{h}_{i} - \overline{\mathbf{h}}_c\|_2^2
\]
denotes the within-class variance around the class mean $\overline{\mathbf{h}}_c$. Unlike NC1, NCC explicitly assesses class separability locally, at the level of individual class pairs; nevertheless, it still constitutes an anisotropic measure of discrimination.

As an additional metric, we also consider the Intrinsic Dimension (ID) of data representations in deep networks, estimated by \citealp{ansuini2019intrinsic} using the global ID estimator introduced by \citealp{facco_estimating_2017}. This approach relies on computing the ratio between the distances to the second and first nearest neighbors for each data point. Formally, let $\{\mathbf{x}_i\}_{i=1}^m$ denote $m$ points uniformly sampled from a manifold with intrinsic dimension $d$. Let $r_i^{(1)}$ and $r_i^{(2)}$ be the distances from point $i$ to its first and second nearest neighbors, respectively. It follows that $\forall i, \mu_i = r_i^{(2)}/r_i^{(1)}$ is distributed according to a Pareto distribution with density $f(\mu_i \mid d) = d \mu_i^{-(d+1)}$. Leveraging this, the likelihood of the vector $\boldsymbol{\mu} = (\mu_1, \dots, \mu_m)$ becomes $P(\boldsymbol{\mu} \mid d) = d^m \prod_{i=1}^m \mu_i^{-(d+1)}$, and ID can be computed by
\begin{equation}
\label{eq:intrinsic_dim}
   \qquad \text{ID} = \max_d \:\:P(\boldsymbol{\mu} \mid d).
\end{equation}

\paragraph{The Inverse Fisher Criterion} 

Finally, to track NC signals online and identify both the split point and the split epoch, we use the ratio between within-class and between-class variance, referred to here as the Inverse Fisher Criterion (IFC):
\begin{equation}
   \text{IFC}= \frac{\sigma_W}{\sigma_B} = \frac{\frac{1}{N} \sum_{c=1}^C \sum_{i=1}^{N_c} (\mathbf{h}_{i} -\overline{\mathbf{h}}_c)^\top  (\mathbf{h}_{i}- \overline{\mathbf{h}}_c)}{\frac{1}{C}\sum_{c=1}^C (\overline{\mathbf{h}}_c - \overline{\mathbf{h}})^\top  (\overline{\mathbf{h}}_c - \overline{\mathbf{h}})} = \frac{\text{Tr}(\Sigma_W)}{\text{Tr}(\Sigma_B)}\label{eq:ifc}
\end{equation}
since it is the inverse of the linear discriminant criterion defined in \citealp{fisher_use_1936} for a binary task. In comparison to the NC1, the IFC is an isotropic approximation in which all the directions in feature space contribute equally.
Its efficient computation, obtained by using the dot product instead of the outer product in equation \eqref{eq:variance_matrices}, comes at the cost of losing the information regarding the pairwise covariances in the original matrices and considering only the elements on the main diagonal. Note that when $C=2$, both IFC and NCC reduce to equivalent variance-separation ratios, differing only by a constant scaling factor. However, for $C>2$, the two criteria diverge, and NCC explicitly accounts for pairwise class separations, whereas IFC provides a global measure of class separability.

\subsection{Metrics behavior}
\label{app:metric_comparison}
Here, we provide a comprehensive analysis of the behavior of all considered metrics across all examined model-dataset combinations. Each figure corresponds to a single model, with each row representing a different dataset. The columns report the behavior of the previously defined metrics. Specifically, from left to right, we present: the convergence of the cosines between the class-mean embeddings and their ETF configuration (\textsc{COS}), the intrinsic dimension (ID) of the representation space, the Neural Collapse Clustering (NCC) metric as a proxy for NC1, and the proposed Inverse Fisher Criterion (IFC).

\begin{figure*}[h]
    \centering
    
    \begin{subfigure}{\textwidth}
        \centering
        \includegraphics[width=\textwidth]{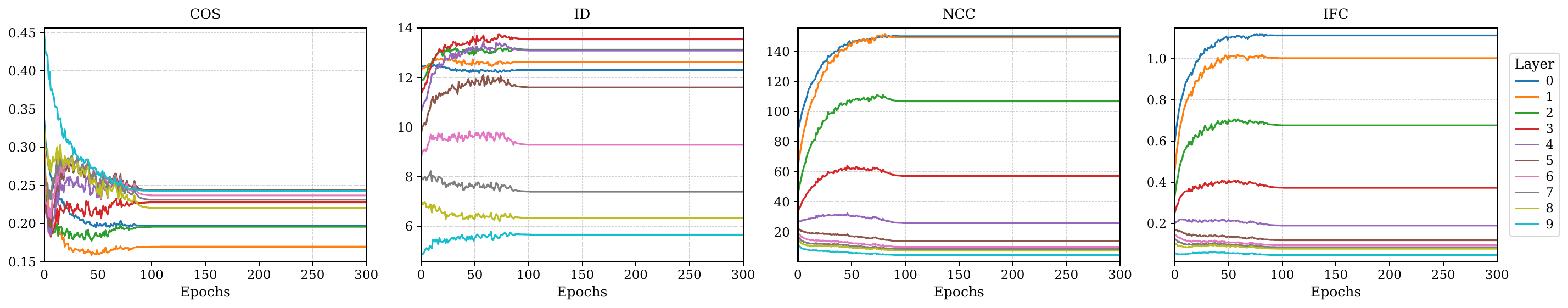}
        \caption{Behavior of different representation metrics across layers for a MLP10 trained on Fashion-MNIST.}
    \end{subfigure}


    \begin{subfigure}{\textwidth}
        \centering
        \includegraphics[width=\textwidth]{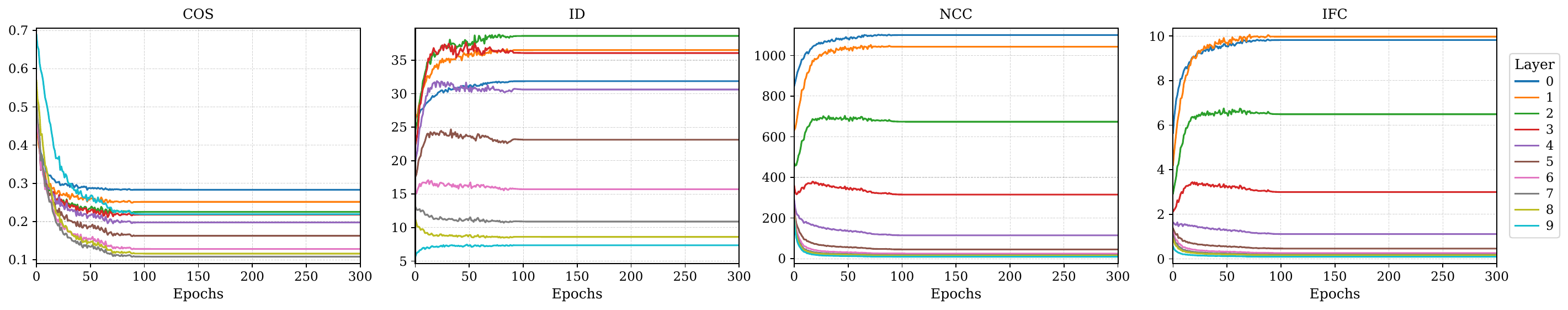} 
        \caption{Behavior of different representation metrics across layers for a MLP10 trained on CIFAR-10.}
    \end{subfigure}


    \begin{subfigure}{\textwidth}
        \centering
        \includegraphics[width=\textwidth]{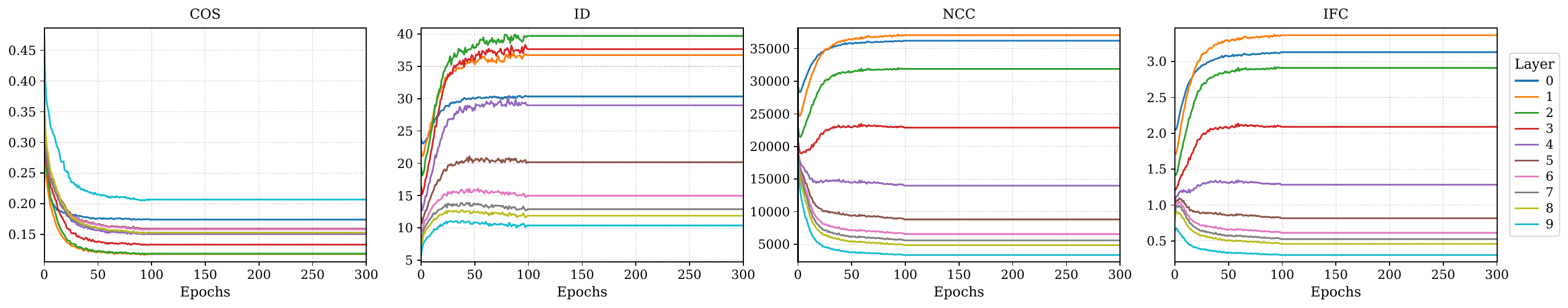} 
        
        \caption{Behavior of different representation metrics across layers for a MLP10 trained on CIFAR-100.}
    \end{subfigure}

    \caption{Candidate metrics of all MLP10 layers across different datasets.}
    \label{fig:mlp10_metrics}
\end{figure*}

\begin{figure*}[h]
    \centering
    
    \begin{subfigure}{1.0\textwidth}
        \centering
        \includegraphics[width=\textwidth]{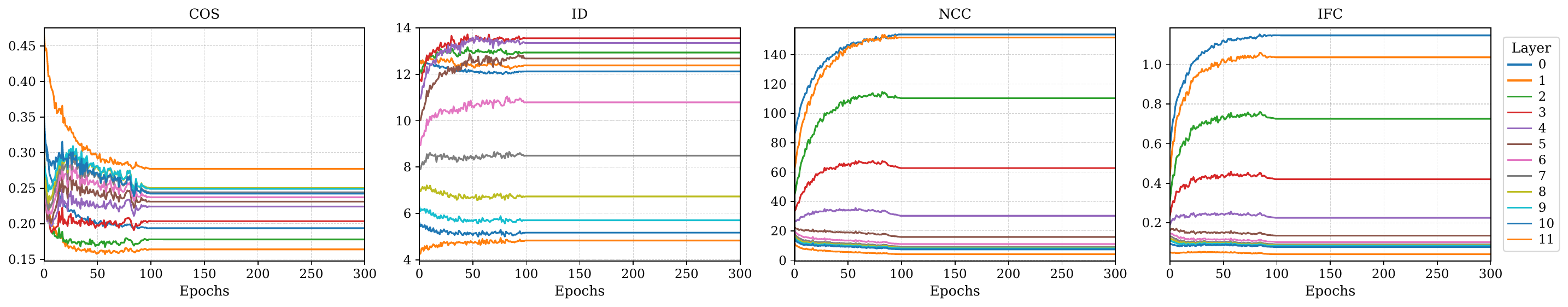}
        \caption{Behavior of different representation metrics across layers for a MLP12 trained on Fashion-MNIST.}
    \end{subfigure}


    \begin{subfigure}{1.0\textwidth}
        \centering
        \includegraphics[width=\textwidth]{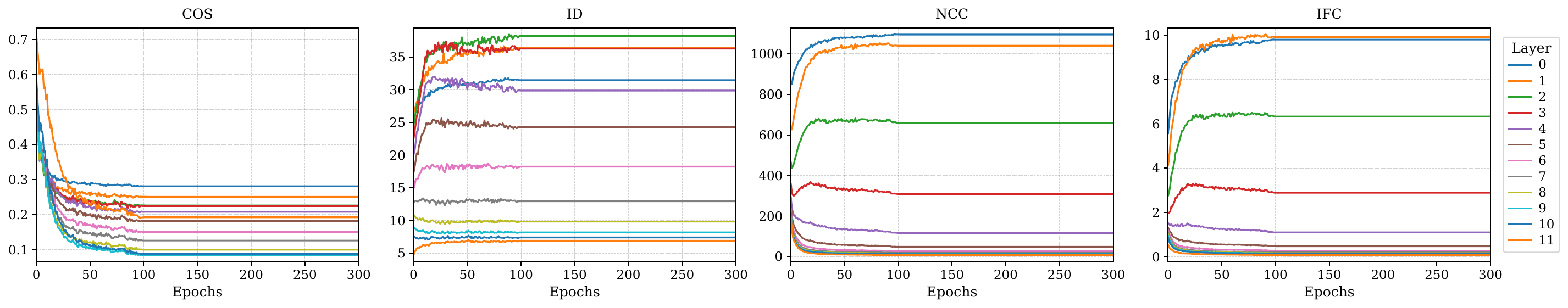} 
        \caption{Behavior of different representation metrics across layers for a MLP12 trained on CIFAR-10.}
    \end{subfigure}


    \begin{subfigure}{1.0\textwidth}
        \centering
        \includegraphics[width=\textwidth]{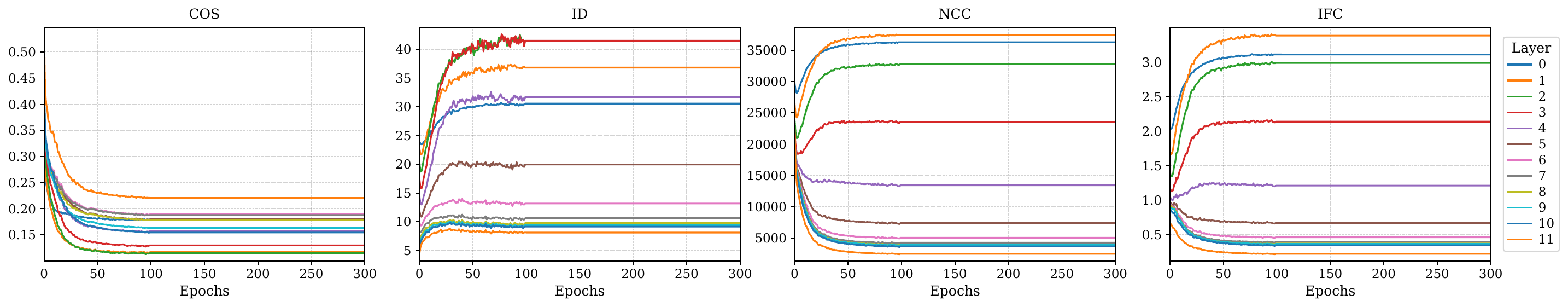} 
        
        \caption{Behavior of different representation metrics across layers for a MLP12 trained on CIFAR-100.}
    \end{subfigure}

    \caption{Candidate metrics of all MLP12 layers across different datasets.}
    \label{fig:mlp12_metrics}
\end{figure*}

\begin{figure*}[h]
    \centering
    
    \begin{subfigure}{1.0\textwidth}
        \centering
        \includegraphics[width=\textwidth]{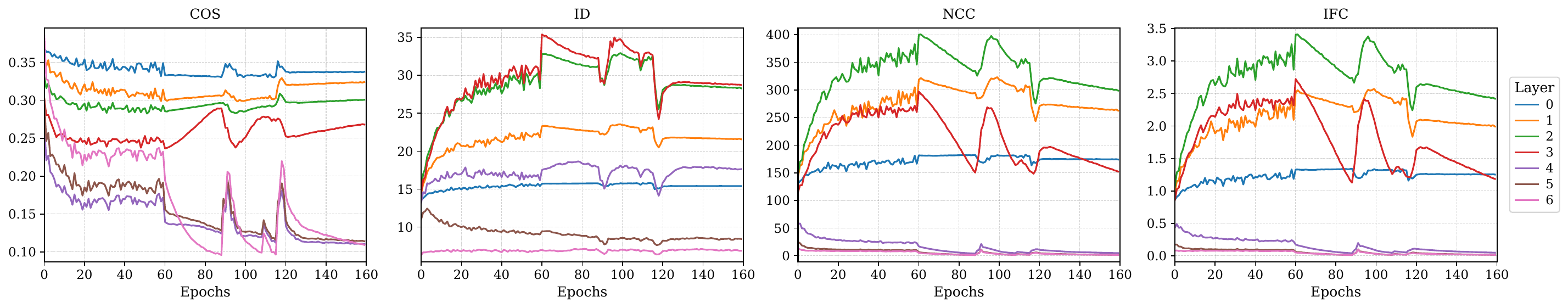}
        \caption{Behavior of different representation metrics across layers for a ResNet10 trained on Fashion-MNIST.}
    \end{subfigure}


    \begin{subfigure}{1.0\textwidth}
        \centering
        \includegraphics[width=\textwidth]{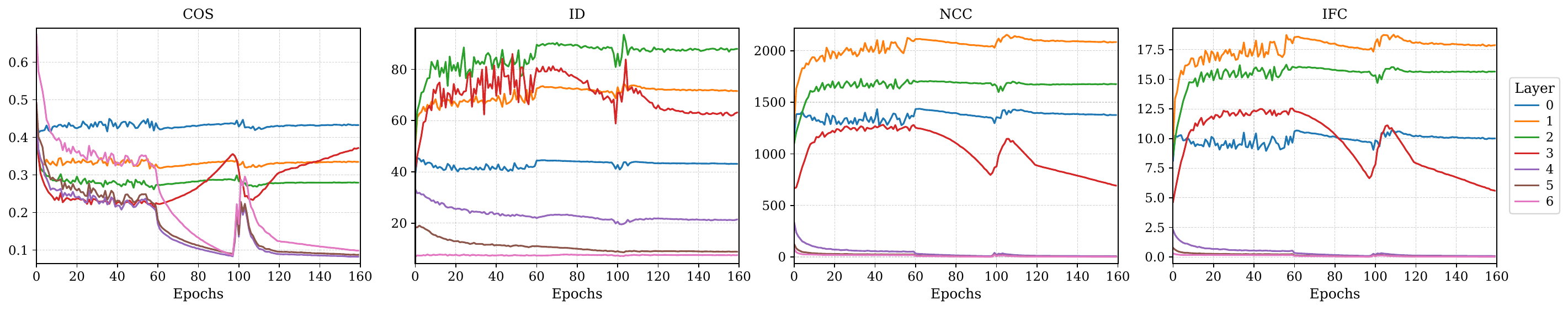} 
        \caption{Behavior of different representation metrics across layers for a ResNet10 trained on CIFAR-10.}
    \end{subfigure}


    \begin{subfigure}{1.0\textwidth}
        \centering
        \includegraphics[width=\textwidth]{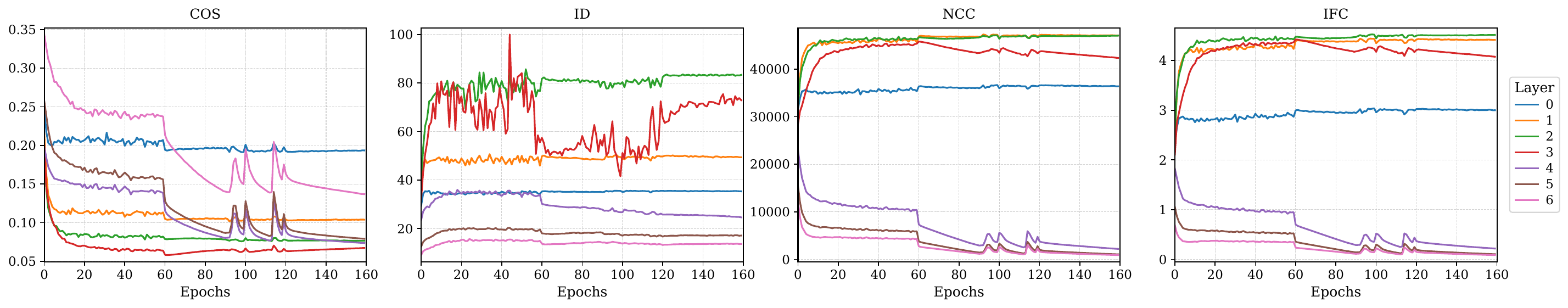} 
        
        \caption{Behavior of different representation metrics across layers for a ResNet10 trained on CIFAR-100.}
    \end{subfigure}

    \caption{Candidate metrics of all ResNet10 layers across different datasets.}
    \label{fig:resnet10_metrics}
\end{figure*}

\begin{figure*}[h]
    \centering
    
    \begin{subfigure}{1.0\textwidth}
        \centering
        \includegraphics[width=\textwidth]{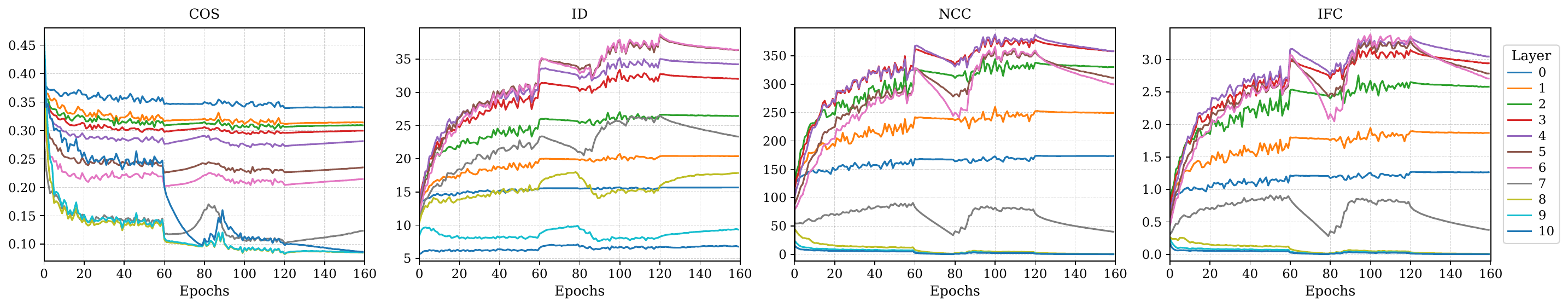}
        \caption{Behavior of different representation metrics across layers for a ResNet18 trained on Fashion-MNIST.}
    \end{subfigure}


    \begin{subfigure}{1.0\textwidth}
        \centering
        \includegraphics[width=\textwidth]{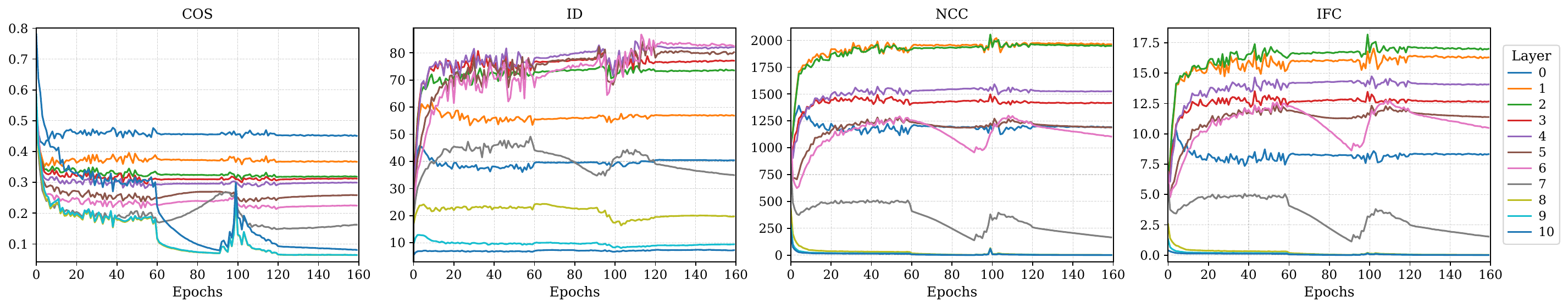} 
        \caption{Behavior of different representation metrics across layers for a ResNet18 trained on CIFAR-10.}
    \end{subfigure}


    \begin{subfigure}{1.0\textwidth}
        \centering
        \includegraphics[width=\textwidth]{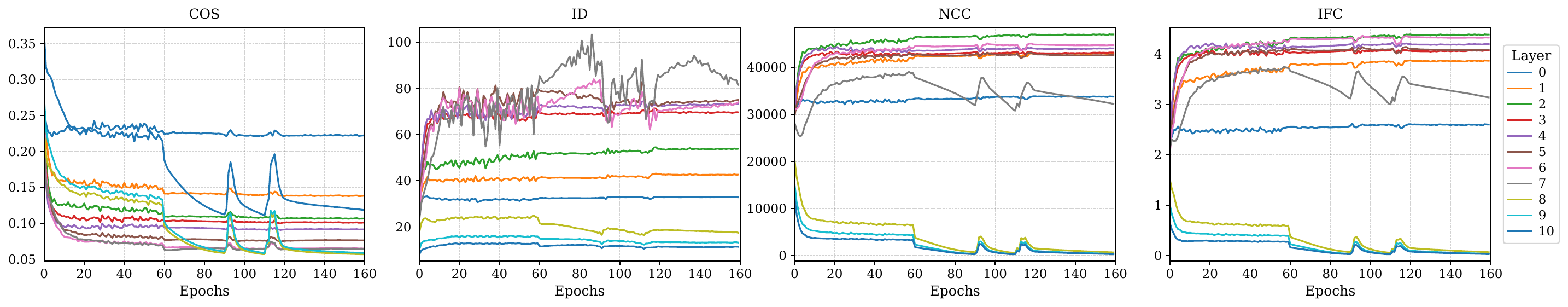} 
        
        \caption{Behavior of different representation metrics across layers for a ResNet18 trained on CIFAR-100.}
    \end{subfigure}

    \caption{Candidate metrics of all ResNet18 layers across different datasets.}
    \label{fig:resnet18_metrics}
\end{figure*}

\begin{figure*}[h]
    \centering
    
    \begin{subfigure}{1.0\textwidth}
        \centering
        \includegraphics[width=\textwidth]{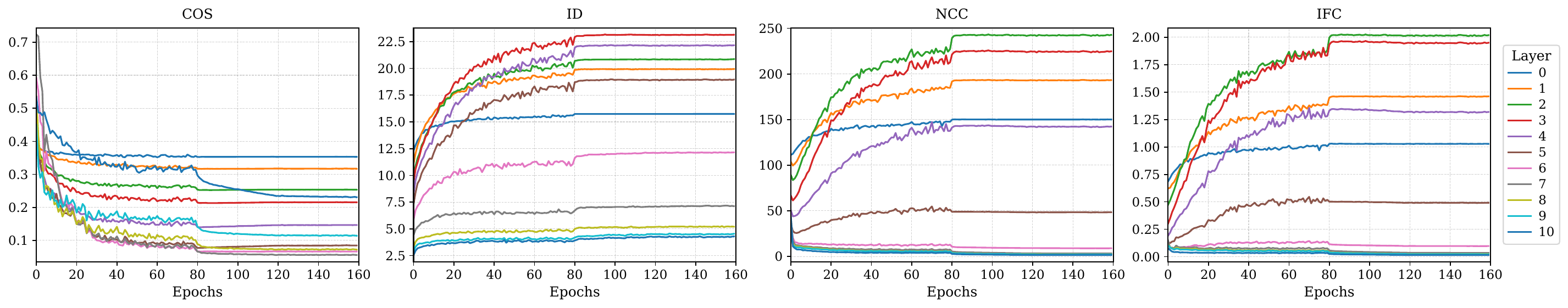}
        \caption{Behavior of different representation metrics across layers for a VGG11 trained on Fashion-MNIST.}
    \end{subfigure}


    \begin{subfigure}{1.0\textwidth}
        \centering
        \includegraphics[width=\textwidth]{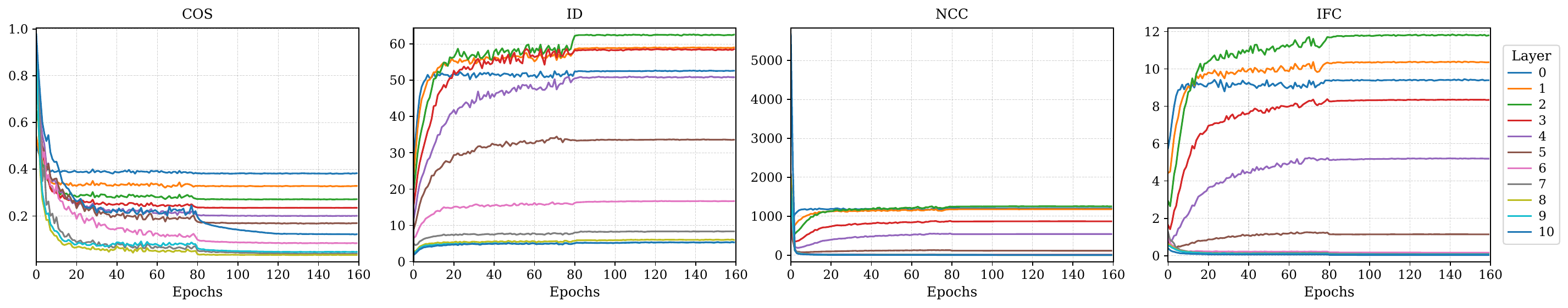} 
        \caption{Behavior of different representation metrics across layers for a VGG11 trained on CIFAR-10}
    \end{subfigure}


    \begin{subfigure}{1.0\textwidth}
        \centering
        \includegraphics[width=\textwidth]{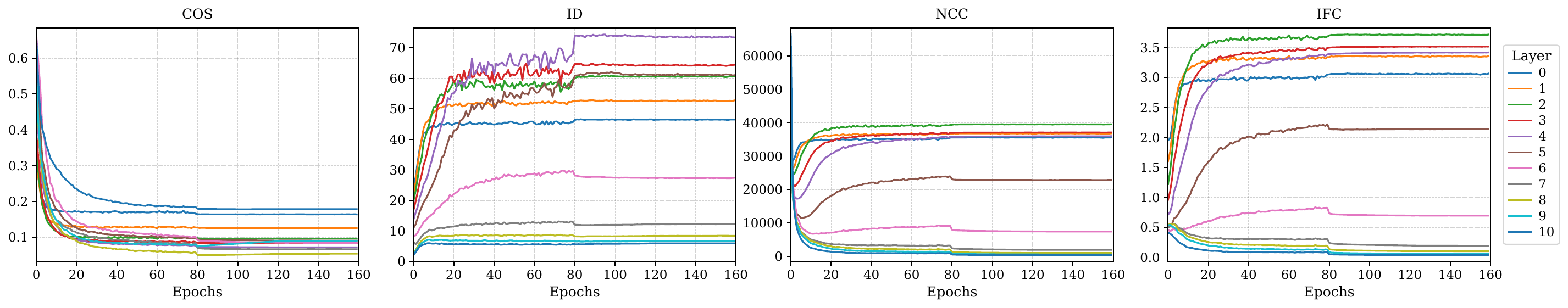} 
        
        \caption{Behavior of different representation metrics across layers for a VGG11 trained on CIFAR-100}
    \end{subfigure}

    \caption{Candidate metrics of all VGG11 layers across different datasets.}
    \label{fig:vgg11_metrics}
\end{figure*}

\begin{figure*}[ht]
    \centering
    
    \begin{subfigure}{1.0\textwidth}
        \centering
        \includegraphics[width=\textwidth]{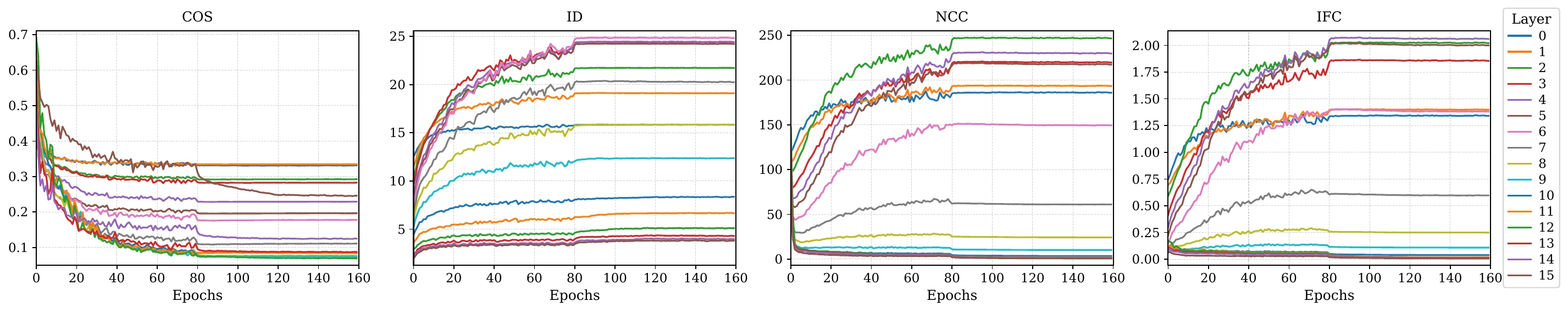}
        \caption{Behavior of different representation metrics across layers for a VGG16 trained on Fashion-MNIST.}
    \end{subfigure}


    \begin{subfigure}{1.0\textwidth}
        \centering
        \includegraphics[width=\textwidth]{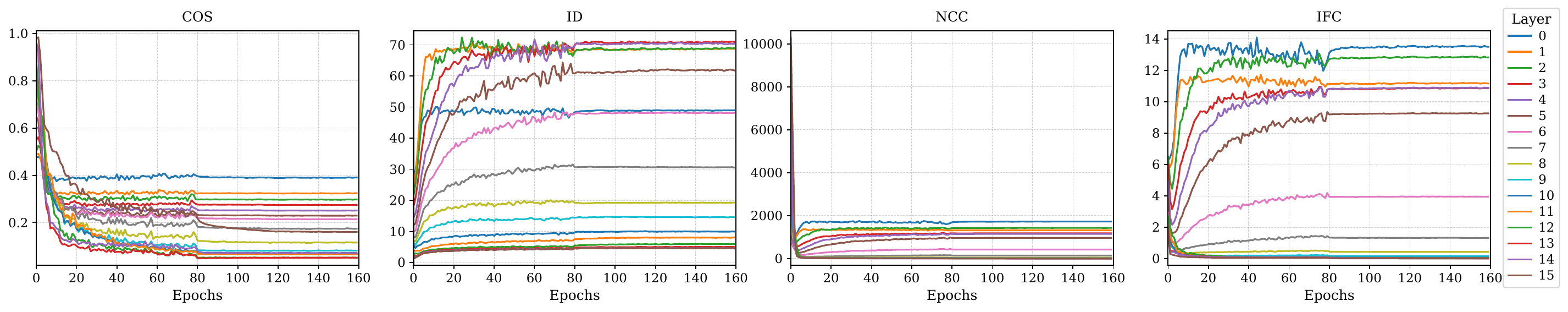} 
        \caption{Behavior of different representation metrics across layers for a VGG16 trained on CIFAR-10.}
    \end{subfigure}


    \begin{subfigure}{1.0\textwidth}
        \centering
        \includegraphics[width=\textwidth]{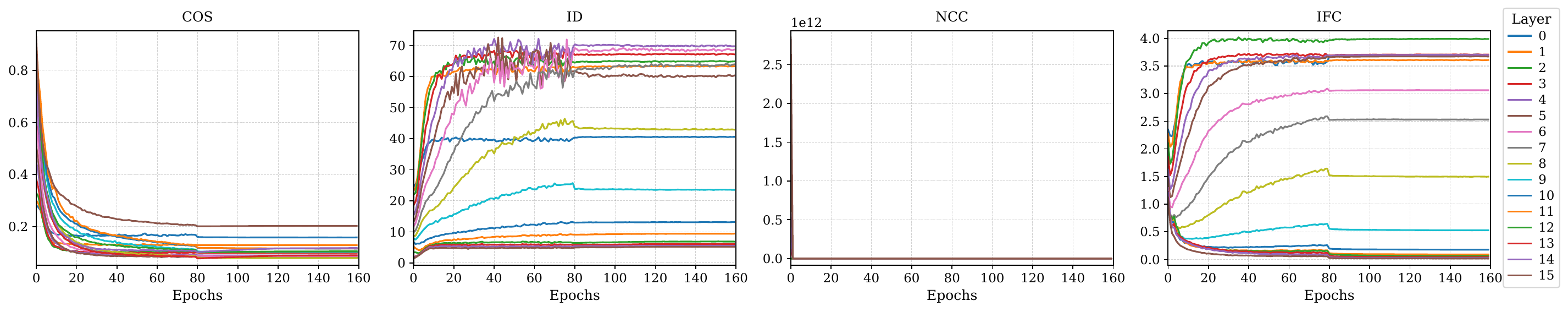} 
        
        \caption{Behavior of different representation metrics across layers for a VGG16 trained on CIFAR-100.}
    \end{subfigure}

    \caption{Candidate metrics of all VGG16 layers across different datasets.}
    \label{fig:vgg16_metrics}
\end{figure*}

As shown in Figures~\ref{fig:mlp10_metrics}, \ref{fig:mlp12_metrics}, \ref{fig:resnet10_metrics}, \ref{fig:resnet18_metrics}, \ref{fig:vgg11_metrics}, \ref{fig:vgg16_metrics}, a common trend across the considered metrics is that, as training progresses, the value associated with each layer converges toward its ground-truth level, despite the noise in the early stages of training. An effective metric should therefore display distinct behaviors across different categories of layers early in training, allowing the identification of an appropriate split point without requiring extensive optimization. However, neither COS nor ID exhibits a clear or consistent pattern. Across datasets and architectures, the evolution of these metrics varies substantially
making it difficult to meaningfully partition the network layers into two distinct groups.

In contrast, both NCC and IFC display a consistent pattern: early layers tend to increase the initial value of the metric, whereas later layers tend to decrease it. The behavior of intermediate layers is more ambiguous, as their values may oscillate around the initial level. Although the two metrics exhibit similar trends, IFC is regarded in the literature as a stable criterion for NC1 \cite{hui_limitations_2022} and is also computationally less expensive; we therefore adopt IFC as the selected metric. The lower stability of the NCC metric was also empirically observed in certain configurations, such as VGG16 on CIFAR-100 in Figure~\ref{fig:vgg16_metrics}, where the recorded values exhibit very large magnitudes, despite being averaged over multiple runs.

As illustrated in Figures~\ref{fig:mlp10_metrics}, \ref{fig:mlp12_metrics}, MLPs exhibit a clear and stable IFC pattern through the training epochs: early layers specialize in increasing class separability, while later layers tend to reduce it. This behavior is consistent across datasets and model depths, as is also observed for MLP10 and MLP12 over all the tested datasets. A similar trend occurs for the VGG architecture in Figures~\ref{fig:vgg11_metrics}, \ref{fig:vgg16_metrics}. In contrast, the analysis of ResNets, Figures \ref{fig:resnet10_metrics}, \ref{fig:resnet18_metrics}, is more involved, likely due to residual connections that make the split less granular. As a result, some layers that initially appear to act as feature extractors may be characterized as classifier layers as training progresses, a behavior that typically characterizes the intermediate layers. An exception is Layer~0, which sometimes exhibits behavior that is difficult to interpret, possibly due to its distinct functional role or to the limited number of parameters available at early stages to induce a meaningful transformation. Consequently, this layer is always treated as part of the feature extractor. 

\begin{figure*}[ht]
    \centering
    
    \begin{subfigure}{1.0\textwidth}
        \centering
        \includegraphics[width=\textwidth]{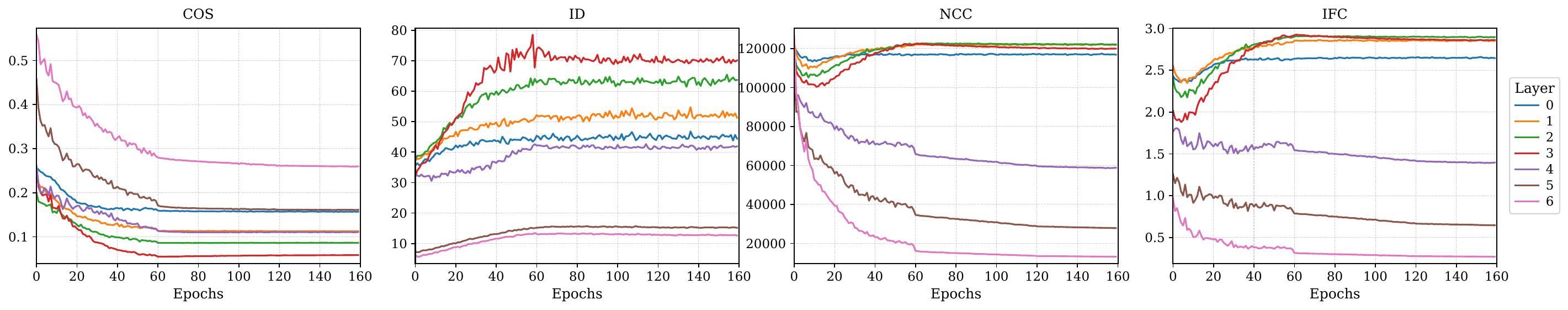} 
        
        \caption{Behavior of different representation metrics across layers for a ResNet10 trained on CUB.}
    \end{subfigure}


    \begin{subfigure}{1.0\textwidth}
        \centering
        \includegraphics[width=\textwidth]{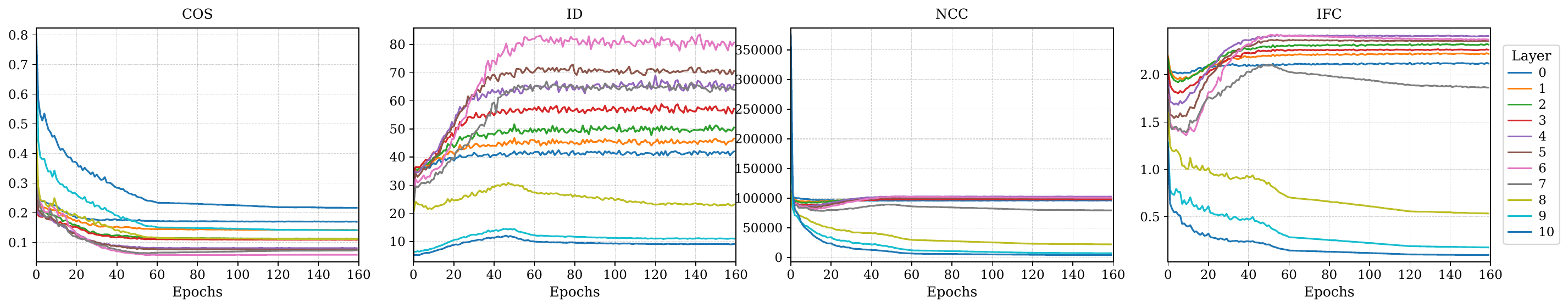} 
        
        \caption{Behavior of different representation metrics across layers for a ResNet18 trained on CUB.}
    \end{subfigure}


    \begin{subfigure}{1.0\textwidth}
        \centering
        \includegraphics[width=\textwidth]{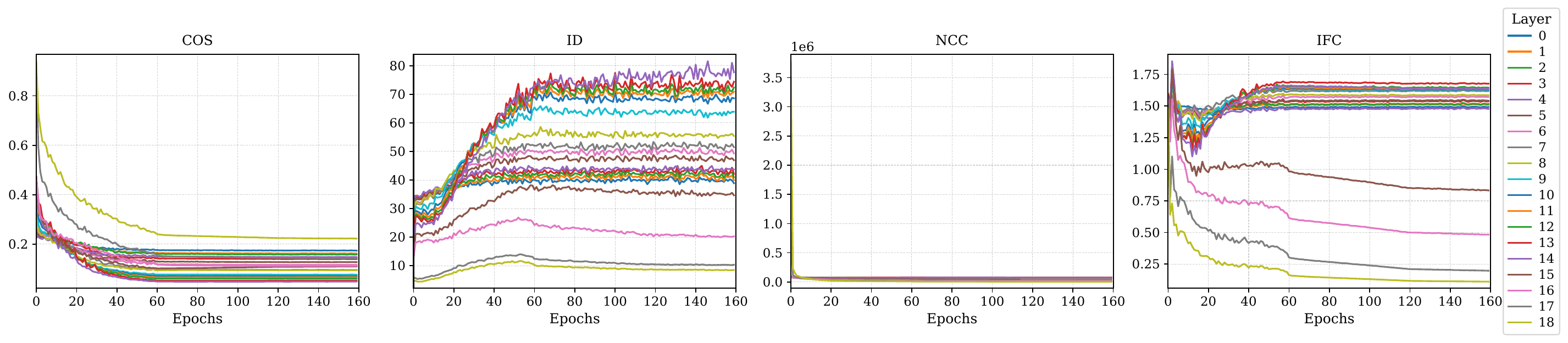} 
        
        \caption{Behavior of different representation metrics across layers for a ResNet34 trained on CUB.}
    \end{subfigure}

    \caption{Candidate metrics of all ResNets layers on the CUB dataset.}
    \label{fig:cub_metrics}
\end{figure*}

\clearpage

\subsection{Inverse Fisher Criterion analysis}
\label{app:IFC_analysis}
\begin{figure*}[h]
\centering

\includegraphics[width=\columnwidth]{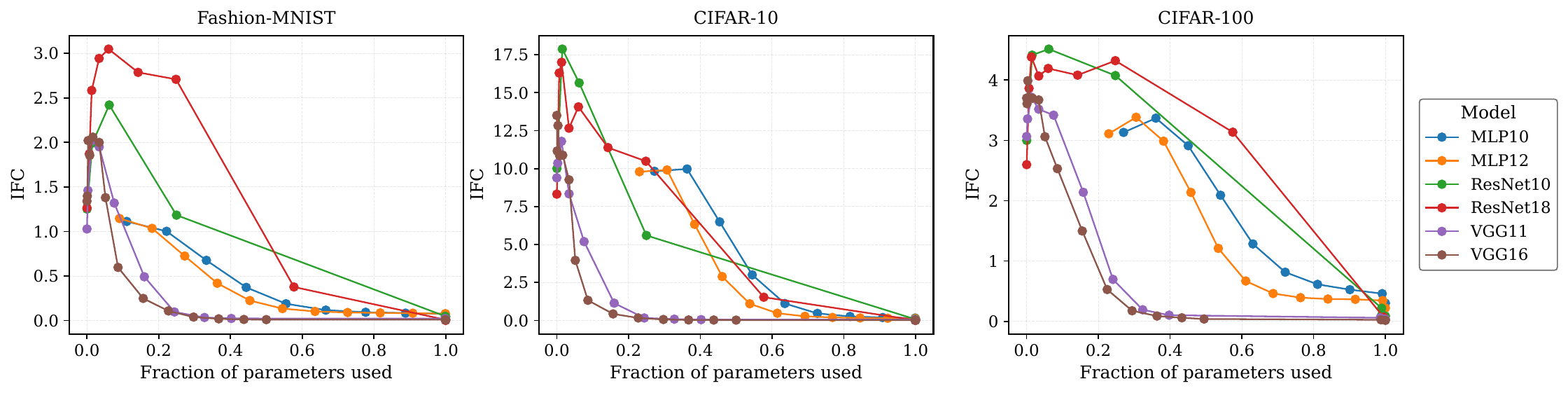}

\caption{The Inverse Fisher Criterion as a function of the fraction of parameters used across various datasets and models.}
\label{fig:ifc_params_count}
\end{figure*}

To investigate the presence of systematic patterns in the behavior of the IFC across models and datasets, Figure~\ref{fig:ifc_params_count} reports the IFC curves at the end of training for all considered models and datasets, excluding CUB. Since different architectures have different depths, the x-axis is normalized by reporting the fraction of the total number of parameters accounted for up to each layer, yielding a common range in $(0,1]$. Despite differences in absolute IFC values, the curves exhibit a clear clustering by architectural family, with models of the same type reaching their IFC maxima at approximately the same parameter fraction. A similar pattern is also observed for the locations of the IFC minima. This suggests that models of different depths but belonging to the same architectural family tend to exploit the initial layers in a similar manner to increase the IFC, thereby mapping samples into a representation space where classes are less entangled. The remaining layers, in contrast, markedly reduce variability, in line with the behaviors described by the Neural Collapse and Tunnel Effect phenomena.

\subsection{Numerical Rank analysis}
\label{app:numerical_rank}
Given the eigen-decomposition $\Sigma = Q\Lambda Q^{-1}$ of the sample covariance matrix $\Sigma \in \mathbb{R}^{d\times d}$, its Numerical Rank (NR) is estimated in \citealp{masarczyk2023tunnel} as
\begin{equation}
    \text{NR} =\sum_{i=1}^d \mathbf{1}[{\lambda_i > \tau}], \qquad s.t. \:\:\lambda_1 \leq \dots\leq\lambda_d
\end{equation}
where $\mathbf{1}[\cdot]$ is an indicator function and $\tau=1\mathrm{e}^{-3}\cdot \lambda_1$ is a threshold defined on the maximum eigenvalue $\lambda_1$.

\begin{figure*}[h]
    \centering
    
    \begin{subfigure}{\textwidth}
        \centering
        \includegraphics[width=\textwidth]{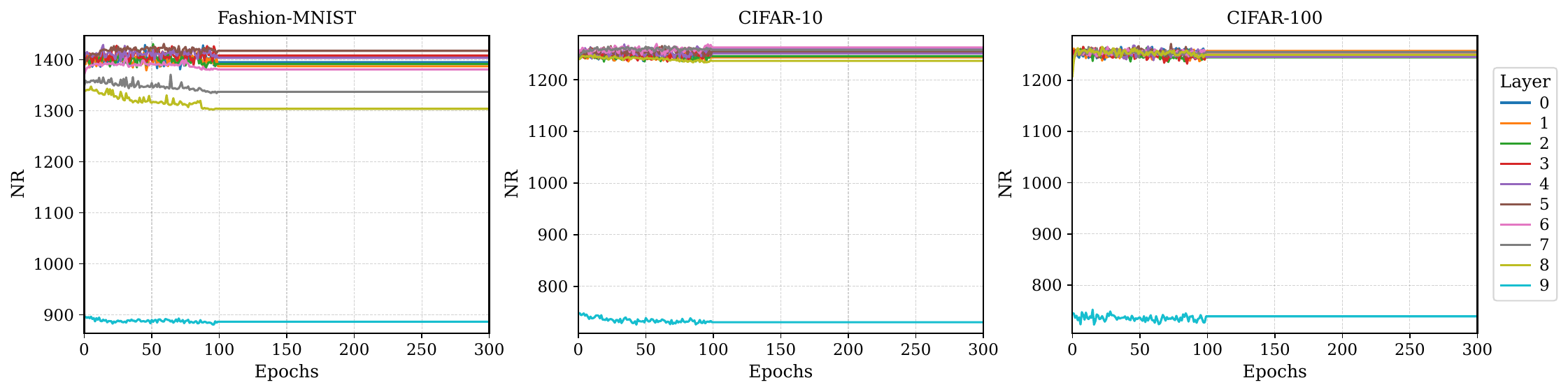}
        \caption{Behavior of the NR across layers for a MLP10 trained on different datasets.}
    \end{subfigure}

    \begin{subfigure}{\textwidth}
        \centering
        \includegraphics[width=\textwidth]{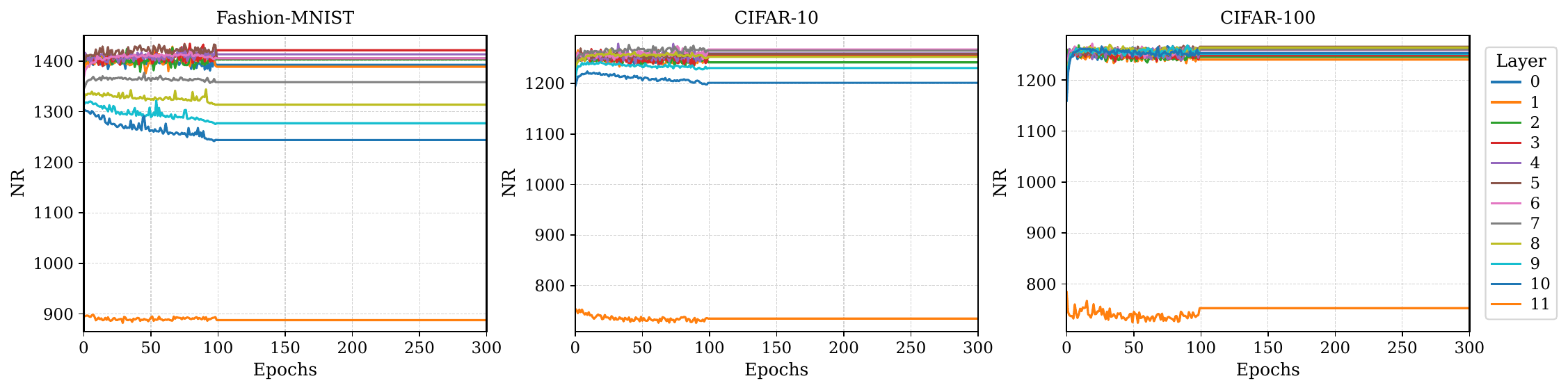}
        \caption{Behavior of the NR across layers for a MLP12 trained on different datasets.}
    \end{subfigure}
    
    \caption{Numerical rank of all MLPs layers across different datasets.}
    \label{fig:numerical_rank_mlp}
\end{figure*}

By analyzing the behavior of the NR, we observe a pattern that consistently holds across all model-dataset pairs. The final layer invariably exhibits the lowest numerical rank, while the earliest layers attain the highest values. However, for MLP architectures in Figure\ref{fig:numerical_rank_mlp}, no clear depth- or dataset-independent pattern emerges that would indicate a plausible split point. In many cases, all or nearly all layers appear to be informative, with the exception of the final layer. A partial exception is observed for Fashion-MNIST, where some of the last layers exhibit slightly lower NR values for both MLP10 and MLP12.

\begin{figure*}[h]
    \centering
    
     \begin{subfigure}{\textwidth}
        \centering
        \includegraphics[width=\textwidth]{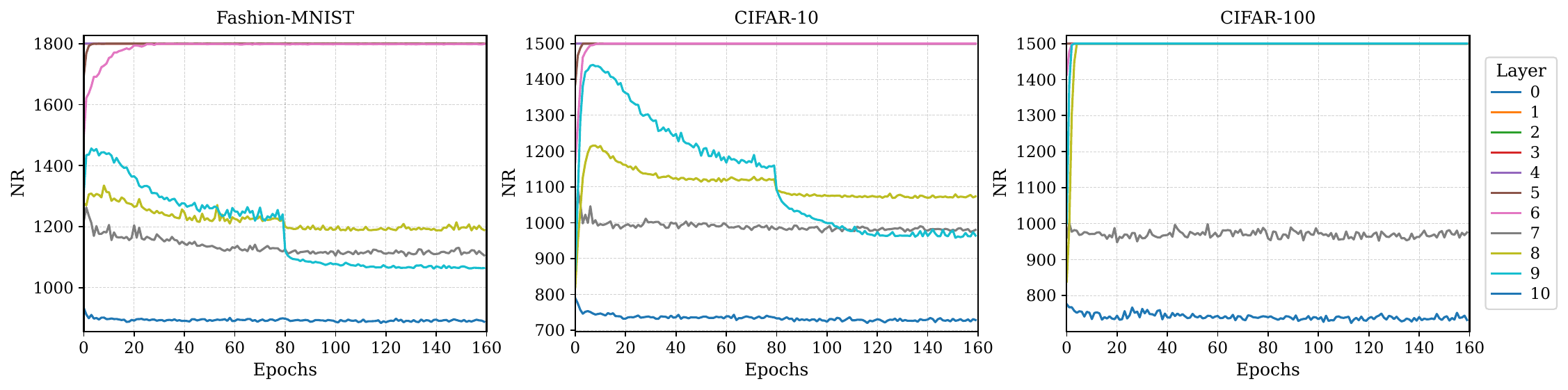} 
        
        \caption{Behavior of the NR across layers for a VGG11 trained on different datasets.}
    \end{subfigure}

    \begin{subfigure}{\textwidth}
        \centering
        \includegraphics[width=\textwidth]{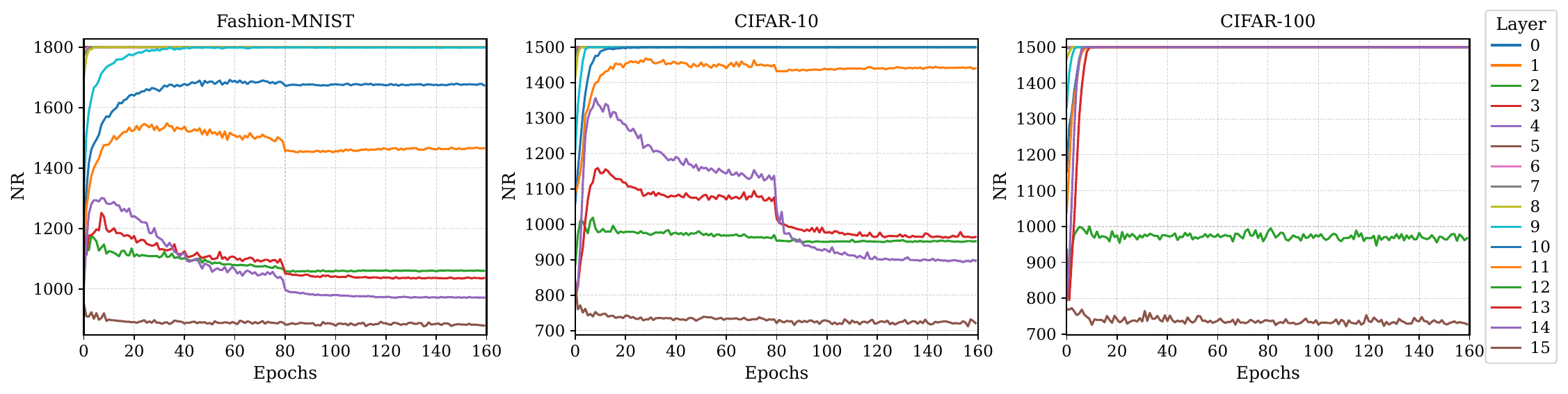} 
        
        \caption{Behavior of the NR across layers for a VGG16 trained on different datasets.}
    \end{subfigure}
    
    \caption{Numerical rank of all VGG layers across different datasets.}
    \label{fig:numerical_rank_vgg}
\end{figure*}

\begin{figure*}[h]
    \centering
    
    \begin{subfigure}{\textwidth}
        \centering
        \includegraphics[width=\textwidth]{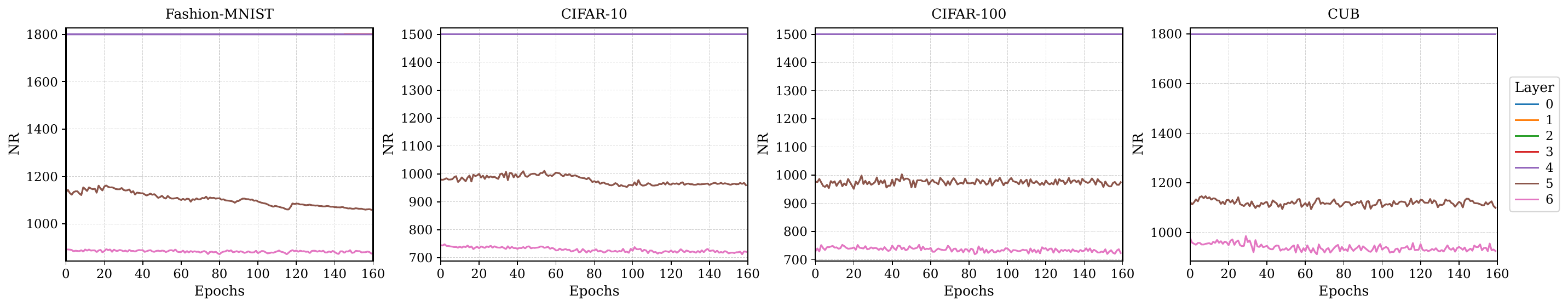} 
        
        \caption{Behavior of the NR across layers for a ResNet10 trained on different datasets.}
    \end{subfigure}
    \begin{subfigure}{\textwidth}
        \centering
        \includegraphics[width=\textwidth]{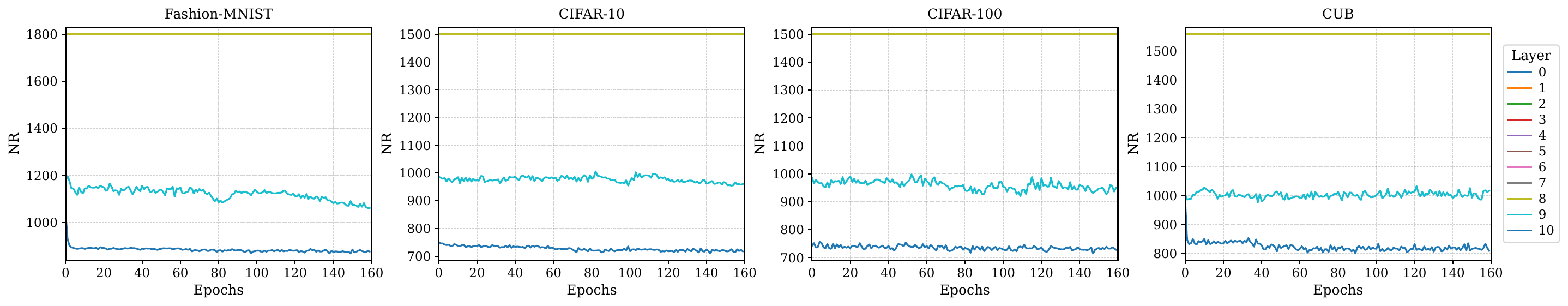} 
        
        \caption{Behavior of the NR across layers for a ResNet18 trained on different datasets.}
    \end{subfigure}
    \begin{subfigure}{\textwidth}
        \centering
        \includegraphics[width=0.3\textwidth]{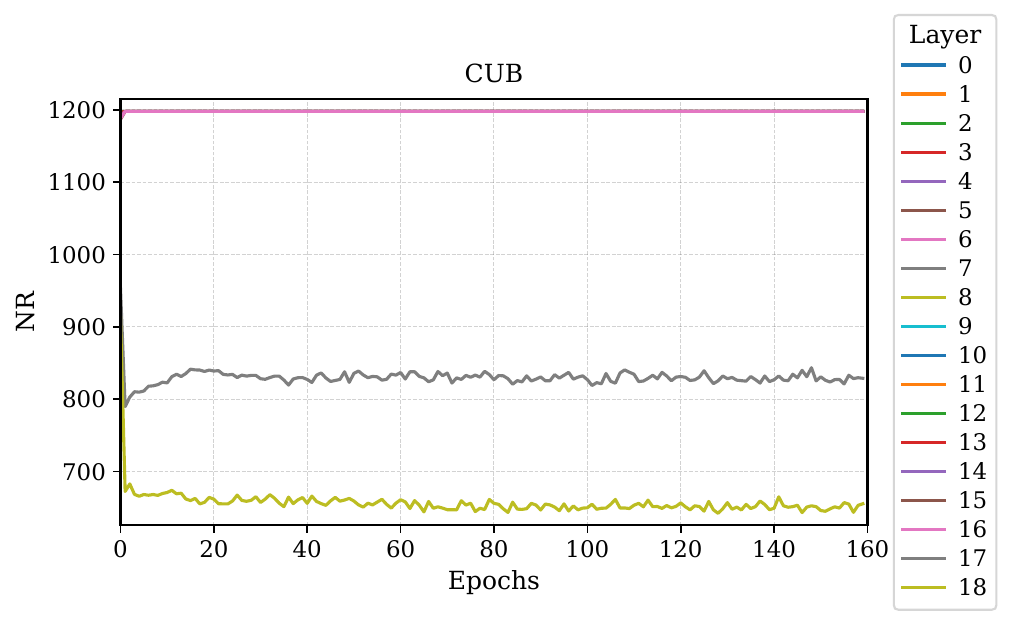} 
        
        \caption{Behavior of the NR across layers for a ResNet34 trained on CUB.}
    \end{subfigure}

    \caption{Numerical rank of all ResNets layers across different datasets.}
    \label{fig:numerical_rank_resnet}
\end{figure*}

In contrast, a clearer pattern emerges for ResNet architectures in Figure \ref{fig:numerical_rank_resnet} and becomes even more pronounced for VGG in Figure \ref{fig:numerical_rank_vgg} networks, where the penultimate layers display substantially lower numerical rank compared to the initial ones, suggesting a potential split point based on this behavior. Nevertheless, due to the high computational cost of this metric and its lack of consistency across all settings, we do not consider numerical rank as a suitable candidate for our proposed technique.

\clearpage


\clearpage
\section{\ours Split Analysis}
\label{app:split_analysis}

\begin{table}[h]
    \centering
    \caption{Stopping epoch, split layer, model reduction and fraction of total training across different models and datasets.}
    \begin{tabular}{ll r@{}l r@{}l c c}
    \toprule
    Model & Dataset & \multicolumn{2}{c}{Epoch} &
    \multicolumn{2}{c}{Layer} & \%Reduction & \%Training \\
    \midrule
    MLP10 & CIFAR-10 & 24.0 & \stdf{8.6} & 4.0 & \stdf{0.0} & 45.4\% & 8.0\% \\
    MLP10 & CIFAR-100 & 29.4 & \stdf{5.9} & 5.0 & \stdf{0.0} & 36.1\% & 9.8\% \\
    MLP10 & Fashion-MNIST & 57.2 & \stdf{38.0} & 4.6 & \stdf{0.6} & 48.8\% & 19.1\% \\
    \midrule
    MLP12 & CIFAR-10 & 23.8 & \stdf{8.1} & 4.2 & \stdf{0.5} & 52.3\% & 7.9\% \\
    MLP12 & CIFAR-100 & 26.6 & \stdf{3.3} & 5.0 & \stdf{0.0} & 45.8\% & 8.9\% \\
    MLP12 & Fashion-MNIST & 64.8 & \stdf{35.9} & 5.0 & \stdf{0.0} & 54.5\% & 21.6\% \\
    \midrule
    ResNet10 & CIFAR-10 & 16.0 & \stdf{0.0} & 4.0 & \stdf{0.0} & 71.7\% & 10.0\% \\
    ResNet10 & CIFAR-100 & 16.0 & \stdf{0.0} & 4.0 & \stdf{0.0} & 42.1\% & 10.0\% \\
    ResNet10 & Fashion-MNIST & 19.4 & \stdf{2.2} & 4.0 & \stdf{0.0} & 71.7\% & 12.1\% \\
    ResNet10 & Cub & 40.2 & \stdf{17.51} & 4.0 & \stdf{0.0} & 74.36\% & 25.0\% \\
    \midrule
    ResNet18 & CIFAR-10 & 37.2 & \stdf{22.3} & 7.4 & \stdf{0.6} & 60.8\% & 23.3\% \\
    ResNet18 & CIFAR-100 & 17.2 & \stdf{2.7} & 8.0 & \stdf{0.0} & 35.2\% & 10.8\% \\
    ResNet18 & Fashion-MNIST & 22.2 & \stdf{6.4} & 8.0 & \stdf{0.0} & 41.6\% & 13.9\% \\
    ResNet18 & Cub & 34.2 & \stdf{3.1} & 8.4 & \stdf{0.8} & 41.85\% & 21.3\% \\
    \midrule
    Resnet34 & Cub & 51.0 & \stdf{28.1} & 15.2 & \stdf{0.4} & 44.15\% & 31.8\% \\
    \midrule
    VGG11 & CIFAR-10 & 25.0 & \stdf{9.7} & 5.4 & \stdf{0.6} & 88.8\% & 15.6\% \\
    VGG11 & CIFAR-100 & 22.2 & \stdf{1.6} & 6.8 & \stdf{0.5} & 76.9\% & 13.9\% \\
    VGG11 & Fashion-MNIST & 29.4 & \stdf{11.5} & 6.8 & \stdf{0.8} & 77.2\% & 18.4\% \\
    \midrule
    VGG16 & CIFAR-10 & 39.2 & \stdf{7.0} & 7.0 & \stdf{0.8} & 94.2\% & 24.5\% \\
    VGG16 & CIFAR-100 & 32.0 & \stdf{7.8} & 8.4 & \stdf{0.9} & 85.8\% & 20.0\% \\
    VGG16 & Fashion-MNIST & 62.2 & \stdf{25.3} & 10.4 & \stdf{0.6} & 74.4\% & 38.9\% \\
    \bottomrule
    \end{tabular}
    \label{tab:models_split_analysis}
\end{table}

Table~\ref{tab:models_split_analysis} reports, for each tested model-dataset combination, the split layer and stopping epoch identified by \ours. For every configuration, we also report the resulting parameter reduction obtained by replacing the truncated network with a single fully connected classification head, as well as the fraction of training spent using the full model.

\subsection{Experiments on the tunnel set}
\label{app:tunnel_set}
In this section, we study how the identified split layer and stopping epoch vary with both the size and the construction of the tunnel set. We focus on a subset of the model-dataset pairs that exhibit high variability in Table~\ref{tab:models_split_analysis}, namely MLP10 on Fashion-MNIST and VGG16 on CIFAR-10.

\paragraph{Experiments on tunnel set dimension}

To analyze the dependence of layer and epoch identification on the size of the tunnel set, we vary the size of the selected tunnel set by sampling from the training data without replacement, proportionally to class frequencies, using sampling ratios of 1\%, 3\%, 5\%, and 10\%. For sampling rates above 10\%, the cost of storing embeddings for metric computation and the associated computational time become non-negligible, particularly for large-scale datasets such as CIFAR-10 and CIFAR-100.

\begin{table}[h]
    \centering
    \caption{Influence of the tunnel set size on \ours for the determination of the split layer and the concluding training epoch.}
    \begin{tabular}{ll c r@{}l r@{}l}
    \toprule
    Model & Dataset & \%Tunnel Set & \multicolumn{2}{c}{Epoch} & \multicolumn{2}{c}{Layer} \\
    \midrule
    MLP10 & Fashion-MNIST & 1\% & 47.75 & \stdf{29.77} & 4.5 & \stdf{0.58} \\
    MLP10 & Fashion-MNIST & 3\% & 35.00 & \stdf{17.07}  & 4.75 & \stdf{0.5} \\
    MLP10 & Fashion-MNIST & 5\% & 35.00 & \stdf{17.07} & 4.75 & \stdf{0.58}\\
    MLP10 & Fashion-MNIST & 10\% & 35.75 & \stdf{17.59} & 4.75 & \stdf{0.5}\\
    \midrule
    VGG16 & CIFAR-10 & 1\% & 31.50 & \stdf{8.10} & 6.5 &\stdf{0.96}\\
    VGG16 & CIFAR-10 & 3\% & 35.75 & \stdf{11.44}  & 6.5 &\stdf{0.58}\\
    VGG16 & CIFAR-10 & 5\% & 33.75 & \stdf{9.43}  & 6.5 &\stdf{0.58}\\
    VGG16 & CIFAR-10 & 10\% & 29.50 & \stdf{3.42} & 6.25 & \stdf{0.58}\\
    \bottomrule
    \end{tabular}
    \label{tab:tunnel_set_dimension}
\end{table}

As shown in Table~\ref{tab:tunnel_set_dimension}, the effect of tunnel-set size on the identified split layer is negligible: varying the subset from 1\% to 10\% of the training set produces virtually no change. Since, as discussed in the main paper, the selected layer matters much more than the exact split epoch, we fix the tunnel-set proportion to 5\% throughout both the main paper and the appendix.

\paragraph{Experiments on optimization and tunnel set creation}

To investigate whether the identified layers and epochs depend on the dataset construction procedure or on the optimization process, we try using one random seed for constructing the tunnel set and a different seed for optimizing the model, in order to assess the sensitivity of the results to these sources of randomness. Varying only the random seed used to generate the tunnel set allows us to assess whether the observed differences are solely attributable to the tunnel set sampling or instead arise from the underlying optimization process. As discussed in the main paper, the layer at which the split is identified is the most critical factor, whereas the specific epoch has a limited impact on the generalization performance of the simplified model. For this reason, our analysis primarily concentrates on the identified layer.

\begin{table}[h]
    \centering
    \caption{Influence of the tunnel and optimization seed on \ours for identifying the split layer and the terminating training epoch.}
    \begin{tabular}{ll c c c c}
    \toprule
    Model & Dataset & Tunnel Seed & Optimization Seed & Epoch & Layer \\
    \midrule
    MLP10 & Fashion-MNIST & 42 & 42 & 101 & 4\\
    MLP10 & Fashion-MNIST & 42 & 1024 &  21 & 5 \\
    MLP10 & Fashion-MNIST & 1024 & 42 & 91 & 4\\
    MLP10 & Fashion-MNIST & 1024 & 1024 & 23 & 5\\
    \midrule
    VGG16 & CIFAR-10 & 42 & 42 & 29 & 7\\
    VGG16 & CIFAR-10 & 42 & 1024 &  34 & 6\\
    VGG16 & CIFAR-10 & 1024 & 42 &  25 &  7\\
    VGG16 & CIFAR-10 & 1024 & 1024 & 23 &  6\\
    \bottomrule
    \end{tabular}
    \label{tab:tunnel_set_def}
\end{table}

As shown in Table~\ref{tab:tunnel_set_def}, the dominant factor in determining the split layer appears to be the optimization seed, and thus the network initialization and the subsequent optimization trajectory. However, it is difficult to conclusively disentangle how these factors are interrelated. Even if, in principle, both the size and the construction of the tunnel set should influence the identification of the split layer, as observed in Table~\ref{tab:tunnel_set_dimension}, for large datasets, even a small percentage is sufficient to obtain a subset that is reasonably representative of the training set. So, the optimization of the model appears to play a more significant role. Indeed, as emphasized in the literature~\cite{hui_limitations_2022, sukenik_neural_2024, han_flatness_2025}, Neural Collapse is primarily an optimization-driven phenomenon. This effect is clearly illustrated in the MLP10–Fashion-MNIST setting, where the optimization seed determines the selected split layer independently of the tunnel set construction. For a fixed tunnel set, changing the optimization seed is sufficient to alter the identified layer as reported in Table~\ref{tab:tunnel_set_def}. In nearly all the cases, this results in shifts of only one layer forward or backward, among the middle ones near the transition point. A closer inspection suggests that, in order to be slightly more robust and less sensitive to the very first IFC value, simple mitigations such as averaging the reference value over the first few epochs, or introducing a brief warm-up before fixing the baseline, may further reduce occasional split-point anomalies. These results should be interpreted as an initial robustness analysis rather than a complete one: they show that the criterion is reasonably stable to tunnel-set subsampling, but also that optimization effects matter. A systematic study of learning-rate schedules, weight decay, class imbalance, dataset scale, and other training choices remains outside the scope of the present work.

\subsection{Inverse Fisher Criterion convergence}

We report here the convergence rate of the Inverse Fisher Criterion (IFC) for each layer for all the model-dataset pairs toward its final value attained at the end of training. As shown, only a few epochs are sufficient for the metric to approach its asymptotic value. Moreover, if the objective is merely to determine whether a given layer tends to increase or decrease the IFC over training, this criterion enables early termination of the full-model training process.

\begin{figure*}[h]
    \centering
    
    \begin{subfigure}{\textwidth}
        \centering
        \includegraphics[width=\textwidth]{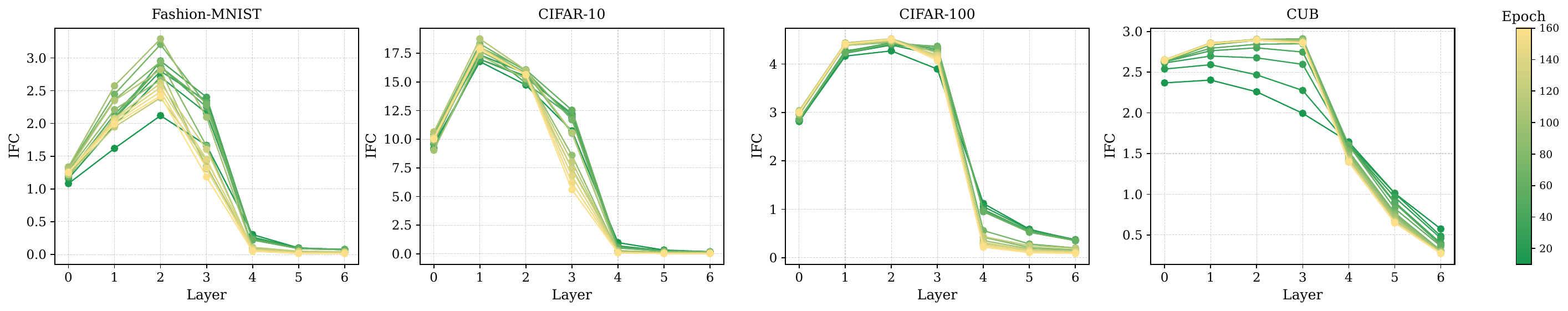}
        \caption{The x-axis corresponds to the network layers, ranging from the first to the final hidden layer of ResNet10, while the y-axis corresponds to the IFC metric. For a fixed layer, the points along the vertical line show the evolution of the IFC value across training epochs.}
    \end{subfigure}

    \vspace{0.3em} 

    \begin{subfigure}{\textwidth}
        \centering
        \includegraphics[width=\textwidth]{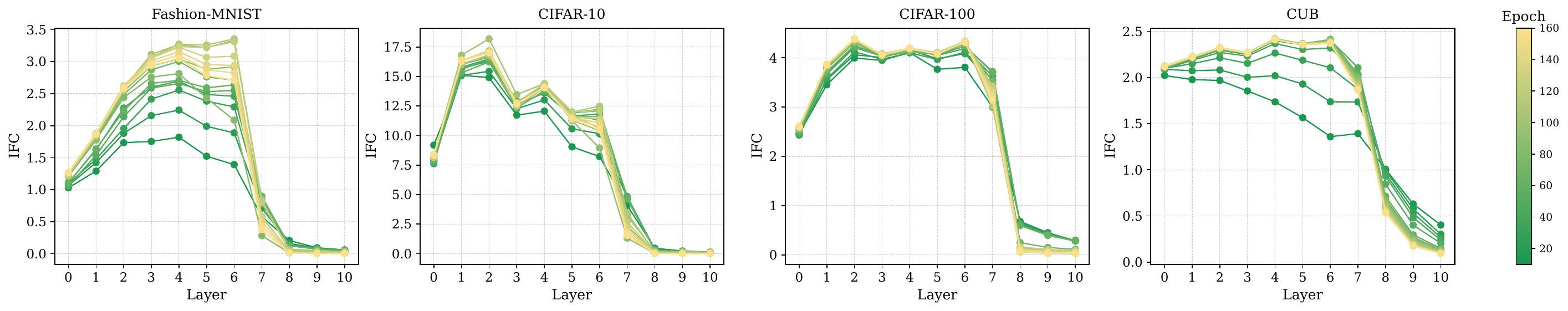}
        \caption{The x-axis corresponds to the network layers, ranging from the first to the final hidden layer of ResNet18, while the y-axis corresponds to the IFC metric. For a fixed layer, the points along the vertical line show the evolution of the IFC value across training epochs.}
    \end{subfigure}

    \vspace{0.3em} 

    \begin{subfigure}{\textwidth}
        \centering
        \includegraphics[width=0.3\textwidth]{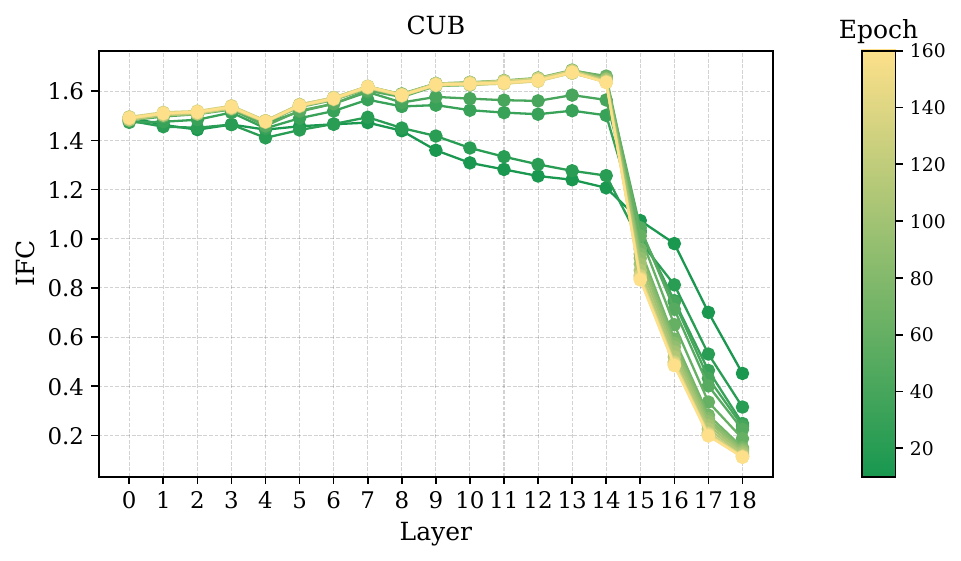}
        \caption{The x-axis corresponds to the network layers, ranging from the first to the final hidden layer of ResNet34, while the y-axis corresponds to the IFC metric. For a fixed layer, the points along the vertical line show the evolution of the IFC value across training epochs.}
    \end{subfigure}
    
    \caption{Convergence through epochs of the Inverse Fisher Criterion for all ResNets layers across different datasets.}
    \label{fig:ifc_convergence_resnet}
\end{figure*}
\begin{figure*}[h]
    \centering
    
    \begin{subfigure}{\textwidth}
        \centering
        \includegraphics[width=\textwidth]{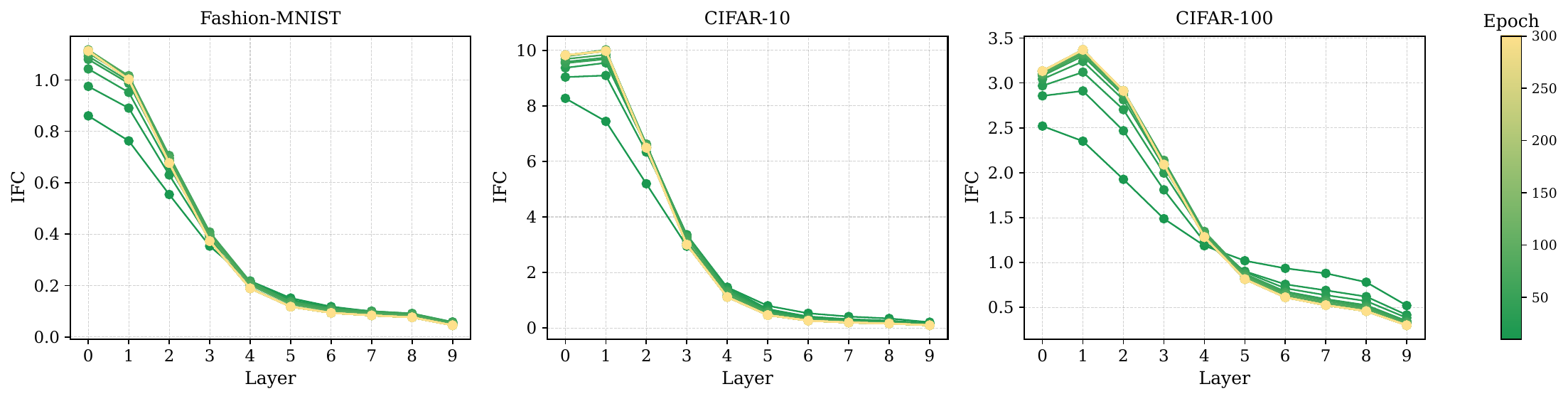}
        \caption{The x-axis corresponds to the network layers, ranging from the first to the final hidden layer of MLP10, while the y-axis corresponds to the IFC metric. For a fixed layer, the points along the vertical line show the evolution of the IFC value across training epochs.}
    \end{subfigure}

    \begin{subfigure}{\textwidth}
        \centering
        \includegraphics[width=\textwidth]{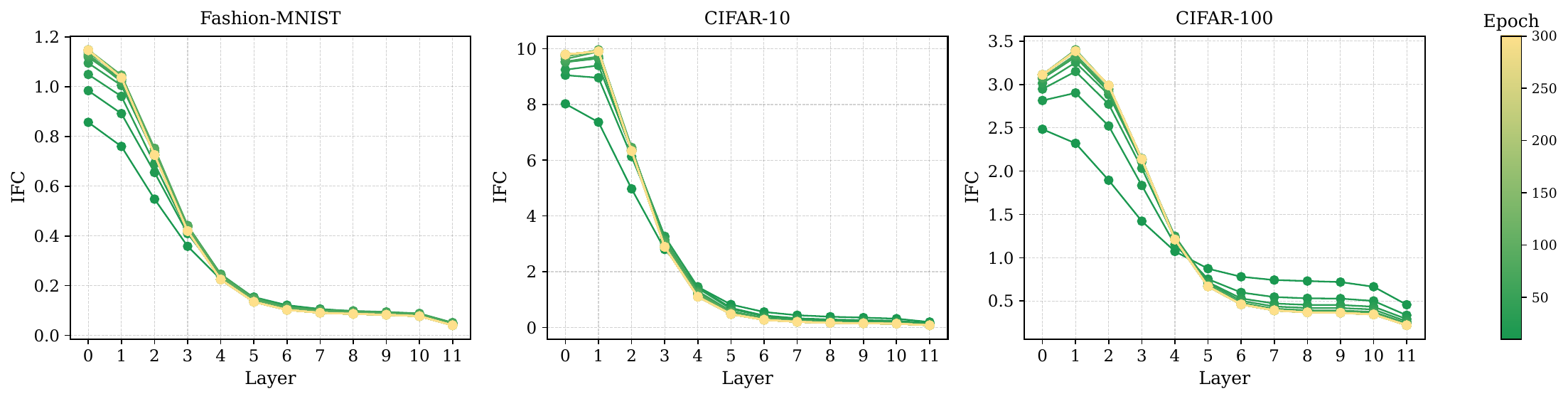}
        \caption{The x-axis corresponds to the network layers, ranging from the first to the final hidden layer of MLP12, while the y-axis corresponds to the IFC metric. For a fixed layer, the points along the vertical line show the evolution of the IFC value across training epochs.}
    \end{subfigure}
    
    \caption{Convergence through epochs of the Inverse Fisher Criterion for all MLPs layers across different datasets.}
    \label{fig:ifc_convergence_mlp}
\end{figure*}
\begin{figure*}[h]
    \centering
    
     \begin{subfigure}{\textwidth}
        \centering
        \includegraphics[width=\textwidth]{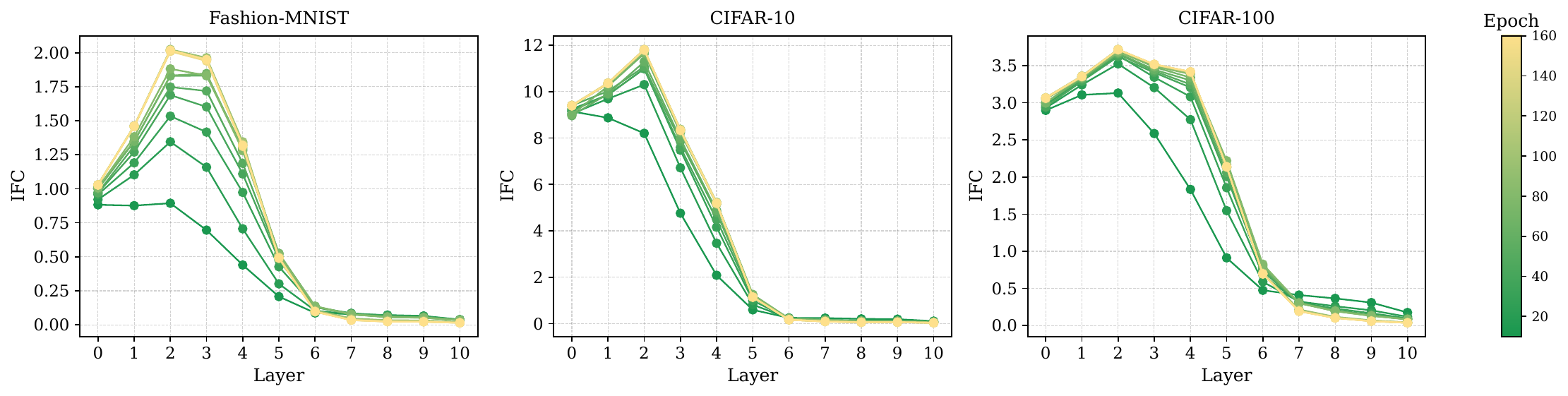} 
        
        \caption{The x-axis corresponds to the network layers, ranging from the first to the final hidden layer of VGG11, while the y-axis corresponds to the IFC metric. For a fixed layer, the points along the vertical line show the evolution of the IFC value across training epochs.}
    \end{subfigure}

    \begin{subfigure}{\textwidth}
        \centering
        \includegraphics[width=\textwidth]{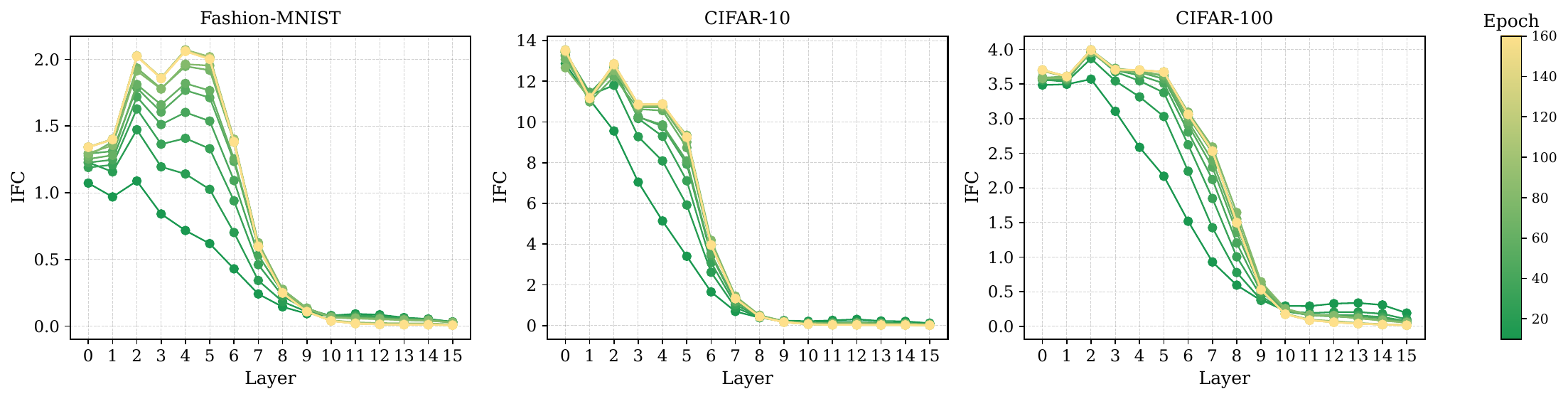} 
        
        \caption{The x-axis corresponds to the network layers, ranging from the first to the final hidden layer of VGG16, while the y-axis corresponds to the IFC metric. For a fixed layer, the points along the vertical line show the evolution of the IFC value across training epochs.}
    \end{subfigure}
    
    \caption{Convergence through epochs of the Inverse Fisher Criterion for all VGGs layers across different datasets.}
    \label{fig:ifc_vgg}
\end{figure*}

\clearpage

\subsection{Centered Kernel Alignment analysis}
\label{app:CKA}

The Centered Kernel Alignment (CKA)~\cite{cristianini_kernel-target_2001, cortes_algorithms_2012,kornblith_similarity_2019} measures the similarity between the representations learned at different layers of a neural network. Let $X \in \mathbb{R}^{m \times d_1}$ and $Y \in \mathbb{R}^{m \times d_2}$ denote the representations of two layers evaluated on the same set of $m$ examples, possibly with different feature dimensionalities. The matrices $K = XX^\top$ and $L = YY^\top$ encode pairwise similarities between examples induced by the representations in $X$ and $Y$, respectively. The Hilbert-Schmidt Independence Criterion (HSIC) quantifies the similarity between these centered similarity matrices, and we follow \citealp{song_feature_2012} by using an unbiased estimator:

\begin{equation*}
\begin{aligned}
\text{HSIC}_1(K,L)
&= \frac{1}{m(m-3)} \Big( \text{Tr}(\widetilde{K}\widetilde{L}) + \frac{\mathbf{1}^\top \widetilde{K} \mathbf{1}\mathbf{1}^\top \widetilde{L}\mathbf{1}}{(m-1)(m-2)} - \frac{2}{m-2}\mathbf{1}^\top \widetilde{K}\widetilde{L}\mathbf{1} \Big),
\end{aligned}
\end{equation*}

that is independent of the sample size. Here $\widetilde{K}$ and $\widetilde{L}$ are obtained by setting the diagonal entries of $K$ and $L$ to zero, $\mathbf{1} \in \mathbb{R}^m$ denotes the one vector, and $\text{Tr}(\cdot)$ is the trace operator. The resulting CKA metric is defined as
\begin{equation}
    \text{CKA}(K,L) = \frac{\text{HSIC}_1(K,L)}{\sqrt{\text{HSIC}_1(K,K)\,\text{HSIC}_1(L,L)}},
\end{equation}

This similarity measure allows us to identify groups of layers that behave similarly and, consequently, to detect transitions in the functional role of the network. In all the following figures, CKA is computed between each layer and its successor throughout training, for different models and datasets.

\begin{figure*}[h]
    \centering
    
    \begin{subfigure}{\textwidth}
        \centering
        \includegraphics[width=\textwidth]{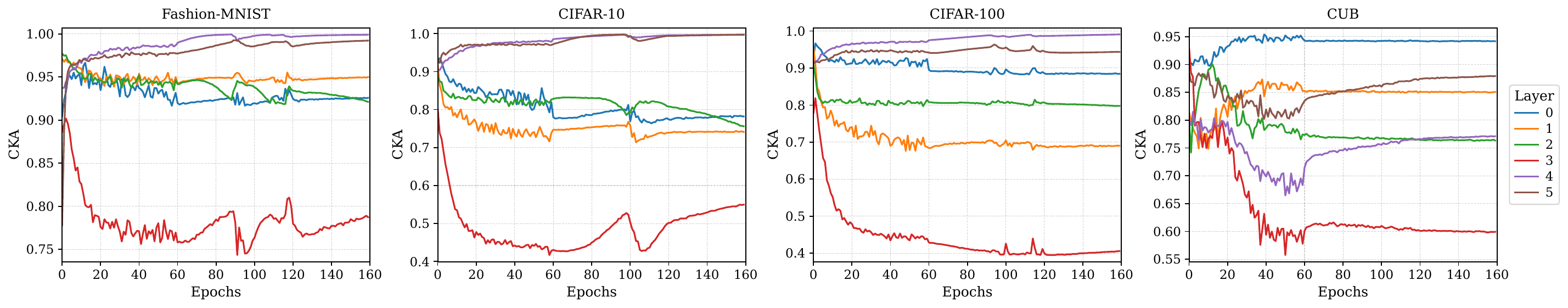} 
        
        \caption{Evolution during training of the CKA for consecutive layers in a ResNet10 trained on different datasets.}
    \end{subfigure}
    \begin{subfigure}{\textwidth}
        \centering
        \includegraphics[width=\textwidth]{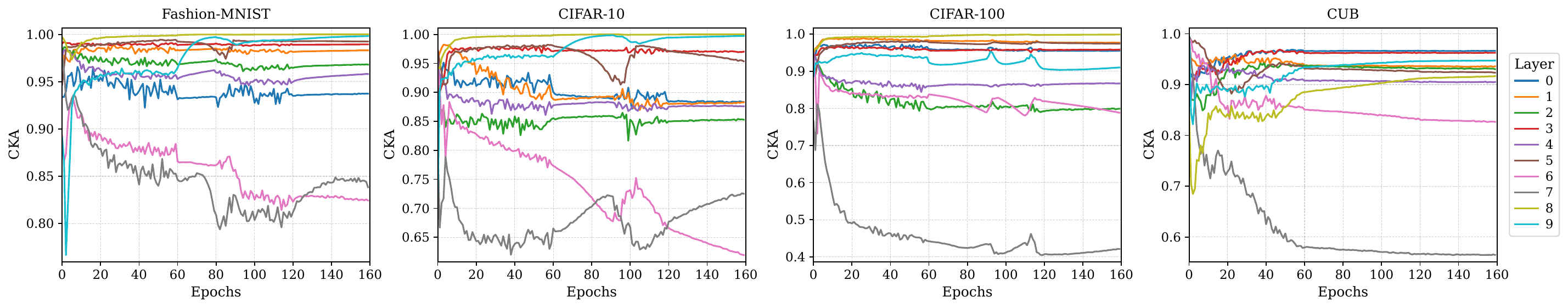} 
        \caption{Evolution during training of the CKA for consecutive layers in a ResNet18 trained on different datasets.}
        \label{subfig:resnet18_cka}
    \end{subfigure}

    \begin{subfigure}{\textwidth}
        \centering
        \includegraphics[width=0.3\textwidth]{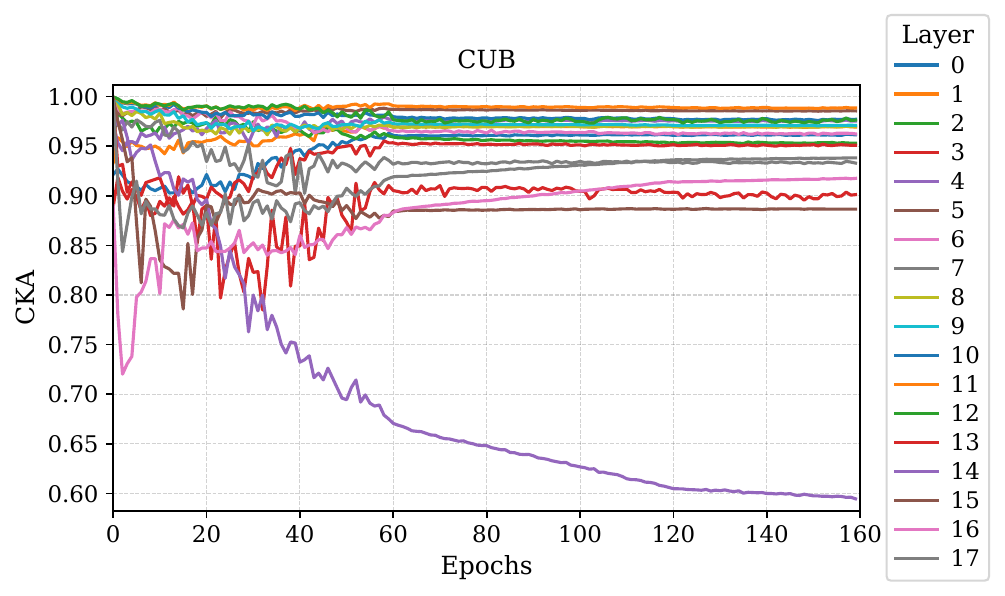} 
        
        \caption{Evolution during training of the CKA for consecutive layers in a ResNet34 trained on CUB.}
    \end{subfigure}
    
   \caption{Convergence through epochs of the CKA between consecutive layers of different ResNets trained on various datasets.}
    \label{fig:cka_resnet}
\end{figure*}

\begin{figure*}[h]
    \centering
    
     \begin{subfigure}{\textwidth}
        \centering
        \includegraphics[width=0.83\textwidth]{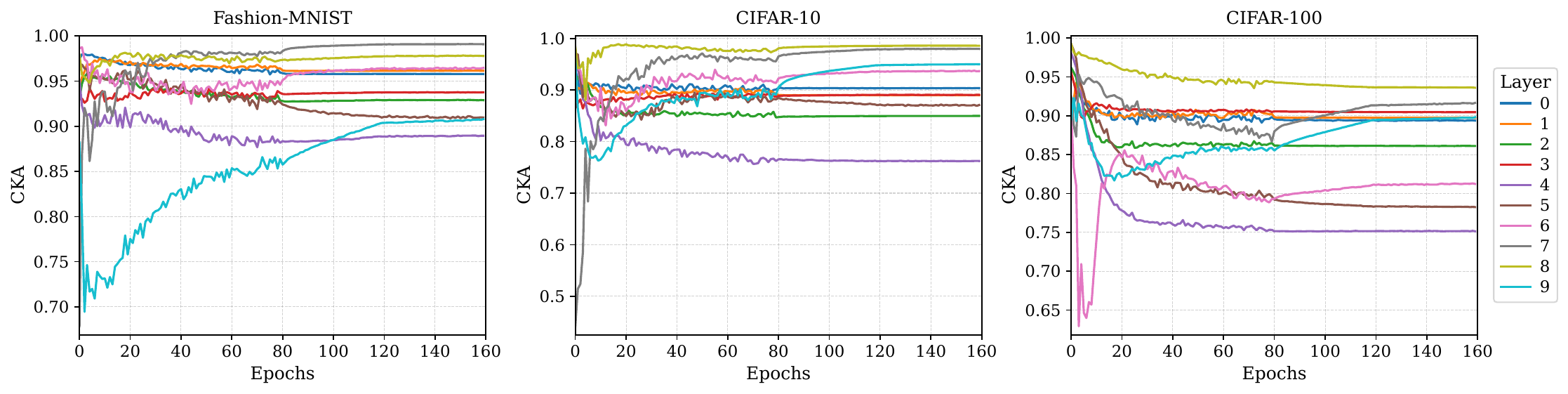} 
        \caption{Evolution during training of the CKA for consecutive layers in a VGG11 trained on different datasets.}
        \label{subfig:vgg11_cka}
    \end{subfigure}

    \begin{subfigure}{\textwidth}
        \centering
        \includegraphics[width=0.83\textwidth]{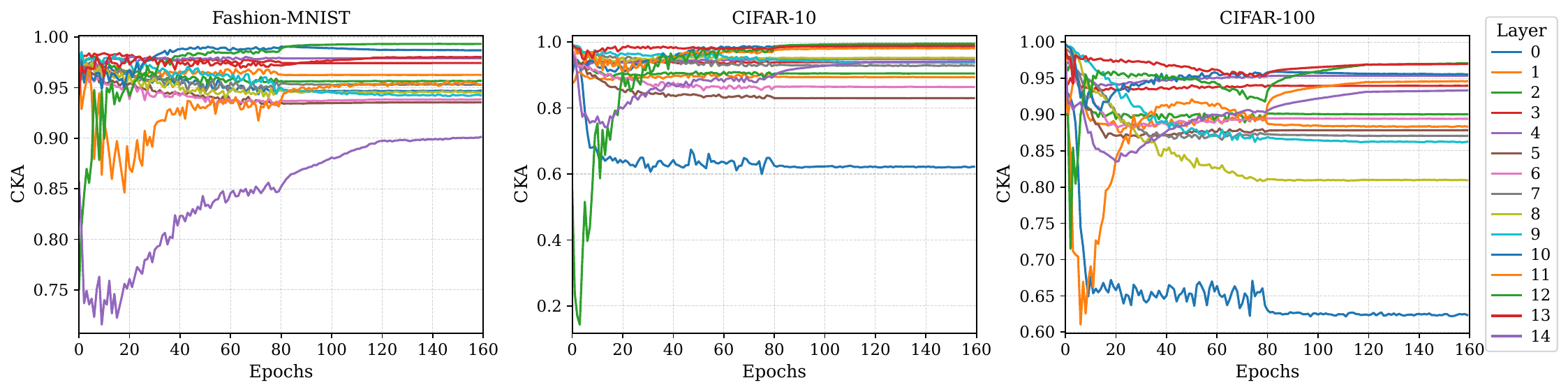} 
        
        \caption{Evolution during training of the CKA for consecutive layers in a VGG16 trained on different datasets.}
    \end{subfigure}

    \caption{Convergence through epochs of the CKA between consecutive layers of different VGGs trained on various datasets.}
    \label{fig:cka_vgg}
\end{figure*}

Considering, for example, ResNet18 on CIFAR-10 in Figure~\ref{subfig:resnet18_cka}, the two layers exhibiting the lowest CKA values are Layers~6 and~7, which is consistent with the split point of $7.4$ reported in Table~\ref{tab:models_split_analysis}. Instead, for VGG11 on CIFAR-100 in Figure~\ref{subfig:vgg11_cka}, Layers~5 and~6, which are indicated as the splitting point in the same table, display low CKA values, although the minimum is attained at Layer~4. This pattern is consistently observed only for ResNet architectures, whereas for MLPs and VGGs, the identified split does not always coincide with the minimum CKA value. This behavior can be explained by the observation that the splitting process is unlikely to be sharp and is instead inherently gradual and continuous through layers. Indeed, the CKA values are often uniformly high, thereby limiting the discriminative power of local minima. However, in ResNet architectures, where residual blocks, rather than individual layers, are considered, this gradual nature allows representational differences to accumulate across successive blocks, resulting in a clearer alignment between the split identified by the proposed metrics and the observed changes in representation.

\begin{figure*}[h]
    \centering
    
    \begin{subfigure}{\textwidth}
        \centering
        \includegraphics[width=0.83\textwidth]{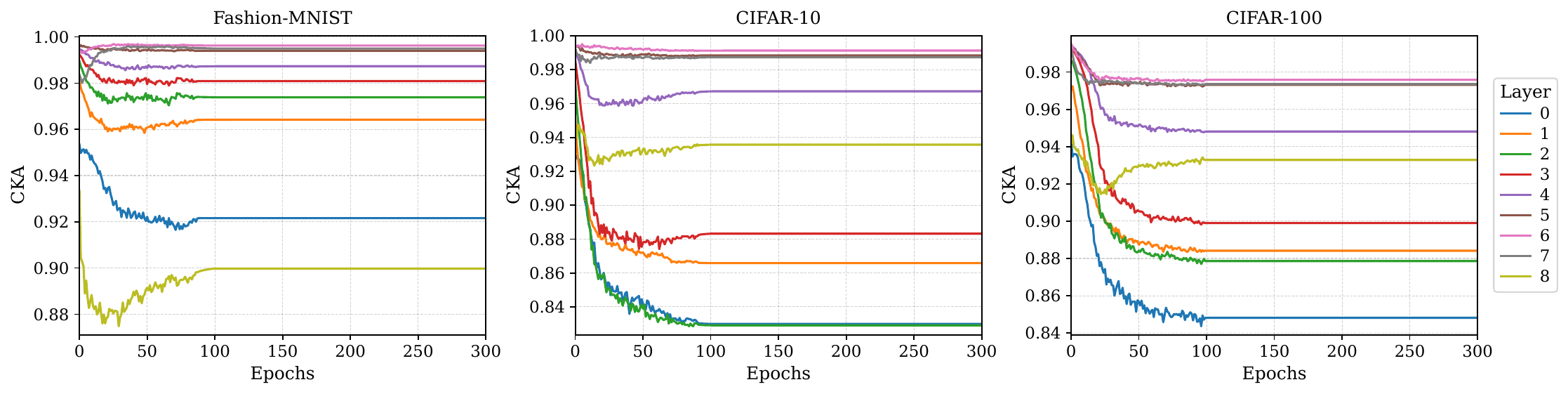}
        \caption{Evolution during training of the CKA for consecutive layers in a MLP10 trained on different datasets.}
    \end{subfigure}

    \begin{subfigure}{\textwidth}
        \centering
        \includegraphics[width=0.83\textwidth]{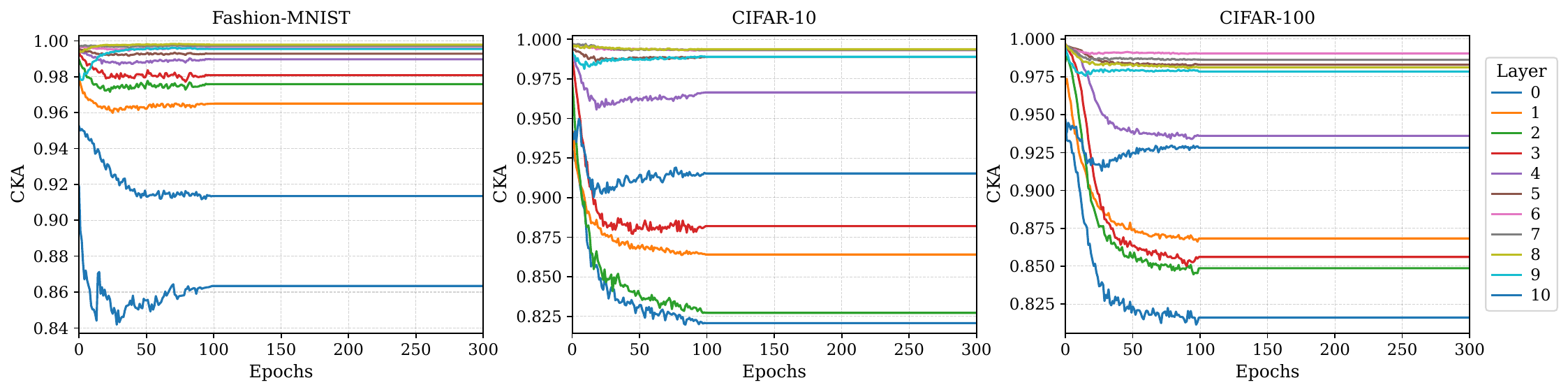}
        \caption{Evolution during training of the CKA for consecutive layers in a MLP12 trained on different datasets.}
    \end{subfigure}
    
    \caption{Convergence through epochs of the CKA between consecutive layers of different MLPs trained on various datasets.}
    \label{fig:cka_mlp}
\end{figure*}


\clearpage

\section{Experiments on generalization}
\label{app:res}

\subsection{Experiments on different classification heads}
\label{app:exps_clas_heads}

To analyze the interplay between parameter count, split depth, and classification-head design, we focus on ResNet10 trained on CIFAR-100.

\begin{table}[h]
    \centering
    \caption{Analysis of the performance of three different classification heads in terms of parameter count and accuracy across two critical split layers for ResNet-10 on CIFAR-100.}
    \resizebox{\columnwidth}{!}{
    \begin{tabular}{llll c r@{}l}
    \toprule
    Model & Dataset & Layer & Head type & Num. parameters & \multicolumn{2}{c}{Accuracy} \\
    \midrule
    ResNet10 & CIFAR-100 & 2 & Fully connected & 6{,}629{,}440 & 24.25 &\stdf{0.99}  \\
    ResNet10 & CIFAR-100 & 2 & Avg. Pooling + Fully connected & 82{,}240 & \textbf{45.35} & \stdf{0.64} \\
    ResNet10 & CIFAR-100 & 2 & Conv2D + Fully connected & 178{,}240 & 15.93 &\stdf{0.22} \\
    \midrule
    ResNet10 & CIFAR-100 & 5 & Fully connected & 5{,}717{,}312 & \textbf{60.73} &\stdf{0.41} \\
    ResNet10 & CIFAR-100 & 5 & Avg. Pooling + Fully connected & 4{,}949{,}312 & 60.10 & \stdf{0.17}  \\
    ResNet10 & CIFAR-100 & 5 & Conv2D + Fully connected & 4{,}899{,}712 & 38.38&\stdf{0.77}  \\
    \bottomrule
    \end{tabular}}
    \label{tab:ablation_head}
\end{table}

This setting is particularly informative because, as shown in Table~\ref{tab:ablation_head}, splitting the network at layers 2 or 5 and attaching a simple fully connected head can produce a model with \emph{more} parameters than the original architecture, which contains $4{,}949{,}412$ parameters. Interestingly, for this configuration, the split layer selected across all optimization seeds in Table~\ref{tab:models_split_analysis} is consistently layer 4, which still yields a $42\%$ reduction in total parameters. This phenomenon appears mainly in ResNet architectures and in datasets with many classes, such as CIFAR-100 and CUB, where flattening high-dimensional embeddings at intermediate depths can create a large classification head. Since this issue arises precisely in the more challenging settings, we investigate which head designs best mitigate this parameter explosion. Given a feature tensor of size $B \times C \times H \times W$, where $B$ is the batch size, $C$ is the number of channels, and $H, W$ are the height and width of the feature map, we evaluate three alternative classification heads:
\begin{enumerate}
    \item \textbf{Fully connected}: the feature map is directly flattened, yielding a representation of size $B \times (C \cdot H \cdot W)$;
    \item \textbf{Average pooling + fully connected}: a global adaptive average pooling is first applied, followed by flattening, resulting in a $B \times C$ representation;
    \item \textbf{Conv2D + fully connected}: the features are first averaged across channels at each spatial location and then flattened, producing a $B \times (H \cdot W)$ representation.
\end{enumerate}

Table~\ref{tab:ablation_head} highlights several noteworthy patterns. First, the number of parameters is not directly correlated with generalization performance; rather, performance depends critically on how the learned representations are processed. For instance, when the network is split at layer 2, an average pooling operation followed by a fully connected layer outperforms a single fully connected head, despite using only $2.7\%$ of the original parameters. A second important observation is that the split depth itself plays a significant role, in agreement with the Tunnel Effect and the Information Bottleneck Principle. Specifically, splitting the network at layer 5 instead of layer 2 yields higher final accuracy, even though fewer parameters are retained. This suggests that deeper layers produce more informative and compact representations. Furthermore, for a fixed split point, average pooling consistently leads to a larger parameter reduction and to a higher test accuracy than a convolutional layer, motivating our choice of this design for the CUB experiments. Among the lightweight alternatives explored here, channel-wise compression followed by a small fully connected head therefore appears to offer the best trade-off between compactness and accuracy. Finally, the impact of the classification head diminishes as the split occurs deeper in the network: at later layers, the representations have already undergone substantial processing, effectively separating signal from noise, and consequently, the performance differences between head designs at layer 5 are much less pronounced than those observed at layer 2.

\subsection{Layer and epoch dependence on generalization}

Here, we report how the generalization performance of the different models, evaluated across the selected datasets, depends on the choice of split layer (where the classification head is attached) and on the starting epoch from which the simplified model, derived from the full network, is trained.

\begin{figure*}[h]
    \centering
    
    \begin{subfigure}{\textwidth}
        \centering
        \includegraphics[width=\textwidth]{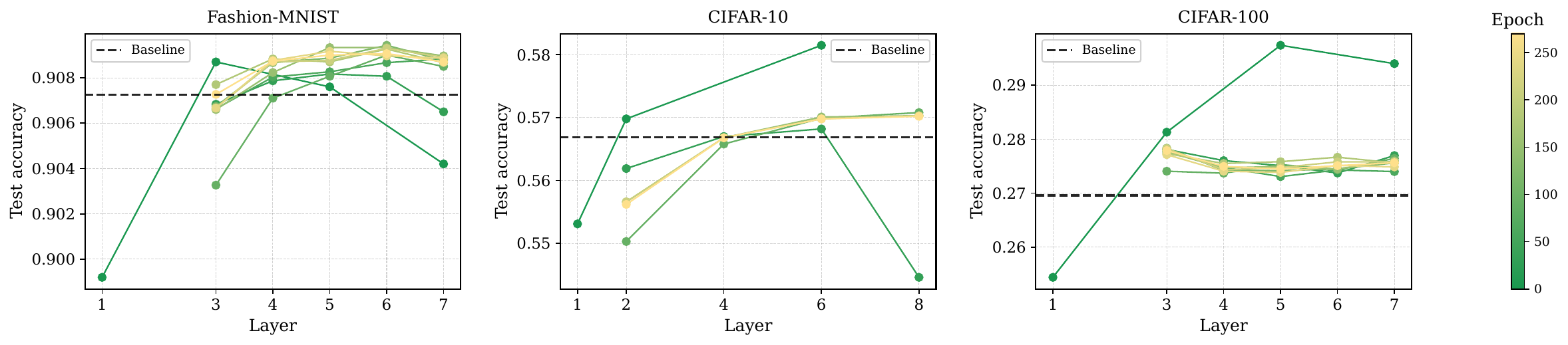} 
        
        \caption{Test accuracy of a MLP10 on different split layers varying the starting training epoch, through different datasets.}
    \end{subfigure}

    \begin{subfigure}{\textwidth}
        \centering
        \includegraphics[width=\textwidth]{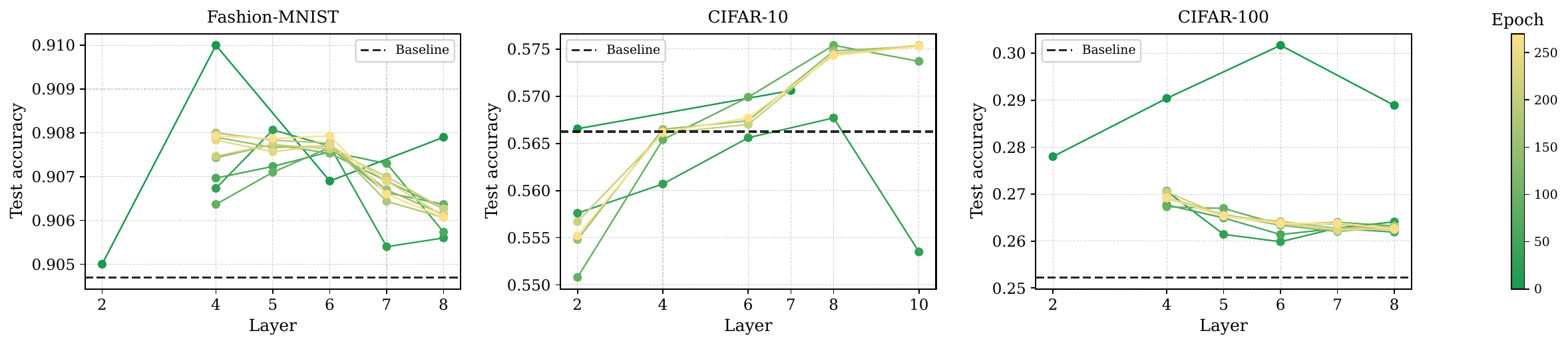} 
        
        \caption{Test accuracy of a MLP12 on different split layers varying the starting training epoch, through different datasets.}
    \end{subfigure}

    \caption{Generalization ability of MLPs through different model depths, splitting layers, and starting training epochs.}
    \label{fig:mlp_oracle}
\end{figure*}
\begin{figure*}[h]
    \centering
    
    \begin{subfigure}{\textwidth}
        \centering
        \includegraphics[width=\textwidth]{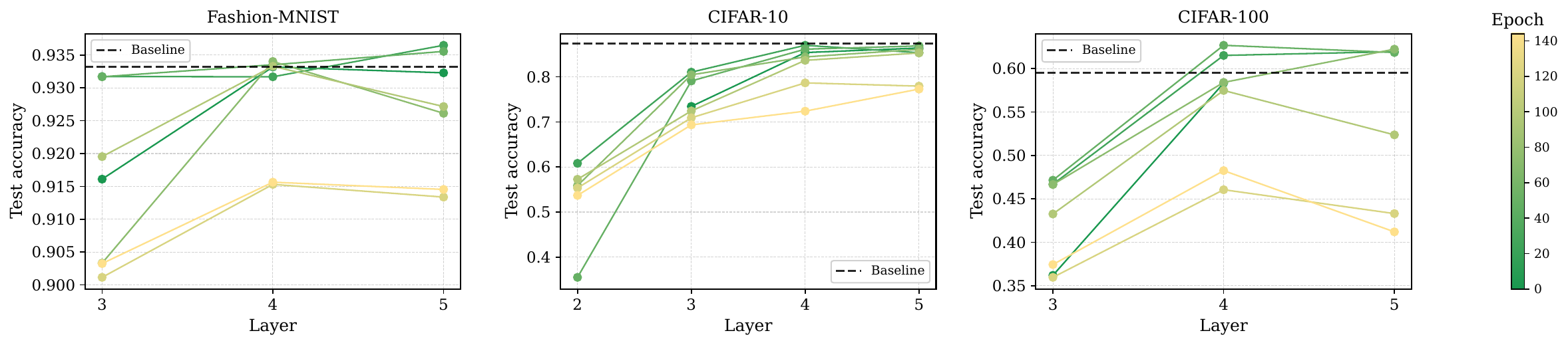} 
        
        \caption{Test accuracy of a ResNet10 on different split layers varying the starting training epoch, through different datasets.}
    \end{subfigure}

    \begin{subfigure}{\textwidth}
        \centering
        \includegraphics[width=\textwidth]{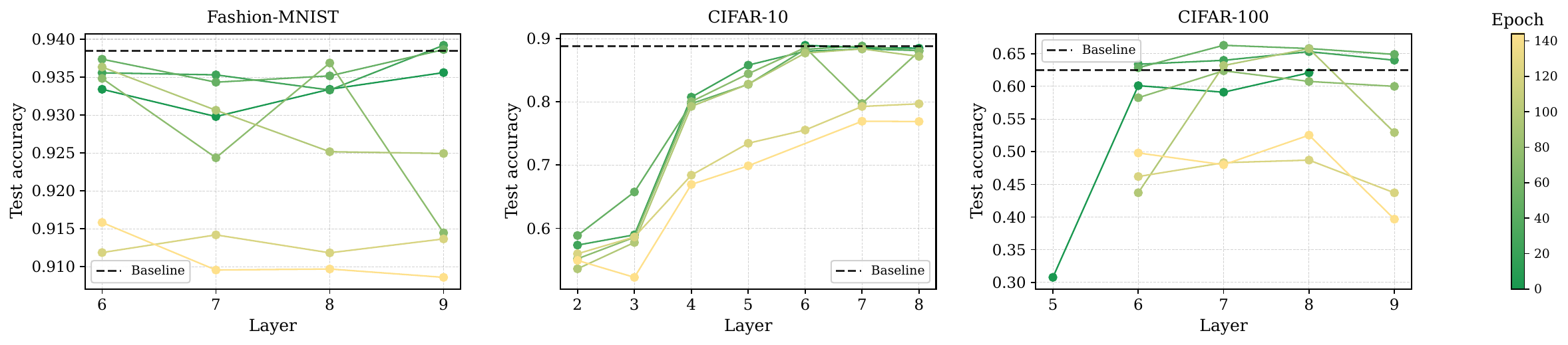} 
        
        \caption{Test accuracy of a ResNet18 on different split layers varying the starting training epoch, through different datasets.}
    \end{subfigure}

    \caption{Generalization ability of ResNets through different model depths, splitting layers, and starting training epochs.}
    \label{fig:resnet_oracle}
\end{figure*}
\begin{figure*}[h]
    \centering
    
    \begin{subfigure}{\textwidth}
        \centering
        \includegraphics[width=\textwidth]{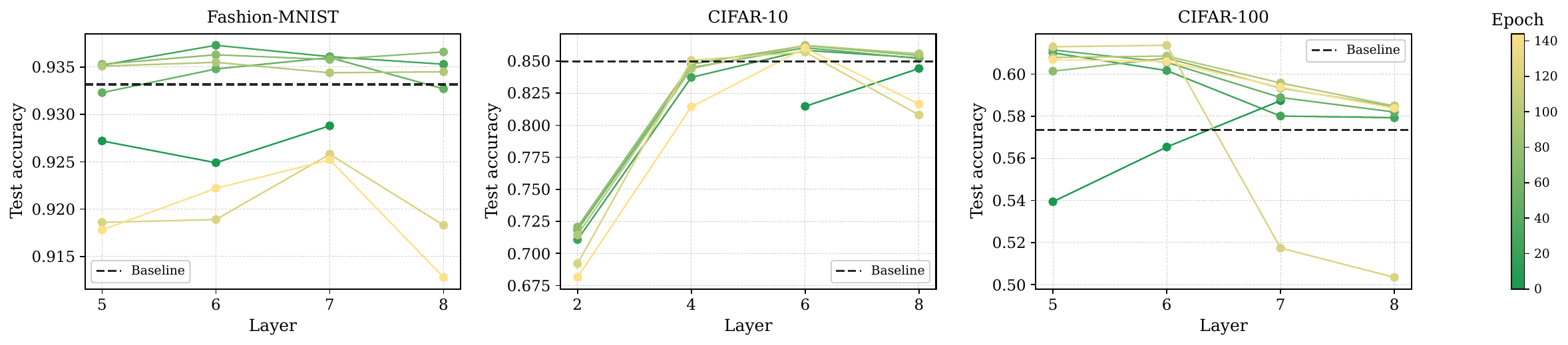} 
        
        \caption{Test accuracy of a VGG11 on different split layers varying the starting training epoch, through different datasets.}
    \end{subfigure}

    \begin{subfigure}{\textwidth}
        \centering
        \includegraphics[width=\textwidth]{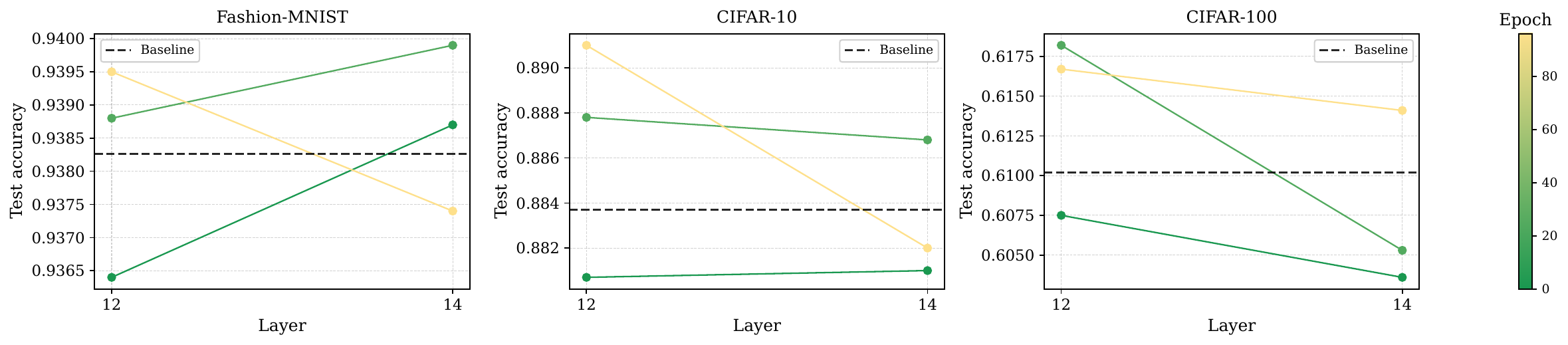} 
        
        \caption{Test accuracy of a VGG16 on different split layers varying the starting training epoch, through different datasets.}
    \end{subfigure}

    \caption{Generalization ability of VGGs through different model depths, splitting layers, and starting training epochs.}
    \label{fig:vgg_oracle}
\end{figure*}
\begin{figure*}[h]
    \centering
    
    \begin{subfigure}{\textwidth}
        \centering
        \includegraphics[width=\textwidth]{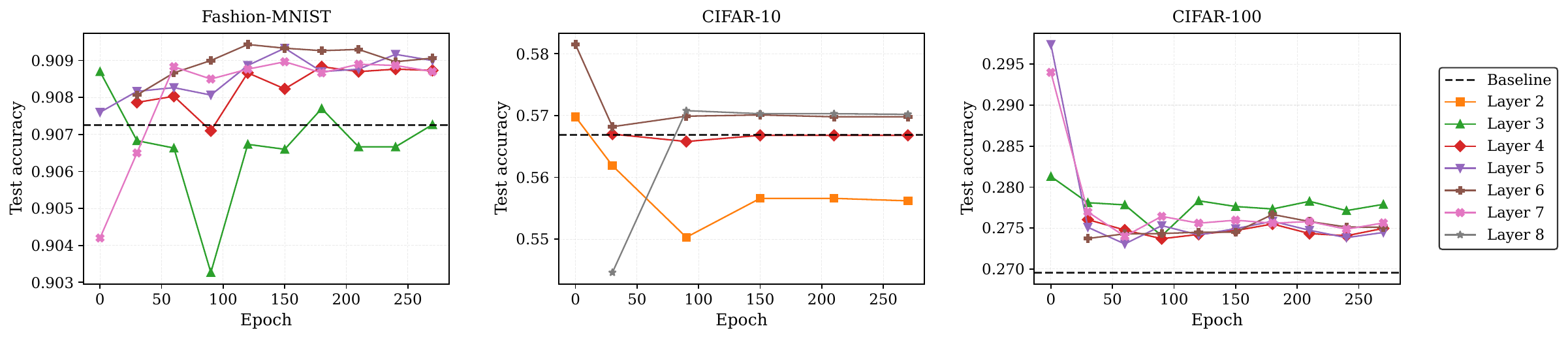}
        
        \caption{Test accuracy of a MLP10 on different starting training epochs varying the split layers, through different datasets.}
    \end{subfigure}

    \begin{subfigure}{\textwidth}
        \centering
        \includegraphics[width=\textwidth]{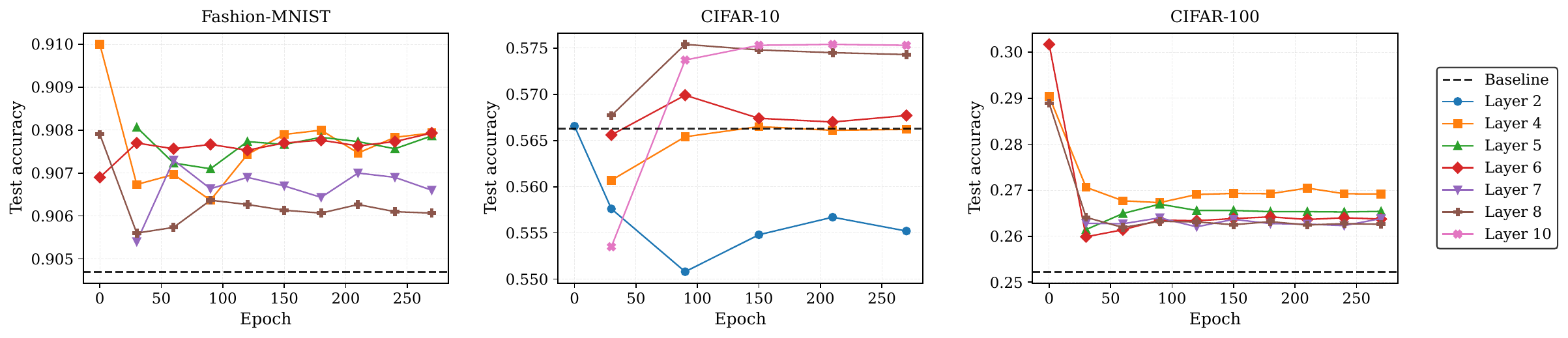}
        
        \caption{Test accuracy of a MLP12 on different starting training epochs varying the split layers, through different datasets.}
    \end{subfigure}

    \caption{Generalization ability of MLPs through different model depths, datasets, starting training epochs, and splitting layers.}
    \label{fig:mlp_oracle_epochs}
\end{figure*} 
\begin{figure*}[h]
    \centering
    
    \begin{subfigure}{\textwidth}
        \centering
        \includegraphics[width=\textwidth]{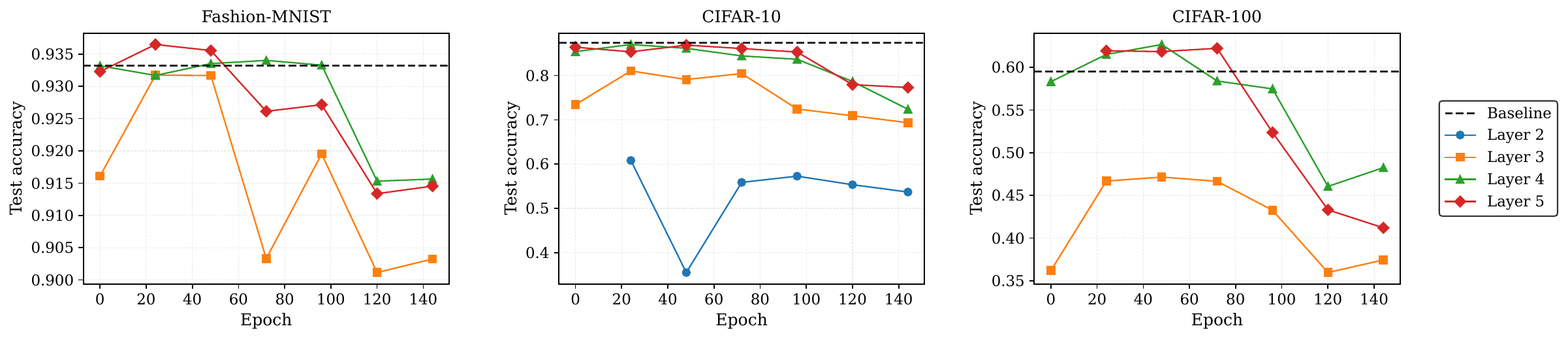} 
        
        \caption{Test accuracy of a ResNet10 on different starting training epochs varying the split layers, through different datasets.}
    \end{subfigure}

    \begin{subfigure}{\textwidth}
        \centering
        \includegraphics[width=\textwidth]{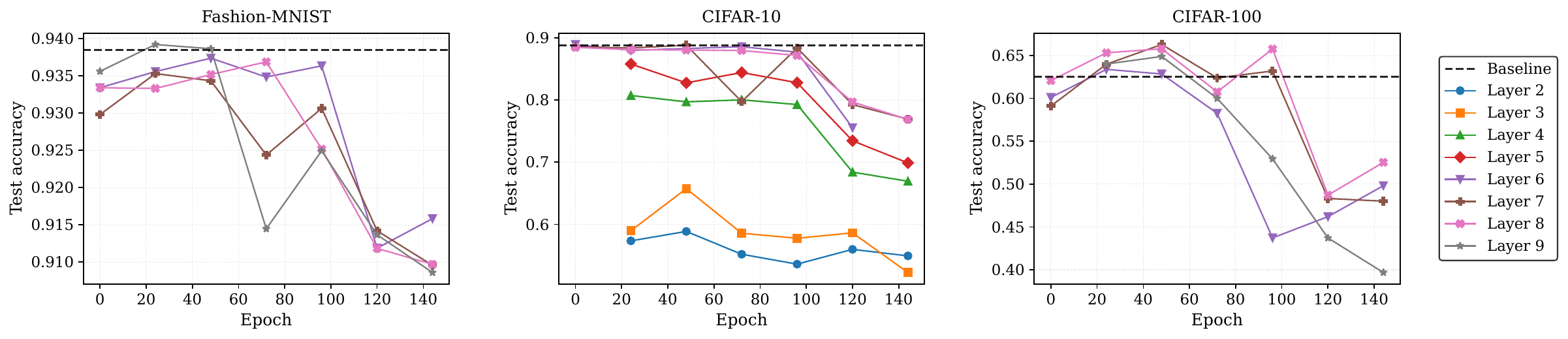}
        
        \caption{Test accuracy of a ResNet18 on different starting training epochs varying the split layers, through different datasets.}
    \end{subfigure}

    \caption{Generalization ability of ResNets through different model depths, datasets, starting training epochs, and splitting layers.}
    \label{fig:resnet_oracle_epochs}
\end{figure*}
\begin{figure*}[h]
    \centering
    
    \begin{subfigure}{\textwidth}
        \centering
        \includegraphics[width=\textwidth]{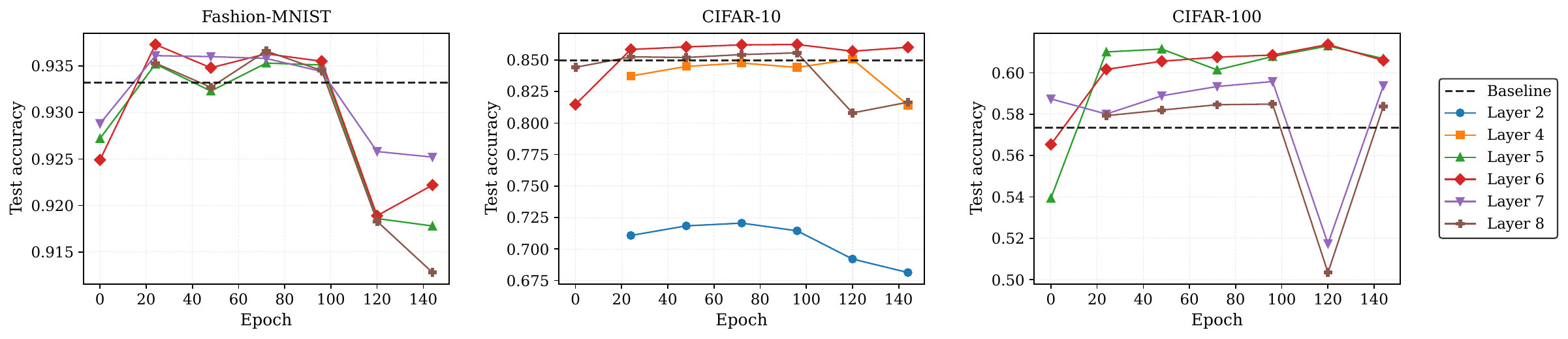}
        
        \caption{Test accuracy of a VGG11 on different starting training epochs varying the split layers, through different datasets.}
    \end{subfigure}

    \begin{subfigure}{\textwidth}
        \centering
        \includegraphics[width=\textwidth]{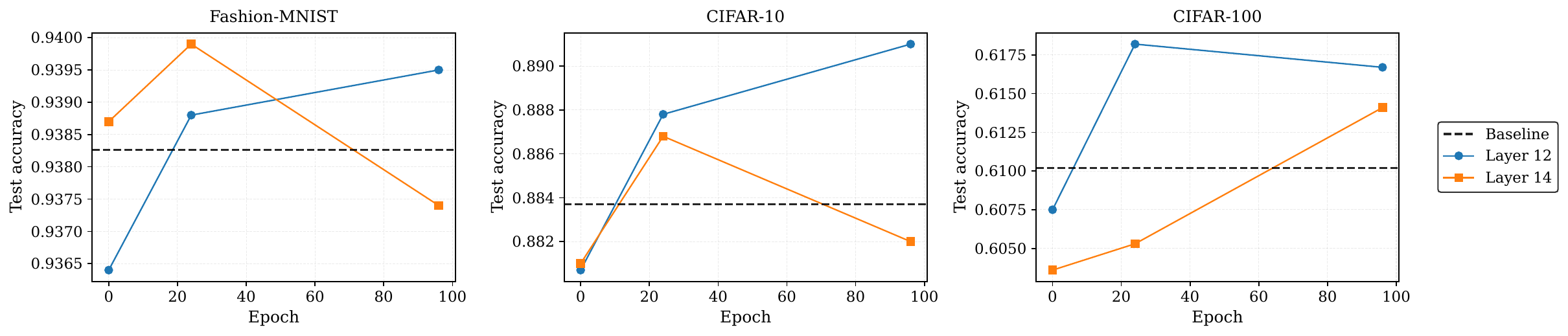}
        
        \caption{Test accuracy of a VGG16 on different starting training epochs varying the split layers, through different datasets.}
    \end{subfigure}

    \caption{Generalization ability of VGGs through different model depths, datasets, starting training epochs, and splitting layers.}
    \label{fig:vgg_oracle_epochs}
\end{figure*}

\clearpage

\section{Accuracy and FLOPs Results}
\label{app:acc_res}

\subsection{Accuracy Analysis}
In all the following tables, the mean accuracy and standard deviation are over five independent runs. The symbol $^*$ indicates that at least one run failed during training.

\begin{table}[h]
    \centering
    \caption{Test accuracy on Fashion-MNIST. Bold marks the best result, and underlining marks the second best.}
    \resizebox{\columnwidth}{!}{
    \begin{tabular}{l|r@{}l r@{}l  r@{}l r@{}l  r@{}l r@{}l}
    \toprule
    \multirow{2}{*}{Method} & \multicolumn{12}{c}{Models} \\
    & \multicolumn{2}{c}{MLP10} & \multicolumn{2}{c}{MLP12} & \multicolumn{2}{c}{ResNet10} & \multicolumn{2}{c}{ResNet18} & \multicolumn{2}{c}{VGG11} & \multicolumn{2}{c}{VGG16} \\
    \midrule
        Full training & 90.73 &\stdf{0.10} & 90.47 &\stdf{0.13} & \underline{93.32} &\stdf{0.23} & \textbf{93.85} &\stdf{0.12} & \underline{93.32} &\stdf{0.17} & 93.83 &\stdf{0.19}\\
       $\text{\ours}_{\text{LP}}$  & \textbf{90.82} &\stdf{0.06} & \underline{90.67} &\stdf{0.13} & 93.27 &\stdf{0.16} & 93.57 &\stdf{0.07} & \textbf{93.37} &\stdf{0.12} & \underline{93.87} &\stdf{0.10}\\
       $\text{\ours}_{\text{DCL}}$ & \underline{90.79} &\stdf{0.10} & \textbf{90.76} &\stdf{0.15} & 92.46 &\stdf{0.41} & \underline{93.68} &\stdf{0.02} & 93.31 &\stdf{0.14} & \textbf{93.93} &\stdf{0.12} \\
       $\text{\ours}_{\text{FIX}}$ & 84.15 &\stdf{5.41} & $89.87^*$ &\stdf{0.15} & 84.70 &\stdf{0.46} & 91.46 &\stdf{0.63} & 82.72 &\stdf{15.14} & 93.56 &\stdf{0.22} \\
       $\text{EB-LTH}_{30}$ & \multicolumn{2}{c}{--} & \multicolumn{2}{c}{--} & \textbf{93.46} &\stdf{0.15} & 92.32 &\stdf{0.40} & $85.05^*$ &\stdf{16.03} & $82.36^*$ &\stdf{22.69}\\
       $\text{EB-LTH}_{50}$  & \multicolumn{2}{c}{--} & \multicolumn{2}{c}{--} & 93.17 &\stdf{0.43} & 92.52 &\stdf{0.41} & $77.11^*$ &\stdf{31.62} & $85.55^*$ &\stdf{16.18}\\
       $\text{EB-LTH}_{70}$  & \multicolumn{2}{c}{--} & \multicolumn{2}{c}{--} & 92.89 &\stdf{0.18} & 92.26 &\stdf{0.68} & $80.33^*$ &\stdf{21.56} & $75.49^*$ &\stdf{31.43}\\
       \midrule
       $\%$Reduction & \multicolumn{2}{c}{48.8\%} & \multicolumn{2}{c}{54.5\%} & \multicolumn{2}{c}{71.7\%} & \multicolumn{2}{c}{41.6\%} & \multicolumn{2}{c}{77.2\%} & \multicolumn{2}{c}{$77.23\%$} \\
       $\%$Training & \multicolumn{2}{c}{19.1\%} & \multicolumn{2}{c}{21.6\%} & \multicolumn{2}{c}{12.1\%} & \multicolumn{2}{c}{13.9\%} & \multicolumn{2}{c}{18.4\%} & \multicolumn{2}{c}{38.75\%} \\
    \bottomrule
    \end{tabular}}
    \label{tab:fmnist}
\end{table}

\begin{table}[h]
    \centering
    \caption{Test accuracy on CIFAR-10. Bold marks the best result, and underlining marks the second best.}
    \resizebox{\columnwidth}{!}{
    \begin{tabular}{l|r@{}l r@{}l  r@{}l r@{}l  r@{}l r@{}l}
    \toprule
    \multirow{2}{*}{Method} & \multicolumn{12}{c}{Models} \\
    & \multicolumn{2}{c}{MLP10} & \multicolumn{2}{c}{MLP12} & \multicolumn{2}{c}{ResNet10} & \multicolumn{2}{c}{ResNet18} & \multicolumn{2}{c}{VGG11} & \multicolumn{2}{c}{VGG16} \\
    \midrule
        Full training & \underline{56.68} &\stdf{0.31} & \underline{56.63} &\stdf{0.32} & \textbf{87.40} &\stdf{0.09} & 88.79 &\stdf{0.50} & 84.96 &\stdf{0.30} & \underline{88.37} &\stdf{0.20}\\
       $\text{\ours}_{\text{LP}}$  & \textbf{56.82} &\stdf{0.32} & \textbf{56.82} &\stdf{0.30} & 86.05 &\stdf{0.57} & \underline{88.89} &\stdf{0.83} & \underline{85.04} &\stdf{0.46} & 88.36 &\stdf{0.23}\\
       $\text{\ours}_{\text{DCL}}$  & 56.53 &\stdf{0.20} & 56.37 &\stdf{1.25} & \underline{86.61} &\stdf{0.30} & 87.56 &\stdf{1.59} & \textbf{85.35} &\stdf{0.35} & \textbf{88.83} &\stdf{0.40} \\
       $\text{\ours}_{\text{FIX}}$  & 54.31 &\stdf{0.46} & 54.45 &\stdf{0.17} & 42.69 &\stdf{2.38} & $81.99^*$ &\stdf{1.83} & 74.85 &\stdf{4.95} & 83.46 &\stdf{1.34} \\
       $\text{EB-LTH}_{30}$ & \multicolumn{2}{c}{--} & \multicolumn{2}{c}{--} & 86.03 &\stdf{1.32} & \textbf{89.08} &\stdf{0.21} & 78.39 &\stdf{12.50} & 86.67 &\stdf{0.88} \\
       $\text{EB-LTH}_{50}$ & \multicolumn{2}{c}{--} & \multicolumn{2}{c}{--} & 85.95 &\stdf{1.05} & 88.88 &\stdf{0.23} & 78.14 &\stdf{13.55} & 74.83 &\stdf{24.85}\\
       $\text{EB-LTH}_{70}$  & \multicolumn{2}{c}{--} & \multicolumn{2}{c}{--} & 85.80 &\stdf{0.81} & 87.24 &\stdf{2.15} & 77.45 &\stdf{13.20} & $85.73^*$ &\stdf{1.10} \\
       \midrule
       $\%$Reduction & \multicolumn{2}{c}{45.4\%} & \multicolumn{2}{c}{52.3\%} & \multicolumn{2}{c}{71.7\%} & \multicolumn{2}{c}{60.8\%} & \multicolumn{2}{c}{94.2\%} & \multicolumn{2}{c}{$84.06\%$}\\
       $\%$Training & \multicolumn{2}{c}{8.0\%} & \multicolumn{2}{c}{7.9\%} & \multicolumn{2}{c}{10.0\%} & \multicolumn{2}{c}{23.3\%} & \multicolumn{2}{c}{15.6\%} & \multicolumn{2}{c}{52.0\%} \\
       
    \bottomrule
    \end{tabular}}
    \label{tab:cifar10}
\end{table}

\begin{table}[t]
    \centering
    \caption{Test accuracy on CIFAR-100, reported as mean and standard deviation over 5 runs. Bold marks the best result, and underlining marks the second best.}
    \resizebox{\columnwidth}{!}{
    \begin{tabular}{l|r@{}l r@{}l  r@{}l r@{}l  r@{}l r@{}l}
    \toprule
    \multirow{2}{*}{Method} & \multicolumn{12}{c}{Models} \\
    & \multicolumn{2}{c}{MLP10} & \multicolumn{2}{c}{MLP12} & \multicolumn{2}{c}{ResNet10} & \multicolumn{2}{c}{ResNet18} & \multicolumn{2}{c}{VGG11} & \multicolumn{2}{c}{VGG16} \\
    \midrule
        Full training & \underline{26.96} &\stdf{0.77} & \underline{25.23} &\stdf{0.65} & 59.53 &\stdf{0.74} & \underline{62.51} &\stdf{0.35} & \underline{57.34} &\stdf{0.34} & \underline{61.02} &\stdf{0.13}\\
       $\text{\ours}_{\text{LP}}$  & \textbf{27.03} &\stdf{0.32} & \textbf{26.20} &\stdf{0.41} & \textbf{60.98} &\stdf{0.58} & \textbf{63.70} &\stdf{0.33} & \textbf{58.17} &\stdf{0.37} & \textbf{62.76} &\stdf{0.56}\\
       $\text{\ours}_{\text{DCL}}$  & 24.66 & \stdf{0.77} & 24.80 & \stdf{0.30} & 44.47 &\stdf{1.24} & 54.44 & \stdf{0.52} & 54.83 & \stdf{1.24} & 57.42 & \stdf{3.37}\\
       $\text{\ours}_{\text{FIX}}$  & $22.97^*$&\stdf{0.30}  & $24.16^*$&\stdf{0.20}  & $13.01^*$ &\stdf{1.16} &  $40.85^*$ &\stdf{4.63} & $35.84^*$ &\stdf{2.03} & $43.97^*$ &\stdf{6.31} \\
       $\text{EB-LTH}_{30}$ & \multicolumn{2}{c}{--} & \multicolumn{2}{c}{--} & \underline{60.40} &\stdf{0.62} & 61.09 &\stdf{1.43} & $50.30^*$ &\stdf{2.52} & 58.85 &\stdf{4.75}\\
       $\text{EB-LTH}_{50}$  & \multicolumn{2}{c}{--} & \multicolumn{2}{c}{--} & 58.32 &\stdf{0.58} & 59.27 &\stdf{1.70} & $38.15^*$ &\stdf{6.37} & 60.42 &\stdf{0.54}\\
       $\text{EB-LTH}_{70}$  & \multicolumn{2}{c}{--} & \multicolumn{2}{c}{--} & 49.49 &\stdf{0.49} & 57.30 &\stdf{1.06} & $8.02^*$ &\stdf{0.00} & 51.60 &\stdf{11.05}\\
       \midrule
       $\%$Reduction & \multicolumn{2}{c}{36.1\%} & \multicolumn{2}{c}{45.8\%} & \multicolumn{2}{c}{42.1\%} & \multicolumn{2}{c}{35.2\%} & \multicolumn{2}{c}{76.9\%} & \multicolumn{2}{c}{$82.07\%$}\\
       $\%$Training & \multicolumn{2}{c}{9.8\%} & \multicolumn{2}{c}{8.9\%} & \multicolumn{2}{c}{10.0\%} & \multicolumn{2}{c}{10.8\%} & \multicolumn{2}{c}{13.9\%} & \multicolumn{2}{c}{20.0\%}\\
    \bottomrule
    \end{tabular}}
    \label{tab:cifar100}
\end{table}

\begin{table}[]
    \centering
    \caption{Test accuracy on CUB-200-2011, reported as mean and standard deviation over 5 runs. Bold marks the best result, and underlining marks the second best.}
    
    \begin{tabular}{l|r@{}l r@{}l r@{}l}
    \toprule
    \multirow{2}{*}{Method} & \multicolumn{6}{c}{Models} \\
     & \multicolumn{2}{c}{ResNet10} & \multicolumn{2}{c}{ResNet18} & \multicolumn{2}{c}{ResNet34} \\
    \midrule
        Full training & 36.02 &\stdf{1.12} & 48.70 &\stdf{0.97} & 47.75 &\stdf{0.78} \\
       $\text{\ours}_{\text{LP}}$ & 25.04 &\stdf{2.38} & \underline{49.17} &\stdf{0.91} & \textbf{50.94} &\stdf{2.87} \\
       $\text{\ours}_{\text{FIX}}$  & 25.76 &\stdf{2.75} & 30.24 &\stdf{18.19} & 36.23 &\stdf{5.98} \\
       $\text{EB-LTH}_{30}$ & 35.20 &\stdf{1.18} & 48.76 &\stdf{0.78} & \underline{47.91} &\stdf{2.30} \\
       $\text{EB-LTH}_{50}$  & \textbf{36.52} &\stdf{0.87} & \textbf{49.49} &\stdf{0.97} & 47.16 &\stdf{2.42} \\
       $\text{EB-LTH}_{70}$  & \underline{36.51} &\stdf{0.93} & 47.90 &\stdf{0.79} & 46.40 &\stdf{2.10} \\
       \midrule
       $\%$Reduction & \multicolumn{2}{c}{$74.36\%$} & \multicolumn{2}{c}{$41.85\%$} & \multicolumn{2}{c}{$44.15\%$}  \\
       $\%$Training & \multicolumn{2}{c}{25.0\%} & \multicolumn{2}{c}{21.3\%} & \multicolumn{2}{c}{31.8\%} \\
    \bottomrule
    \end{tabular}
    \label{tab:cub}
\end{table}

As reported in Tables~\ref{tab:fmnist}, \ref{tab:cifar10}, and \ref{tab:cifar100}, $\text{EB-LTH}_{70}$ and $\text{\ours}_{\text{FIX}}$, which typically retain the smallest number of parameters, are associated with the lowest accuracies, often exhibiting substantial drops compared to full training. For ResNet architectures, EB-LTH with parameter reductions of 30\% and 50\% generally achieves performance comparable to the fully trained models. However, this approach is not applicable to MLPs due to the absence of channel-wise normalization, and it yields significantly lower average accuracies for VGG architectures. In this case, the degradation is caused by runs that, despite fitting the training set successfully, display poor generalization performance on the test set, thereby increasing the observed variance. 
The two strongest variants overall are $\text{\ours}_{\text{LP}}$ and $\text{\ours}_{\text{DCL}}$, both of which remain consistently close to the full-model accuracy. However, the declarative ETF variant does not justify its additional computational cost and was therefore not evaluated on CUB in Table~\ref{tab:cub}. Overall, these results support our central hypothesis: \ours can match full-training performance while reducing the number of parameters by between $35.2\%$ and $94.2\%$, depending on the model-dataset pair. Among the tested heads, linear probing provides the best robustness-to-cost trade-off, while global-pooling and channel-averaging variants are attractive when further reducing head size is important.

For CUB, the most challenging case is ResNet10, where $\text{\ours}_{\text{LP}}$ performs substantially worse than full training. A plausible explanation is the combination of a relatively shallow backbone, the fine-grained 200-class nature of the task, and the aggressive reduction reported in Table~\ref{tab:cub}, which leaves the truncated model with insufficient capacity. By contrast, deeper ResNet variants on the same dataset remain competitive after truncation, suggesting that the issue is not the dataset alone but the interaction between task complexity and the available post-split capacity. More broadly, the fact that the same procedure remains effective across Fashion-MNIST, CIFAR-10, CIFAR-100, and CUB indicates that the criterion is not tied to a single class cardinality or difficulty regime, although this should not be confused with a full robustness guarantee.

\subsection{FLOPs Analysis}
\label{app:flops_res}

Tables~\ref{tab:flops_fmnist}--\ref{tab:flops_cub} report the training FLOPs for the same configurations considered in the main accuracy comparison. Here, $\text{\ours}_{\text{AVG}}$ denotes the variant in which the new classification head uses average pooling before the final linear layer. Reduced FLOPs are reported in GFLOPs and correspond to the resulting computational cost after simplification, while `Reduced FLOPs Percentage' denotes the relative reduction with respect to full training, denoted with `Total GFLOPs'. Bold marks the lowest reduced FLOPs within each model block.

\begin{table*}[t]
\centering
\scriptsize
\caption{FLOPs comparison on Fashion-MNIST. `Reduced FLOPs' denotes the resulting training cost after simplification, expressed in GFLOPs; lower is better. Bold marks the lowest reduced FLOPs within each model block.}
\resizebox{\textwidth}{!}{
\begin{tabular}{l l l r r c}
\toprule
Method & Model & Dataset & Reduced GFLOPs & Total GFLOPs & Reduced FLOPs Percentage \\
\midrule
$\text{\ours}_{\text{LP}}$ & MLP10 & Fashion-MNIST & \textbf{376,062.2} & 684,234.3 & \textbf{45.04} \\
\midrule
$\text{\ours}_{\text{LP}}$ & MLP12 & Fashion-MNIST & \textbf{477,084.7} & 836,121.8 & \textbf{42.94} \\
\midrule
$\text{\ours}_{\text{LP}}$ & ResNet10 & Fashion-MNIST & 7,759,320.2 & 9,745,615.7 & 20.38 \\
$\text{\ours}_{\text{AVG}}$ & ResNet10 & Fashion-MNIST & 7,753,838.2 & 9,745,615.7 & 20.44 \\
$\text{EB-LTH}_{30}$ & ResNet10 & Fashion-MNIST & 7,302,692.2 & 9,745,615.7 & 25.07 \\
$\text{EB-LTH}_{50}$ & ResNet10 & Fashion-MNIST & 5,989,811.8 & 9,745,615.7 & 38.54 \\
$\text{EB-LTH}_{70}$ & ResNet10 & Fashion-MNIST & \textbf{3,350,533.0} & 9,745,615.7 & \textbf{65.62} \\
\midrule
$\text{\ours}_{\text{LP}}$ & ResNet18 & Fashion-MNIST & 18,882,659.2 & 21,384,561.5 & 11.70 \\
$\text{\ours}_{\text{AVG}}$ & ResNet18 & Fashion-MNIST & 18,880,104.3 & 21,384,561.5 & 11.71 \\
$\text{EB-LTH}_{30}$ & ResNet18 & Fashion-MNIST & 17,643,962.3 & 21,384,561.5 & 17.49 \\
$\text{EB-LTH}_{50}$ & ResNet18 & Fashion-MNIST & 15,083,537.4 & 21,384,561.5 & 29.47 \\
$\text{EB-LTH}_{70}$ & ResNet18 & Fashion-MNIST & \textbf{11,874,908.9} & 21,384,561.5 & \textbf{44.47} \\
\midrule
$\text{\ours}_{\text{LP}}$ & VGG11 & Fashion-MNIST & 5,389,148.4 & 6,581,083.2 & 18.11 \\
$\text{\ours}_{\text{AVG}}$ & VGG11 & Fashion-MNIST & 5,388,663.3 & 6,581,083.2 & 18.12 \\
$\text{EB-LTH}_{30}$ & VGG11 & Fashion-MNIST & 4,954,354.5 & 6,581,083.2 & 24.72 \\
$\text{EB-LTH}_{50}$ & VGG11 & Fashion-MNIST & 3,854,302.2 & 6,581,083.2 & 41.43 \\
$\text{EB-LTH}_{70}$ & VGG11 & Fashion-MNIST & \textbf{2,708,481.6} & 6,581,083.2 & \textbf{58.84} \\
\midrule
$\text{\ours}_{\text{LP}}$ & VGG16 & Fashion-MNIST & 11,650,819.6 & 12,764,986.8 & 8.73 \\
$\text{\ours}_{\text{AVG}}$ & VGG16 & Fashion-MNIST & 11,650,456.7 & 12,764,986.8 & 8.73 \\
$\text{EB-LTH}_{30}$ & VGG16 & Fashion-MNIST & 9,227,832.3 & 12,764,986.8 & 27.71 \\
$\text{EB-LTH}_{50}$ & VGG16 & Fashion-MNIST & 7,004,604.5 & 12,764,986.8 & 45.13 \\
$\text{EB-LTH}_{70}$ & VGG16 & Fashion-MNIST & \textbf{4,706,410.7} & 12,764,986.8 & \textbf{63.13} \\
\bottomrule
\end{tabular}}
\label{tab:flops_fmnist}
\end{table*}

\begin{table*}[t]
\centering
\scriptsize
\caption{FLOPs comparison on CIFAR-10. `Reduced FLOPs' denotes the resulting training cost after simplification, expressed in GFLOPs; lower is better. Bold marks the lowest reduced FLOPs within each model block.}
\resizebox{\textwidth}{!}{
\begin{tabular}{l l l r r c}
\toprule
Method & Model & Dataset & Reduced GFLOPs & Total GFLOPs & Reduced FLOPs Percentage \\
\midrule
$\text{\ours}_{\text{LP}}$ & MLP10 & CIFAR-10 & \textbf{405,070.4} & 696,880.3 & \textbf{41.87} \\
\midrule
$\text{\ours}_{\text{LP}}$ & MLP12 & CIFAR-10 & \textbf{413,721.2} & 823,507.2 & \textbf{49.76} \\
\midrule
$\text{\ours}_{\text{LP}}$ & ResNet10 & CIFAR-10 & 6,471,409.7 & 8,162,594.8 & 20.72 \\
$\text{\ours}_{\text{AVG}}$ & ResNet10 & CIFAR-10 & 6,466,742.2 & 8,162,594.8 & 20.78 \\
$\text{EB-LTH}_{30}$ & ResNet10 & CIFAR-10 & 6,026,228.6 & 8,162,594.8 & 26.17 \\
$\text{EB-LTH}_{50}$ & ResNet10 & CIFAR-10 & 4,959,861.1 & 8,162,594.8 & 39.24 \\
$\text{EB-LTH}_{70}$ & ResNet10 & CIFAR-10 & \textbf{2,921,712.0} & 8,162,594.8 & \textbf{64.21} \\
\midrule
$\text{\ours}_{\text{LP}}$ & ResNet18 & CIFAR-10 & 14,560,310.8 & 17,865,852.4 & 18.50 \\
$\text{\ours}_{\text{AVG}}$ & ResNet18 & CIFAR-10 & 14,556,324.0 & 17,865,852.4 & 18.52 \\
$\text{EB-LTH}_{30}$ & ResNet18 & CIFAR-10 & 14,760,443.5 & 17,865,852.4 & 17.38 \\
$\text{EB-LTH}_{50}$ & ResNet18 & CIFAR-10 & 12,603,453.8 & 17,865,852.4 & 29.46 \\
$\text{EB-LTH}_{70}$ & ResNet18 & CIFAR-10 & \textbf{10,354,020.3} & 17,865,852.4 & \textbf{42.05} \\
\midrule
$\text{\ours}_{\text{LP}}$ & VGG11 & CIFAR-10 & 3,480,542.3 & 5,524,359.8 & 37.00 \\
$\text{\ours}_{\text{AVG}}$ & VGG11 & CIFAR-10 & 3,478,458.6 & 5,524,359.8 & 37.03 \\
$\text{EB-LTH}_{30}$ & VGG11 & CIFAR-10 & 4,175,332.9 & 5,524,359.8 & 24.42 \\
$\text{EB-LTH}_{50}$ & VGG11 & CIFAR-10 & 3,361,391.1 & 5,524,359.8 & 39.15 \\
$\text{EB-LTH}_{70}$ & VGG11 & CIFAR-10 & \textbf{2,379,600.9} & 5,524,359.8 & \textbf{56.93} \\
\midrule
$\text{\ours}_{\text{LP}}$ & VGG16 & CIFAR-10 & 9,385,368.9 & 10,679,810.3 & 12.12 \\
$\text{\ours}_{\text{AVG}}$ & VGG16 & CIFAR-10 & 9,384,195.8 & 10,679,810.3 & 12.13 \\
$\text{EB-LTH}_{30}$ & VGG16 & CIFAR-10 & 7,747,368.6 & 10,679,810.3 & 27.46 \\
$\text{EB-LTH}_{50}$ & VGG16 & CIFAR-10 & 5,898,107.0 & 10,679,810.3 & 44.77 \\
$\text{EB-LTH}_{70}$ & VGG16 & CIFAR-10 & \textbf{4,020,066.0} & 10,679,810.3 & \textbf{62.36} \\
\bottomrule
\end{tabular}}
\label{tab:flops_cifar10}
\end{table*}

\begin{table*}[t]
\centering
\scriptsize
\caption{FLOPs comparison on CIFAR-100. `Reduced FLOPs' denotes the resulting training cost after simplification, expressed in GFLOPs; lower is better. Bold marks the lowest reduced FLOPs within each model block.}
\resizebox{\textwidth}{!}{
\begin{tabular}{l l l r r c}
\toprule
Method & Model & Dataset & Reduced GFLOPs & Total GFLOPs & Reduced FLOPs Percentage \\
\midrule
$\text{\ours}_{\text{LP}}$ & MLP10 & CIFAR-100 & \textbf{468,087.1} & 702,436.8 & \textbf{33.36} \\
\midrule
$\text{\ours}_{\text{LP}}$ & MLP12 & CIFAR-100 & \textbf{476,467.2} & 829,063.7 & \textbf{42.53} \\
\midrule
$\text{\ours}_{\text{LP}}$ & ResNet10 & CIFAR-100 & 6,514,232.3 & 8,164,079.4 & 20.21 \\
$\text{\ours}_{\text{AVG}}$ & ResNet10 & CIFAR-100 & 6,467,557.4 & 8,164,079.4 & 20.78 \\
$\text{EB-LTH}_{30}$ & ResNet10 & CIFAR-100 & 5,972,934.6 & 8,164,079.4 & 26.84 \\
$\text{EB-LTH}_{50}$ & ResNet10 & CIFAR-100 & 5,740,794.2 & 8,164,079.4 & 29.68 \\
$\text{EB-LTH}_{70}$ & ResNet10 & CIFAR-100 & \textbf{2,738,808.3} & 8,164,079.4 & \textbf{66.45} \\
\midrule
$\text{\ours}_{\text{LP}}$ & ResNet18 & CIFAR-100 & 15,725,818.6 & 17,867,337.0 & 11.99 \\
$\text{\ours}_{\text{AVG}}$ & ResNet18 & CIFAR-100 & 15,703,746.9 & 17,867,337.0 & 12.11 \\
$\text{EB-LTH}_{30}$ & ResNet18 & CIFAR-100 & 14,665,884.5 & 17,867,337.0 & 17.92 \\
$\text{EB-LTH}_{50}$ & ResNet18 & CIFAR-100 & 12,604,938.4 & 17,867,337.0 & 29.45 \\
$\text{EB-LTH}_{70}$ & ResNet18 & CIFAR-100 & \textbf{10,355,504.9} & 17,867,337.0 & \textbf{42.04} \\
\midrule
$\text{\ours}_{\text{LP}}$ & VGG11 & CIFAR-100 & 4,484,300.9 & 5,536,216.7 & 19.00 \\
$\text{\ours}_{\text{AVG}}$ & VGG11 & CIFAR-100 & 4,480,040.9 & 5,536,216.7 & 19.08 \\
$\text{EB-LTH}_{30}$ & VGG11 & CIFAR-100 & 4,193,029.7 & 5,536,216.7 & 24.26 \\
$\text{EB-LTH}_{50}$ & VGG11 & CIFAR-100 & 4,390,765.5 & 5,536,216.7 & 20.69 \\
$\text{EB-LTH}_{70}$ & VGG11 & CIFAR-100 & \textbf{2,694,567.0} & 5,536,216.7 & \textbf{51.33} \\
\midrule
$\text{\ours}_{\text{LP}}$ & VGG16 & CIFAR-100 & 7,552,664.7 & 10,691,667.2 & 29.36 \\
$\text{\ours}_{\text{AVG}}$ & VGG16 & CIFAR-100 & 7,532,908.2 & 10,691,667.2 & 29.54 \\
$\text{EB-LTH}_{30}$ & VGG16 & CIFAR-100 & 8,054,772.6 & 10,691,667.2 & 24.66 \\
$\text{EB-LTH}_{50}$ & VGG16 & CIFAR-100 & 6,501,385.0 & 10,691,667.2 & 39.19 \\
$\text{EB-LTH}_{70}$ & VGG16 & CIFAR-100 & \textbf{4,635,359.3} & 10,691,667.2 & \textbf{56.65} \\
\bottomrule
\end{tabular}}
\label{tab:flops_cifar100}
\end{table*}

\begin{table*}[t]
\centering
\scriptsize
\caption{FLOPs comparison on CUB. `Reduced FLOPs' denotes the resulting training cost after simplification, expressed in GFLOPs; lower is better. Bold marks the lowest reduced FLOPs within each model block.}
\resizebox{\textwidth}{!}{
\begin{tabular}{l l l r r c}
\toprule
Method & Model & Dataset & Reduced GFLOPs & Total GFLOPs & Reduced FLOPs Percentage \\
\midrule
$\text{\ours}_{\text{LP}}$ & ResNet10 & CUB & 2,959,982.5 & 3,451,320.3 & 14.24 \\
$\text{\ours}_{\text{AVG}}$ & ResNet10 & CUB & 2,931,039.2 & 3,451,320.3 & 15.07 \\
$\text{EB-LTH}_{30}$ & ResNet10 & CUB & 2,586,576.7 & 3,451,320.3 & 25.06 \\
$\text{EB-LTH}_{50}$ & ResNet10 & CUB & 2,089,449.8 & 3,451,320.3 & 39.46 \\
$\text{EB-LTH}_{70}$ & ResNet10 & CUB & \textbf{1,392,653.8} & 3,451,320.3 & \textbf{59.65} \\
\midrule
$\text{\ours}_{\text{LP}}$ & ResNet18 & CUB & 6,336,440.9 & 7,023,153.1 & 9.78 \\
$\text{\ours}_{\text{AVG}}$ & ResNet18 & CUB & 6,321,479.5 & 7,023,153.1 & 9.99 \\
$\text{EB-LTH}_{30}$ & ResNet18 & CUB & 5,084,861.8 & 7,023,153.1 & 27.60 \\
$\text{EB-LTH}_{50}$ & ResNet18 & CUB & 4,140,481.5 & 7,023,153.1 & 41.05 \\
$\text{EB-LTH}_{70}$ & ResNet18 & CUB & \textbf{2,789,804.7} & 7,023,153.1 & \textbf{60.28} \\
\midrule
$\text{\ours}_{\text{LP}}$ & ResNet34 & CUB & 12,962,096.2 & 14,163,159.6 & 8.48 \\
$\text{\ours}_{\text{AVG}}$ & ResNet34 & CUB & 12,949,153.4 & 14,163,159.6 & 8.57 \\
$\text{EB-LTH}_{30}$ & ResNet34 & CUB & 9,966,783.5 & 14,163,159.6 & 29.63 \\
$\text{EB-LTH}_{50}$ & ResNet34 & CUB & 7,484,140.1 & 14,163,159.6 & 47.16 \\
$\text{EB-LTH}_{70}$ & ResNet34 & CUB & \textbf{4,784,782.8} & 14,163,159.6 & \textbf{66.22} \\
\bottomrule
\end{tabular}}
\label{tab:flops_cub}
\end{table*}

From the FLOPs perspective, EB-LTH is expected to have an advantage, since it is explicitly designed for computational reduction. Even when the parameter reduction is not larger, pruning a fraction of the channels in each layer reduces the amount of computation involved in backpropagation throughout the network, and this is typically the dominant cost during training. By contrast, \ours is designed primarily to identify an early split point that preserves accuracy while simplifying the architecture, rather than to minimize layer-wise training cost directly.

At the same time, the two approaches are not mutually exclusive. In preliminary experiments, we applied $\text{EB-LTH}_{30}$ after \ours on the simplified model for the ResNet10/CIFAR-10 and ResNet18/CIFAR-100 configurations. Even in this strongly compression-oriented setting, the combined strategy still achieved competitive performances of $0.87$ and $0.55$, respectively, while further lowering the resulting FLOPs compared with either method used in isolation. A broader and more systematic study of such hybrid strategies, including combinations with lighter classification heads, is left for future work.

\clearpage


\end{document}